\documentclass[11pt]{article}

\usepackage[margin=1in]{geometry}
\usepackage{amsmath,amssymb,amsthm,mathtools,bm}
\usepackage{booktabs}
\usepackage{enumitem}
\usepackage{natbib}
\usepackage[table]{xcolor}
\usepackage{subfig}
\usepackage{algorithm}
\usepackage{algpseudocode}
\usepackage[hidelinks]{hyperref}
\usepackage{etoc}
\usepackage{tabularx}
\usepackage{fvextra}
\usepackage{upquote}
\fvset{frame=single,framesep=4mm,fontsize=\small,breaklines=true,breaksymbolleft={}}

\newtheorem{remark}{Remark}[section] 
\newtheorem{theorem}{Theorem}[section]
\newtheorem{lemma}{Lemma}[section]
\newtheorem{proposition}{Proposition}[section]
\newtheorem{corollary}{Corollary}[section]
\newtheorem{condition}{Condition}[section]
\theoremstyle{definition}

\newcommand{\E}{\mathbb E}
\newcommand{\Pp}{\mathbb P}
\newcommand{\Var}{\operatorname{Var}}

\newcommand{\R}{\mathbb R}
\newcommand{\1}{\mathbf 1}
\newcommand{\clip}{\operatorname{clip}}
\newcommand{\EL}{\mathrm{EL}}

\usepackage{graphicx,multirow,upgreek,array,fix-cm,float}

\title{Anytime-Valid Empirical Likelihood Inference from LLM Annotations with Adaptive Human Auditing}
\author{Hongfu Gao$^{1}$ \and Jiang Hu$^{2}$ \and Pengfei Liu$^{3}$ \and Wang Zhou$^{1}$}
\date{}
\begin{document}
\maketitle
\begin{center}
\small
$^{1}$Department of Statistics and Data Science, National University of Singapore\\
$^{2}$School of Mathematics and Statistics, Northeast Normal University\\
$^{3}$School of Mathematics and Statistics, Jiangsu Normal University\\[0.5em]
\texttt{gaohongfu@u.nus.edu}, \texttt{huj156@nenu.edu.cn}\\
\texttt{liupengfei@jsnu.edu.cn}, \texttt{wangzhou@nus.edu.sg}
\end{center}
\etocdepthtag.toc{mtmain}
\begin{abstract}
Large language models (LLMs) are increasingly used to annotate unlabeled data as a fast, low-cost alternative to human annotation.
However, their outputs can be biased and miscalibrated relative to human labels, potentially leading to misleading analysis and inference.
We propose AAELI (Adaptive Human Auditing and Anytime-Valid Empirical Likelihood Inference), an online framework for estimating the population mean of human labels under a limited auditing budget.
AAELI uses LLM outputs to guide adaptive human auditing and combines them with audited labels through predictable empirical-likelihood $e$-processes to construct anytime-valid confidence sequences.
Our main theoretical result shows that a predictably learned efficient projection preserves the first-order profile empirical-likelihood statistic under adaptive auditing, without requiring a prescribed convergence rate for the projection estimator.
This yields an exactly anytime-valid betting construction whose continuous-mixture form incurs a scalar rather than vector-dimensional mixing penalty.
Supporting results establish adaptive empirical-likelihood asymptotics and show that a learned audit rule attains the oracle variance within the stated class of augmented estimating-equation designs under the same audit-probability floor and long-run expected budget.
Experiments on synthetic data and three real-world datasets demonstrate that AAELI maintains empirical anytime coverage while requiring fewer human audits than the strongest competing baseline to reach a target precision.
\end{abstract}


\section{Introduction}
\label{sec:auto:1}
\label{sec:introduction}

In many scientific and social applications, obtaining high-quality labeled data remains a major bottleneck. 
Such labels typically require human expert annotation, which is costly and time-consuming. 
As a cheaper and faster alternative, large language models (LLMs) are increasingly used as automatic judges or annotators \citep{cowenbreen2026multiprediction,song2026demystifying}.
In reinforcement learning, they serve as reward models that score model outputs
\citep{gunjal2026rubrics}.
In content moderation, they detect misinformation in news~\citep{gabriel2022misinfo}.
And in political science, they assess the political leaning of opinion articles \citep{li2026robust}.
These applications show that LLMs are increasingly relied upon to extract information from large-scale unlabeled data.

Nevertheless, using LLM outputs for statistical inference in online annotation pipelines raises two challenges.
First, LLM judgments can be biased and miscalibrated, and different LLMs often make highly correlated errors \citep{chen2024humans,kim2025correlated}.
Treating these judgments as ground truth can be problematic.
When used as reward signals in reinforcement learning, they may encourage reward hacking \citep{thaman2026reward}.
When used as substitutes for human labels, they may lead to misleading statistical conclusions.
This motivates using a limited number of human audits to correct LLM-induced bias and support valid inference.
Second, annotations often arrive sequentially, as in Chatbot Arena \citep{chiang2024chatbot}, and practitioners may monitor results and stop once a target precision is reached.
Fixed-time inference generally does not retain its nominal guarantees under such data-dependent stopping \citep{ramdas2023game}.

In this article, we focus on inference for the population mean of human labels. 
This target is a basic quantity in many evaluation tasks: it measures how often an LLM agrees with human preference \citep{chiang2024chatbot}, or the prevalence of a human-defined attribute, such as political leaning or misinformation, in a population of texts \citep{li2026robust,gabriel2022misinfo}.
Although human labels can be obtained for only a limited subset of instances, low-cost LLM judgments are available for all instances.
We use these judgments to guide human auditing and sharpen inference for the population mean, making efficient use of the human-auditing budget while maintaining anytime validity.

\subsection{Related Works}
\label{sec:auto:1.1}
\label{sec:related_work}

Although quite different in scope, our work is inspired by prediction-powered inference (PPI) \citep{angelopoulos2023prediction}. 
PPI and its extensions
\citep{einbinder2025semi,kluger2025prediction,angelopoulos2026ppi++,chen2026noisy,cowenbreen2026multiprediction,lee2026efficient,ji2026predictions} combine machine learning model predictions with a small set of human labels to correct prediction bias. 
These methods use model predictions to improve estimation, but not to decide which instances receive human labels. 
As a result, human labels may be spent on instances where the predictions are already accurate, rather than where they are most informative. 

To reduce human labeling costs, active statistical inference (ASI) and subsequent work \citep{zrnic2024active,chen2026balanced,li2026robust,sfyraki2026revisiting} use model predictions and uncertainties to guide which instances receive human labels, and correct the resulting selection bias by inverse-probability weighting (IPW).
ASI further derives the variance-minimizing sampling rule for a given predictor, but takes the predictor as given and does not characterize how auxiliary information built from LLM outputs and the audit rule jointly determine the attainable efficiency. More broadly, allocating measurement effort in proportion to conditional standard deviations is classical in survey sampling and adaptive experimental design \citep{neyman1934,hahn2011adaptive}, and sequential designs that learn such allocations online and attain the semiparametric efficiency bound have been developed for average treatment effects \citep{kato2020efficient,cook2024semiparametric}. 
These methods rely on a single plug-in AIPW estimator with fixed-time or asymptotic guarantees.
They neither combine multiple auxiliary moments derived from LLM outputs nor provide exact anytime-valid inference.

Empirical likelihood has a long history as a method for incorporating auxiliary information through moment conditions.
It yields likelihood-ratio confidence sets with asymptotic chi-squared calibration \citep{owen1988single,owen1990regions,owen2001book}, and combines valid moment conditions to attain semiparametric efficiency when they are sufficiently informative \citep{qin1994empirical}. 
Two-sample empirical likelihood \citep{chen2003information} combines a small validation sample containing the primary outcome with a large sample containing only surrogates, and EPI \citep{wang2025epi} extended this idea to prediction-powered inference. 
However, EPI takes the labeled and unlabeled samples as given, with i.i.d.\ labeled data collected in advance, and does not consider how instances are selected for labeling. 
When labels are instead acquired adaptively based on predictions and previously audited data, the resulting estimating equations form a martingale difference sequence rather than an i.i.d.\ sample, so the existing theory does not directly apply.

Fixed-time confidence intervals generally do not guarantee nominal coverage under data-dependent stopping, whereas confidence sequences remain valid uniformly over time \citep{howard2021time,ramdas2023game}. 
They have been constructed via betting for bounded means \citep{waudbysmith2024betting} and asymptotically for general estimators \citep{waudby2024time}. 
ABPPI \citep{kilian2026anytime} extends asymptotic confidence sequences to prediction-powered inference, but assumes i.i.d.\ labeled data that are not selected adaptively.
For bounded mean estimation, ASI \citep{zrnic2024active} explicitly describes how betting methods applied to its sequential AIPW increments yield time-uniform confidence sequences.
Our additional contribution is the predictable estimation of an efficient combination of multiple moments and its first-order connection to profile empirical likelihood under adaptive auditing.
In off-policy evaluation, betting-based confidence sequences have been built from IPW scores under history-dependent logging policies \citep{waudbysmith2024offpolicy}, and asymptotic confidence sequences are available for AIPW estimators under adaptive designs \citep{cook2024semiparametric}.
These methods rely on a single weighted score and do not combine multiple auxiliary moments.
The connection between predictable betting under adaptive auditing and profile empirical likelihood remains to be established.

\subsection{Contributions}
\label{sec:auto:1.2}

In this article, we develop an empirical-likelihood framework that uses LLM judgments to guide adaptive human auditing, improving human-label efficiency while providing anytime-valid confidence sequences for the population mean. 
We propose a novel method, Adaptive Human Auditing and Anytime-Valid Empirical Likelihood Inference (AAELI). 
Methodologically, AAELI consists of three main steps. 
First, it constructs augmented estimating equations that combine inverse-probability-weighted human labels with LLM-based auxiliary moments.
These equations have conditional mean zero at the true population mean and are integrated through empirical likelihood.
Second, it allocates human audits in proportion to the conditional standard deviation of the human label, estimated sequentially from previously audited data. 
Third, it constructs predictable empirical-likelihood $e$-processes and inverts them to obtain anytime-valid confidence sequences.

Our main theoretical contribution connects profile empirical likelihood to predictable efficient betting under adaptive auditing.
Theorem~\ref{thm:profile-bridge} shows that projecting the estimating equations onto an efficient direction learned from past observations preserves the first-order profile empirical-likelihood statistic, without a prescribed convergence rate for the direction estimator.
For continuous mixtures, this reduces the mixing correction from $q\log n$ for the full moment vector to $\log n$ for the scalar projection, while retaining exact anytime validity.
The contribution is this profile-equivalence result for a learned predictable projection.
Exact validity itself follows from the martingale betting construction.
Two supporting developments connect this result to human auditing.
First, we establish asymptotic normality and a profile Wilks theorem for the adaptively generated estimating equations (Theorem~\ref{thm:adaptive-wilks}).
Second, using the classical AIPW variance decomposition and constrained Neyman allocation, we characterize the oracle audit variance and show that it can be attained by sequential learning within the stated class of augmented estimating-equation designs (Theorem~\ref{thm:eff-decomp}, Proposition~\ref{prop:adaptive-variance-bound}, and Corollary~\ref{cor:oracle-recovery}).
A complementary information bound identifies the largest deterministic local log-evidence coefficient attainable in probability under the stated local submodel, which the continuous-mixture construction attains under the corresponding assumptions (Proposition~\ref{prop:evidence-bound}).

Experiments on synthetic and real-world datasets show that AAELI maintains anytime coverage while requiring fewer human labels than existing baselines to reach a target precision.
In real-world evaluations, we apply AAELI with three LLM judges to three datasets: Chatbot Arena \citep{chiang2024chatbot}, AllSides \citep{baly2020we}, and Misinfo \citep{gabriel2022misinfo}. 
For example, on Chatbot Arena, AAELI needs about 870 human labels on average to reach a confidence-sequence width of 0.10, compared with about 1{,}100 for the strongest baseline, RASI, a saving of roughly 21\%.

\subsection{Organization}
\label{sec:auto:1.3}
The rest of the paper is organized as follows.
Section~\ref{sec:setup} introduces the problem formulation.
Section~\ref{sec:methodology} proposes the AAELI method, including its augmented estimating equations, adaptive auditing rule, and predictable empirical-likelihood confidence sequences.
Section~\ref{sec:theory} establishes the theoretical properties of AAELI, including asymptotic normality and Wilks theorems, the efficiency decomposition and oracle audit rule, and predictable empirical-likelihood $e$-processes together with their connection to raw and profile empirical likelihood.
Section~\ref{sec:synthetic} evaluates AAELI in simulations, and Section~\ref{sec:real} applies it to three real-world LLM-evaluation datasets.
Section~\ref{sec:conclusion} concludes the article with discussions.
All technical proofs and additional results are collected in the appendix.

\section{Problem Formulation}
\label{sec:auto:2}

\label{sec:setup}
We consider a sequence of instances $X_t$, $t=1,2,\ldots$, drawn independently from an unknown, fixed distribution $\Pp_X$.
Each instance $X_t$ is associated with a ground-truth label $Y_t\in[0,1]$, drawn conditionally on $X_t$ from an unknown, fixed conditional distribution $\Pp_{Y\mid X}$.
The labels are initially unobserved and costly to obtain through human annotation.
For every instance, however, we observe the outputs $J_t=(J_{t1},\ldots,J_{tK})$ of $K$ low-cost LLM judges, which may be biased or miscalibrated.
We assume that the triples $(X_t,J_t,Y_t)$ are i.i.d.\ across $t$ and write $Z_t=(X_t,J_t)$ for all information available before annotation.
Our target is the population mean
\begin{equation}
\mu_0=\E[Y_t],
\label{eq:target}
\end{equation}
which we estimate by combining LLM judgments available for all instances with selectively acquired human labels under a long-run audit budget $\rho\in(0,1)$.


Human labels are acquired sequentially.
As each instance arrives, we observe $Z_t$ and decide at random whether to send it for human annotation.
Let $R_t\in\{0,1\}$ indicate this decision, so that $Y_t$ is observed only when $R_t=1$.
Let $\mathcal F_t=\sigma\bigl(\mathcal F_0,\{Z_s,R_s,R_sY_s:1\le s\le t\}\bigr)$ denote the observed history and $\mathcal P_t=\sigma(\mathcal F_{t-1},Z_t)$ the information available when the audit decision at time $t$ is made.
The audit probability $\pi_t$ is $\mathcal P_t$-measurable: it may depend on the current pre-audit information and the entire observed history, but not on the unobserved label $Y_t$.
Given $\pi_t$, we draw $R_t=\1\{U_t\le\pi_t\}$, where $U_1,U_2,\ldots$ are i.i.d.\ $\operatorname{Unif}(0,1)$ variables independent of $\mathcal F_0$ and of $\{(Z_t,Y_t)\}_{t\ge1}$, so that $R_t\mid Z_t,Y_t,\mathcal F_{t-1}\sim\operatorname{Bernoulli}(\pi_t)$.
We impose two constraints on $\pi_t$.
First, $\underline\pi\le\pi_t\le1$ for a fixed $\underline\pi\in(0,\rho)$, so that every instance has a positive chance of being audited.
Second, the long-run expected audit rate respects the budget, $\limsup_{n\to\infty}n^{-1}\sum_{t=1}^n\E[\pi_t]\le\rho$.
Since $\E[R_t]=\E[\pi_t]$, this also bounds the long-run expected fraction of human audits.
These requirements are defined formally in Condition~\ref{cond:basic}.

We construct a $(1-\alpha)$ confidence sequence, that is, $\mathcal F_n$-measurable sets $\mathcal C_n$ satisfying
\begin{equation}
\Pp\left(\mu_0\in\mathcal C_n\text{ for all }n\ge1\right)\ge1-\alpha,
\label{eq:anytime-coverage}
\end{equation}
where $\alpha\in(0,1)$ is a prescribed error level.
This property, known as anytime validity, implies $\Pp(\mu_0\in\mathcal C_N)\ge1-\alpha$ for every almost surely finite stopping time $N$.
Hence, we may update the confidence set and stop annotation once the desired precision is reached.
While maintaining this guarantee and respecting the annotation budget, we seek to reduce the number of human labels, $\sum_{t\le n}R_t$, needed to attain a desired width
\begin{equation}
\operatorname{Width}_n=\sup\mathcal C_n-\inf\mathcal C_n .
\label{eq:cs-width}
\end{equation}

\section{Methodology}
\label{sec:auto:3}
\label{sec:methodology}
Our method consists of three main components.
We first construct estimating equations whose conditional mean is zero at $\mu_0$ under adaptive auditing.
We then use an efficiency analysis to guide human audit allocation and learn the auditing rule sequentially from previously audited data. 
Finally, we construct predictable empirical-likelihood $e$-processes and invert them to obtain anytime-valid confidence sequences.

\subsection{Empirical Likelihood for Augmented Estimating Equations}
\label{sec:auto:3.1}
\label{subsec:ee}

We first construct estimating equations for $\mu_0$.
For each instance $t$, we observe the known human audit probability $\pi_t$, the pre-audit information $Z_t$, and the audit indicator $R_t$.
Since $\pi_t$ depends on $Z_t$ and $Z_t$ is predictive of $Y_t$, the average of the audited labels is generally biased for $\mu_0$.
Inverse-probability weighting (IPW) \citep{horvitz1952generalization,robins1994estimation,zrnic2024active} corrects for this selective auditing by upweighting labels that are less likely to be observed, which yields the score
\[
A_t(\mu)=\frac{R_t}{\pi_t}(Y_t-\mu).
\]
Because $R_t\mid Z_t,Y_t,\mathcal F_{t-1}\sim\operatorname{Bernoulli}(\pi_t)$ with $\mathcal P_t$-measurable $\pi_t$ (Section~\ref{sec:setup}), we have $\E[R_t/\pi_t\mid Z_t,Y_t,\mathcal F_{t-1}]=1$, and hence $\E[A_t(\mu)\mid\mathcal F_{t-1}]=\mu_0-\mu$.
However, $A_t(\mu)$ vanishes for unaudited instances, ignores the predictive information in $Z_t$, and can have large variance when audit probabilities are small.

To recover this information, we use the LLM outputs as control variates.
Let $m(z)=\E[Y_t\mid Z_t=z]$, and let $\widehat m_{t-1}$ be an estimator of $m$ trained on the observed history $\mathcal F_{t-1}$.
Let $h_t\in\R^r$ be any $\mathcal P_t$-measurable auxiliary vector, such as $h_t=(1,\widehat m_{t-1}(Z_t))^\top$.
More generally, $h_t$ may contain the raw scores of the individual LLM judges.
Applying the same weighting to $h_t$ and subtracting its unweighted value gives
\[
B_t=\left(\frac{R_t}{\pi_t}-1\right)h_t.
\]
If instance $t$ is audited, then $B_t=(1/\pi_t-1)h_t$; otherwise $B_t=-h_t$.
This term compares the audited instances with all instances in terms of their LLM outputs, so unaudited instances also contribute through $B_t$.
It satisfies $\E[B_t\mid\mathcal F_{t-1}]=0$, so it does not change the target.
When $h_t$ captures variation in the conditional human mean, $B_t$ is correlated with $A_t(\mu_0)$, allowing a suitable control-variate adjustment to reduce variance.
The variance-minimizing adjustment coefficient depends on the unknown joint distribution of the human labels and the LLM judgments.
We therefore stack the two components,
\begin{equation}
G_t(\mu)=\begin{pmatrix}A_t(\mu)\\B_t\end{pmatrix}\in\R^q,
\quad q=r+1,
\label{eq:ee}
\end{equation}
and let empirical likelihood determine the combination.
The following proposition establishes that these stacked equations identify $\mu_0$ and retain a martingale difference structure under adaptive auditing.

\begin{proposition}
\label{prop:estimating-equations}
Suppose that Condition~\ref{cond:basic}~(a)--(c) holds and that $h_t$ is $\mathcal P_t$-measurable with $\E\|h_t\|<\infty$ for every $t\ge1$.
Then, for every $\mu\in\R$,
\[
\E[G_t(\mu)\mid\mathcal F_{t-1}]=\begin{pmatrix}\mu_0-\mu\\0_r\end{pmatrix}.
\]
Consequently, the conditional moment equation uniquely identifies $\mu_0$, and $\{G_t(\mu_0)\}_{t\ge1}$ is a martingale difference sequence with respect to $(\mathcal F_t)_{t\ge0}$.
\end{proposition}

Proposition~\ref{prop:estimating-equations} remains valid even when $\pi_t$ and $h_t$ depend on the entire observed history.
It requires no unbiasedness, calibration, or independence assumptions on the LLM judges.
The resulting martingale structure underlies the confidence sequences in Section~\ref{subsec:AAELI}.

We now make the combination in \eqref{eq:ee} explicit through empirical likelihood.
Let $\Theta\subset[0,1]$ be a compact interval containing $\mu_0$ in its interior.
Given $G_1(\mu),\ldots,G_n(\mu)$, define the empirical likelihood ratio
\begin{equation}
\mathcal R_n(\mu)=\max\left\{\prod_{t=1}^n np_t:\ p_t\ge0,\ \sum_{t=1}^np_t=1,\
\sum_{t=1}^np_tG_t(\mu)=0\right\},
\label{eq:el-ratio}
\end{equation}
with $\mathcal R_n(\mu)=0$ when the constraint set is empty, and the maximum empirical likelihood estimator $\widehat\mu_n\in\arg\max_{\mu\in\Theta}\mathcal R_n(\mu)$.
Since $\mathcal R_n$ is upper semicontinuous on the compact set $\Theta$, this maximum is attained and a measurable maximizer exists (Section~\ref{app:subsection:A.2} of the appendix).
Since $G_t(\mu)$ stacks one score equation with $r$ auxiliary equations for a scalar parameter, the system is overidentified.
Empirical likelihood then reweights the instances so that the auxiliary equations hold exactly in the sample.
This reweighting carries out the control-variate adjustment automatically.
At a fixed terminal time $n$, under the null hypothesis $\mu_0=\mu$, the profile statistic $-2\log\{\mathcal R_n(\mu)/\mathcal R_n(\widehat\mu_n)\}$ is asymptotically $\chi^2_1$ (Theorem~\ref{thm:adaptive-wilks}), giving the asymptotically valid fixed-sample confidence set $\{\mu\in\Theta:-2\log\{\mathcal R_n(\mu)/\mathcal R_n(\widehat\mu_n)\}\le\chi^2_{1,1-\alpha}\}$.
We next study how the auxiliary vector and the audit probabilities determine the variance of the combination of these equations.

\subsection{Efficiency-Guided Adaptive Auditing}
\label{sec:auto:3.2}
\label{subsec:efficiency}
We first consider a fixed design: let $(Z,Y)$ follow the common distribution of $(Z_t,Y_t)$.
Fix an audit rule $\pi(Z)\in[\underline\pi,1]$ and a bounded auxiliary vector $h(Z)\in\R^r$, and let $R\mid Z,Y\sim\operatorname{Bernoulli}(\pi(Z))$.
Define $A=\frac{R}{\pi(Z)}(Y-\mu_0)$ and $B=\left(\frac{R}{\pi(Z)}-1\right)h(Z)$.
As in Section~\ref{subsec:ee}, $A$ and $B$ have mean zero, so the adjusted score $A-b^\top B$ is an unbiased estimating function for every $b\in\R^r$.
Under the regularity conditions of Theorem~\ref{thm:adaptive-wilks}, the limiting variance of $\sqrt n(\widehat\mu_n-\mu_0)$ under this fixed design is
\begin{equation}
V_{\EL}
=
\inf_{b\in\R^r}\Var(A-b^\top B).
\label{eq:optimal-augmentation}
\end{equation}
We write $b^*$ for the minimizer, so that $A-(b^*)^\top B$ is the efficient combination.
This motivates choosing $h$ and $\pi$ to make $V_{\EL}$ small.
Let $\sigma^2(z)=\Var(Y\mid Z=z)$.
Theorem~\ref{thm:eff-decomp} below shows that $V_{\EL}$ decomposes into an auxiliary approximation error plus the benchmark
\begin{equation}
V_{\mathrm{eff}}(\pi)= \Var\{m(Z)\} + \E\left[\frac{\sigma^2(Z)}{\pi(Z)}\right].
\label{eq:efficient-variance}
\end{equation}
The approximation error vanishes when $h(Z)=(1,m(Z))^\top$: with $b=(-\mu_0,1)^\top$, the adjusted score becomes the AIPW score $m(Z)-\mu_0+\frac{R}{\pi(Z)}\{Y-m(Z)\}$, whose variance equals $V_{\mathrm{eff}}(\pi)$, the semiparametric efficiency bound for a fixed audit rule \citep{robins1994estimation}.
This suggests two design choices.
First, we include $\widehat m_{t-1}(Z_t)$ in $h_t$ to approximate the optimal auxiliary vector.
Second, since only $\E[\sigma^2(Z)/\pi(Z)]$ in $V_{\mathrm{eff}}(\pi)$ depends on the audit rule, we choose $\pi$ to minimize this term, which yields the oracle rule
$\pi^*(Z)=\clip_{[\underline\pi,1]}\{\sigma(Z)/\tau^*\}$ with $\E[\pi^*(Z)]=\rho$ (Theorem~\ref{thm:oracle-annotation}), where $\clip_{[a,b]}(x)=\min\{\max(x,a),b\}$ and $V_{\mathrm{oracle}}=V_{\mathrm{eff}}(\pi^*)$.

Since $m$, $\sigma$, and $\tau^*$ are unknown, we learn them online.
During an initial burn-in, we audit uniformly with $\pi_t=\rho$ and set $\widehat m_{t-1}\equiv1/2$, so that the two auxiliary components of $B_t$ are collinear; this affects neither Proposition~\ref{prop:estimating-equations} nor anytime validity, and the regularization in \eqref{eq:predictable-direction} keeps the estimated direction well defined.
The burn-in ends at a time $n_0$, which may be deterministic or a stopping time with $n_0<\infty$ almost surely.
In our experiments, $n_0$ is the first time at which at least $30$ labels are available.
After the burn-in, each time $t$ consists of three steps, all using only data observed before time $t$.

We first fit $\widehat m_{t-1}$ over a function class $\mathcal M$ with values in $[0,1]$ by minimizing the IPW loss
\begin{equation}
\widehat L_{t-1}(f)=\frac1{t-1}\sum_{s<t}\frac{R_s}{\pi_s}\{Y_s-f(Z_s)\}^2,
\label{eq:ipw-learning}
\end{equation}
where the weights correct for selective auditing.
We estimate $s_2(z)=\E[Y^2\mid Z=z]$ by the same loss with response $Y_s^2$ and set $\widehat\sigma_{t-1}=\{[\widehat s_{2,t-1}-\widehat m_{t-1}^2]_+\}^{1/2}$.

Because $Z_s$ is observed for every instance, audited or not, the threshold can be calibrated without labels.
Fix a compact interval $T=[\tau_-,\tau_+]\subset(0,\infty)$ and choose a measurable
\begin{equation}
\widehat\tau_{t-1}\in\arg\min_{\tau\in T}\Bigl|\frac1{t-1}\sum_{s<t}\clip_{[\underline\pi,1]}\Bigl\{\frac{\widehat\sigma_{t-1}(Z_s)}{\tau}\Bigr\}-\rho\Bigr|,
\label{eq:empirical-budget}
\end{equation}
which is well defined even when the empirical budget equation has no exact root.
An external unlabeled pool in $\mathcal F_0$, if available, may be added to the calibration sample.

After observing $Z_t$, we set
\begin{equation}
\pi_t=\clip_{[\underline\pi,1]}\left\{\frac{\widehat\sigma_{t-1}(Z_t)}{\widehat\tau_{t-1}}\right\},
\quad
h_t=\begin{pmatrix}1\\ \widehat m_{t-1}(Z_t)\end{pmatrix},
\label{eq:learned-allocation}
\end{equation}
draw $R_t\sim\operatorname{Bernoulli}(\pi_t)$, and observe $Y_t$ if $R_t=1$.
In practice, the nuisance fits and the threshold may be updated only every $n_{\mathrm{up}}$ steps, with the most recent fit used in between.
Since $\pi_t$ and $h_t$ are $\mathcal P_t$-measurable, Proposition~\ref{prop:estimating-equations} applies.
Under consistent estimation, the procedure attains the oracle variance and meets the budget asymptotically (Corollary~\ref{cor:oracle-recovery}).

\subsection{Predictable Empirical-Likelihood Confidence Sequences}
\label{sec:auto:3.3}
\label{subsec:AAELI}

We construct anytime-valid confidence sequences by combining the components of $G_t(\mu)$ into a scalar betting score $a^\top G_t(\mu)$ for some direction $a\in\R^q$.
Two quantities determine how useful such a score is.
By Proposition~\ref{prop:estimating-equations}, $\E[a^\top G_t(\mu)\mid\mathcal F_{t-1}]=a^\top D\,(\mu-\mu_0)$ with $D=-e_1$, where $e_1$ denotes the first standard basis vector of $\R^q$, so $a^\top D$ measures how strongly the score reacts to a wrong candidate mean.
Its noise is measured by the variance $a^\top\Omega a$, where $\Omega$ denotes the long-run covariance of $G_t(\mu_0)$ (formalized in Condition~\ref{cond:cov}).
The squared signal-to-noise ratio $\mathcal I(a)=\frac{(a^\top D)^2}{a^\top\Omega a}$ is maximized, by the Cauchy--Schwarz inequality, at directions proportional to $\Omega^{-1}D$, with maximum $D^\top\Omega^{-1}D=V_{\EL}^{-1}$.
We normalize the maximizer as $a^*=\Omega^{-1}D/\{D^\top\Omega^{-1}D\}^{1/2}$, so that $(a^*)^\top\Omega a^*=1$.
This direction recovers the efficient combination of Section~\ref{subsec:efficiency}:
$(a^*)^\top G_t(\mu)=-\{A_t(\mu)-(b^*)^\top B_t\}/V_{\EL}^{1/2}$.

Since $\Omega$ is unknown, we estimate $a^*$ at each time $t$ from observations before time $t$, separately for each candidate mean.
For $t\ge2$ and each $\mu\in\Theta$, define
\begin{equation}
\widehat\Omega_{t-1}(\mu)=\varepsilon_{t-1}I_q+\frac1{t-1}\sum_{s<t}G_s(\mu)G_s(\mu)^\top,
\quad
\widetilde a_{t-1}(\mu)=\frac{\widehat\Omega_{t-1}(\mu)^{-1}D}{\{D^\top\widehat\Omega_{t-1}(\mu)^{-1}D\}^{1/2}}.
\label{eq:predictable-direction}
\end{equation}
The regularization $\varepsilon_t>0$, with $\varepsilon_t\downarrow0$, keeps $\widehat\Omega_{t-1}(\mu)$ invertible, in particular during burn-in, when $h_t=(1,1/2)^\top$ makes the auxiliary components of $G_t(\mu)$ collinear.
To keep scores bounded, fix $C_a>0$, set $\widehat a_{t-1}(\mu)=\widetilde a_{t-1}(\mu)\min\{1,C_a/\|\widetilde a_{t-1}(\mu)\|\}$ for $t\ge2$, and initialize $\widehat a_0(\mu)=-\min\{1,C_a\}e_1$.
Then $\|\widehat a_{t-1}(\mu)\|\le C_a$ and $\widehat a_{t-1,1}(\mu)<0$ for every $t\ge1$ and $\mu\in\Theta$.
The scalar score is
\begin{equation}
S_t(\mu)=\widehat a_{t-1}(\mu)^\top G_t(\mu).
\label{eq:predictable-score}
\end{equation}
Because $\widehat a_{t-1}(\mu)$ is $\mathcal F_{t-1}$-measurable, Proposition~\ref{prop:estimating-equations} gives $\E[S_t(\mu_0)\mid\mathcal F_{t-1}]=0$, and the norm bound $C_a$ keeps $S_t(\mu)$ bounded.

Let $\widehat a_{t-1}(\mu)=(\widehat a_{t-1,1}(\mu),\widehat a_{t-1,B}(\mu)^\top)^\top$, $b_{t-1}=-\widehat a_{t-1,B}(\mu)/\widehat a_{t-1,1}(\mu)$, and $S_t(\mu)=\widehat a_{t-1,1}(\mu)\{A_t(\mu)-b_{t-1}^\top B_t\}$.
For $h_t=(1,f_t(Z_t))^\top$ and $b_{t-1}=(b_{0,t-1},b_{1,t-1})^\top$, define $f_{t,\mu}^{\mathrm{cal}}(z)=\mu+b_{0,t-1}+b_{1,t-1}f_t(z)$.
Then
\[
A_t(\mu)-b_{t-1}^\top B_t=f_{t,\mu}^{\mathrm{cal}}(Z_t)-\mu+\frac{R_t}{\pi_t}\{Y_t-f_{t,\mu}^{\mathrm{cal}}(Z_t)\}.
\]
The standard AIPW score is the special case $b_{t-1}=(-\mu,1)^\top$, that is, $f_{t,\mu}^{\mathrm{cal}}=f_t$.
Thus AAELI bets on a scaled AIPW score with predictably learned calibration and normalization; matching these quantities and the betting fractions yields the same calibrated AIPW bettor.
The empirical-likelihood formulation identifies the efficient projection and establishes that learning it predictably preserves the first-order profile statistic (Theorem~\ref{thm:profile-bridge}).

Suppose $\sup_t\|h_t\|\le C_h$ for a deterministic constant $C_h$.
Together with $\pi_t\ge\underline\pi$, $Y_t\in[0,1]$, and $\mu\in\Theta$, this gives a deterministic bound $\sup_{t,\,\mu\in\Theta}\|G_t(\mu)\|\le C_G$ and $|S_t(\mu)|\le C_S=C_aC_G$.
Fix $\eta\in(0,1)$ and a prespecified probability distribution $\varpi$ over betting fractions $\beta\in I_\eta=[-\eta/C_S,\eta/C_S]$, which keeps every betting factor positive.
Define
\begin{equation}
E_n(\mu)=\int_{I_\eta}\prod_{t=1}^n\{1+\beta S_t(\mu)\}\,\varpi(d\beta),\quad E_0(\mu)=1.
\label{eq:profile-eprocess}
\end{equation}
Let $\widetilde{\mathcal R}_n(\mu)$ denote the empirical likelihood ratio with the single constraint $\sum_{t=1}^np_tS_t(\mu)=0$.
For each fixed $\beta$, the product $\prod_{t\le n}\{1+\beta S_t(\mu)\}$ is both the wealth of a bettor against $H_\mu:\mu_0=\mu$ and the exponentiated dual criterion of this scalar empirical likelihood.
Writing $\mathcal K_n(\mu)=\{\beta:1+\beta S_t(\mu)>0\text{ for all }t\le n\}$, whenever $0$ lies in the interior of the convex hull of $\{S_t(\mu)\}_{t\le n}$ we have $\sup_{\beta\in\mathcal K_n(\mu)}\sum_{t\le n}\log\{1+\beta S_t(\mu)\}=-\log\widetilde{\mathcal R}_n(\mu)$, with a unique maximizer.
Since $I_\eta\subseteq\mathcal K_n(\mu)$, the supremum over $I_\eta$ coincides with it whenever this maximizer lies in $I_\eta$ (Lemma~\ref{lem:scalar-dual} of the appendix).
The $e$-process \eqref{eq:profile-eprocess} therefore integrates the exponentiated empirical-likelihood dual criterion over a prespecified safe set of multipliers; its vector analogue, which mixes over all $q$ multipliers, is studied in Section~\ref{subsec:theory-cs}.
Since $\E[S_t(\mu)\mid\mathcal F_{t-1}]=\widehat a_{t-1,1}(\mu)(\mu_0-\mu)$ with $\widehat a_{t-1,1}(\mu)<0$, incorrect candidate means induce a directional drift, and mixing over $\beta$ adapts to its unknown size.
For $\alpha\in(0,1)$, we retain candidate means whose evidence has not reached $1/\alpha$:
\begin{equation}
\mathcal C_n=\{\mu\in\Theta:E_n(\mu)<1/\alpha\}.
\label{eq:confidence-sequence}
\end{equation}
Because each betting factor has conditional mean one at $\mu_0$, $\mathcal C_n$ is an anytime-valid confidence sequence (Theorem~\ref{thm:anytime-validity}).
Since $\mathcal C_n$ need not be an interval, we report its convex hull, which contains $\mu_0$ whenever $\mathcal C_n$ does and therefore inherits anytime validity.
When the conditions of Corollary~\ref{cor:oracle-recovery} hold, $V_{\EL}=V_{\mathrm{oracle}}$, which governs the local evidence expansions of the continuous-mixture construction in Corollary~\ref{cor:alternatives}.
The complete procedure is summarized in Algorithm~\ref{alg:aaeli}.

\begin{algorithm}[t]
\caption{Adaptive Human Auditing and Anytime-Valid Empirical Likelihood Inference}
\label{alg:aaeli}
\begin{algorithmic}[1]
\Require
\Statex A stream of unlabeled instances $X_1,X_2,\ldots$; $K$ LLM judges $\mathcal J_1,\ldots,\mathcal J_K$; target audit rate $\rho$, error level $\alpha$, target width $w^\star$, and burn-in length $n_0$; function class $\mathcal M$ and clipping constant $C_a$.
\State Initialize $E_0(\mu)=1$ in \eqref{eq:profile-eprocess} and $\widehat a_0(\mu)=-\min\{1,C_a\}e_1$ for all $\mu\in\Theta$.
\For{$t=1,2,\ldots$}
  \State Observe $X_t$, query the judges, and set
  $Z_t=(X_t,\mathcal J_1(X_t),\ldots,\mathcal J_K(X_t))$.
  \If{$t\le n_0$} \Comment{\textit{Adaptive Auditing}}
    \State Set $\pi_t=\rho$ and $h_t=(1,1/2)^\top$.
  \Else
    \State Fit $\widehat m_{t-1}$ and $\widehat\sigma_{t-1}$ over $\mathcal M$ by the IPW loss
    \eqref{eq:ipw-learning}, and choose $\widehat\tau_{t-1}$ by \eqref{eq:empirical-budget}.
    \State Set $\pi_t$ and $h_t$ by \eqref{eq:learned-allocation}.
  \EndIf
  \State Draw $R_t\sim\operatorname{Bernoulli}(\pi_t)$; if $R_t=1$, request the human label
  $Y_t$.
  \State For each $\mu\in\Theta$, form $G_t(\mu)$ by \eqref{eq:ee}. \Comment{\textit{Augmented Estimating Equations}}
\For{each $\mu\in\Theta$}
  \Comment{\textit{Predictable Empirical-Likelihood $e$-Process}}
  \State Compute $\widehat a_{t-1}(\mu)$ by \eqref{eq:predictable-direction} and clip it at $C_a$.
  \State Set $S_t(\mu)$ by \eqref{eq:predictable-score};
  compute $E_t(\mu)$ by \eqref{eq:profile-eprocess}.
\EndFor
\State Form $\mathcal C_t$ by \eqref{eq:confidence-sequence}.
\If{$\mathcal C_t\neq\emptyset$ and
     $\operatorname{Width}_t\le w^\star$}
  \State \textbf{stop}
\EndIf
\EndFor
\Ensure $\{\mathcal C_t\}_{t\ge1}$.
\end{algorithmic}
\end{algorithm}

\section{Theoretical Justification}
\label{sec:auto:4}
\label{sec:theory}

We first show that empirical likelihood retains its asymptotic properties under adaptive auditing
and identify the variance $V_{\EL}$ of the efficient combination (Section~\ref{subsec:theory-el}).
We then characterize how the auxiliary information and the audit rule determine $V_{\EL}$
(Section~\ref{subsec:theory-efficiency}).
Finally, we show that mixing the empirical-likelihood dual over its multipliers yields exactly anytime-valid confidence sequences, and that, at the level of the local quadratic approximation, profiling reduces the smooth-mixture correction from $q\log n$ to $\log n$ while recovering the local structure of profile empirical likelihood (Section~\ref{subsec:theory-cs}).
Proofs are deferred to the appendix.

\subsection{Adaptive Empirical Likelihood}
\label{sec:auto:4.1}
\label{subsec:theory-el}

Because the human audit probabilities and auxiliary vectors depend on the history, the estimating
equations $G_t(\mu)$ need not be independent or identically distributed.
We establish asymptotic normality of the empirical-likelihood estimator and a profile Wilks theorem under adaptive auditing.
This requires the conditional covariance of the estimating equations to stabilize.

\begin{condition}
\label{cond:basic}
\begin{enumerate}
\renewcommand{\labelenumi}{(\alph{enumi})}
\item
The pairs $(Z_t,Y_t)$ are i.i.d., with $0\le Y_t\le1$,
and $\mu_0=\E[Y_t]$.
The parameter space $\Theta\subseteq[0,1]$ is a compact
interval containing $\mu_0$ in its interior.

\item
Let $U_t\overset{\mathrm{i.i.d.}}{\sim}\operatorname{Unif}(0,1)$
be independent of the entire sequence
$\{(Z_t,Y_t)\}_{t\ge1}$.
Let $\mathcal F_0$ contain any external training or calibration data, together with all auxiliary randomization used by the learning algorithms, and be independent of $\{(Z_t,Y_t),U_t:t\ge1\}$.
Define recursively $\mathcal P_t=\sigma(\mathcal F_{t-1},Z_t)$, $R_t=\mathbf{1}\{U_t\le\pi_t\}$, and $\mathcal F_t
=\sigma(\mathcal F_{t-1},Z_t,R_t,R_tY_t)$.

\item
The audit probability $\pi_t$ is $\mathcal P_t$-measurable
and satisfies $\underline\pi\le\pi_t\le1$ almost surely,
where $\underline\pi>0$ is deterministic.

\item
The dimension $r$ is fixed, and $h_t\in\R^r$ is
$\mathcal P_t$-measurable with
$\sup_{t\ge1}\|h_t\|\le C_h$ almost surely
for a deterministic constant $C_h<\infty$.
\end{enumerate}
\end{condition}

\begin{condition}
\label{cond:cov}
There is a deterministic positive definite matrix
$\Omega=\begin{pmatrix}\sigma_A^2&\gamma^\top\\ \gamma&\Omega_B\end{pmatrix}$ such that $\frac1n\sum_{t=1}^n\E\bigl[G_t(\mu_0)G_t(\mu_0)^\top\mid\mathcal F_{t-1}\bigr]\xrightarrow{p}\Omega$.
\end{condition}

Condition~\ref{cond:cov} holds under a fixed audit rule and auxiliary vector with nondegenerate
limiting moments, and under the learned rule when it converges (Corollary~\ref{cor:oracle-recovery}).
For the proposed auxiliary vector $h_t=(1,\widehat m_{t-1}(Z_t))^\top$, if $m(Z)=c$ almost surely for some constant $c$ and $\|\widehat m_n-m\|_2\xrightarrow{p}0$, the two auxiliary components become asymptotically linearly dependent.
Consequently, Condition~\ref{cond:cov} cannot hold with a positive definite limiting matrix.
In this case, the redundant auxiliary component should be dropped before applying the asymptotic empirical-likelihood results.
The limiting covariance determines the efficient control-variate coefficient $b^*=\Omega_B^{-1}\gamma$ and the corresponding variance $V_{\EL}=\sigma_A^2-\gamma^\top\Omega_B^{-1}\gamma$.

\begin{theorem}
\label{thm:adaptive-wilks}
Under Conditions~\ref{cond:basic} and~\ref{cond:cov}, the constraint set in \eqref{eq:el-ratio} is nonempty near $\mu_0$ with probability tending to one, $\sqrt n(\widehat\mu_n-\mu_0)=O_p(1)$, and
\[
\sqrt n(\widehat\mu_n-\mu_0)=\frac1{\sqrt n}\sum_{t=1}^n\{A_t(\mu_0)-(b^*)^\top B_t\}+o_p(1)\xrightarrow{d}N(0,V_{\EL}).
\]
Moreover, $-2\log\{\mathcal R_n(\mu_0)/\mathcal R_n(\widehat\mu_n)\}\xrightarrow{d}\chi^2_1.$
\end{theorem}

Although $G_t(\mu_0)$ is a martingale difference sequence whose conditional law changes with the history, the local convex-hull property and the $n^{-1/2}$-localization of $\widehat\mu_n$ are derived from the stated conditions (Section~\ref{app:subsection:A.2} of the appendix).
The linear representation shows that empirical likelihood applies the efficient coefficient $b^*$ automatically, without estimating it separately.
We next examine how the auxiliary information and the audit rule determine $V_{\EL}$.

\subsection{Efficiency Decomposition and Oracle Recovery}
\label{sec:auto:4.2}
\label{subsec:theory-efficiency}

Theorem~\ref{thm:adaptive-wilks} identifies $V_{\EL}$ as the asymptotic variance of the efficient combination $A_t(\mu_0)-(b^*)^\top B_t$.
Our procedure learns the conditional human mean $m(Z)$ for the auxiliary vector and the conditional standard deviation $\sigma(Z)$ for allocating human audits.
We now justify these choices by characterizing how $h$ and $\pi$ determine $V_{\EL}$.
We first analyze the fixed design of Section~\ref{subsec:efficiency}. Corollary~\ref{cor:oracle-recovery} subsequently connects this analysis to the limiting design of the learned procedure.
For a fixed audit rule, the following decomposition is a form of the classical AIPW variance identity \citep{robins1994estimation,tsiatis2006semiparametric}.
We state it for the empirical-likelihood variance $V_{\EL}$, which empirical likelihood attains without estimating the control-variate coefficient (Theorem~\ref{thm:adaptive-wilks}).
Stating it in this form makes explicit, through the penalty term, how judge outputs enter through the auxiliary vector.

\begin{theorem}
\label{thm:eff-decomp}
In the fixed-design setting of Section~\ref{subsec:efficiency}, suppose that $h(Z)$ is bounded,
$\underline\pi\le\pi(Z)\le1$, and $\Omega_B=\E[BB^\top]$ is positive definite. Then
\begin{equation}
V_{\EL}=V_{\mathrm{eff}}(\pi)
+\inf_{b\in\R^r}\E\left[\frac{1-\pi(Z)}{\pi(Z)}\{m(Z)-\mu_0-b^\top h(Z)\}^2\right],
\label{eq:efficiency-decomposition}
\end{equation}
where the infimum is attained at $b^*=\Omega_B^{-1}\E[BA]$.
\end{theorem}

The decomposition separates the variance into the benchmark $V_{\mathrm{eff}}(\pi)$ and a nonnegative penalty for imperfect auxiliary information.
Choosing $h(Z)=(1,m(Z))^\top$ removes this penalty.
For $h(Z)=(1,\widetilde m(Z))^\top$ satisfying the conditions of Theorem~\ref{thm:eff-decomp}, the choice $b=(-\mu_0,1)^\top$ gives
\begin{equation}
0\le V_{\EL}-V_{\mathrm{eff}}(\pi)\le\frac{1-\underline\pi}{\underline\pi}\|\widetilde m-m\|_2^2,
\label{eq:prediction-efficiency}
\end{equation}
where $\|f\|_2^2=\E[f(Z)^2]$.
This bound justifies including $\widehat m_{t-1}(Z_t)$ in $h_t$: consistent prediction eliminates the auxiliary approximation penalty under a fixed audit rule.
In particular, for given pre-audit information $Z$, the auxiliary vector affects $V_{\EL}$ only through the weighted error of projecting $m(Z)-\mu_0$ onto the span of $h(Z)$.
Consequently, an affine bias or rescaling of the judge scores is absorbed by the coefficients, and correlated judges are reweighted automatically through $\Omega_B^{-1}$.
The judges also affect efficiency through the pre-audit information $Z$ itself.
Both $m$ and $\sigma$ are defined conditionally on $Z$, so they change with the judge outputs included in $Z$.
The benchmark $V_{\mathrm{eff}}(\pi)$ changes accordingly.
The following remark shows that richer pre-audit information can only lower the oracle benchmark.

\begin{remark}
\label{rem:information}
Let $Z$ and $Z'$ be two pre-audit information vectors such that $Z$ is a measurable function of $Z'$, for example $Z'$ obtained by adding judge outputs to $Z$.
Let $V^*(W)$ denote the smallest value of $V_{\mathrm{eff}}(\pi)$ computed with information $W\in\{Z,Z'\}$, over audit rules $\pi(W)\in[\underline\pi,1]$ with $\E[\pi(W)]\le\rho$.
Then $V^*(Z')\le V^*(Z)$.
Any feasible rule $\pi(Z)$ is also available under $Z'$.
Using $V_{\mathrm{eff}}(\pi) =\Var(Y)+\E\left[\Var(Y\mid Z)\frac{1-\pi(Z)}{\pi(Z)}\right]$ and $\E[\Var(Y\mid Z')\mid Z]\le\Var(Y\mid Z)$, the same rule has no larger variance under $Z'$.
Taking infima gives $V^*(Z')\le V^*(Z)$.
Thus additional judges cannot hurt the oracle benchmark, although their realized benefit depends on how much information they add beyond $Z$.
With estimated $m$ and $\sigma$, a higher-dimensional $Z$ may increase estimation error and the realized variance in finite samples.
\end{remark}

Even after this penalty is removed, the remaining variance $V_{\mathrm{eff}}(\pi)$ depends on how audits are allocated.
Since its term $\Var\{m(Z)\}$ is unaffected by the audit rule, the remaining design problem is
\begin{equation}
\min_{\underline\pi\le\pi(Z)\le1,\ \E[\pi(Z)]\le\rho}\
\E\left[\frac{\sigma^2(Z)}{\pi(Z)}\right].
\label{eq:oracle-design-problem}
\end{equation}
The following theorem solves this problem and establishes the corresponding lower bound for history-dependent rules.

\begin{theorem}
\label{thm:oracle-annotation}
Suppose that $0<\underline\pi<\rho<1$ and $\Pp\{\sigma(Z)>0\}=1$.
There exists $\tau^*>0$ such that
\begin{equation}
\pi^*(Z)=\operatorname{clip}_{[\underline\pi,1]}\left\{\frac{\sigma(Z)}{\tau^*}\right\},
\quad
\E[\pi^*(Z)]=\rho,
\label{eq:oracle-allocation}
\end{equation}
and $\pi^*$ is the almost surely unique solution of \eqref{eq:oracle-design-problem}.
Moreover, for each $n\ge1$, every adaptive audit rule satisfying the mechanism of Section~\ref{sec:setup} and $n^{-1}\sum_{t=1}^n\E[\pi_t]\le\rho$ obeys
\begin{equation}
\frac1n\sum_{t=1}^n\E\left[\frac{\sigma^2(Z_t)}{\pi_t}\right] \ge\E\left[\frac{\sigma^2(Z)}{\pi^*(Z)}\right].
\label{eq:adaptive-oracle-bound}
\end{equation}
\end{theorem}

The first part of Theorem~\ref{thm:oracle-annotation} is a constrained Neyman allocation \citep{neyman1934}.
The optimal rule assigns higher audit probabilities to instances with greater conditional human uncertainty, subject to the probability bounds.
This motivates learning $\sigma(Z)$ rather than relying on the uncertainty or disagreement of the judges.
When the judges are highly informative, $\sigma(Z)$ may be close to zero on a large set of instances.
The floor $\underline\pi$ then becomes active there, which keeps the inverse-probability weights bounded and ensures that every instance retains a positive chance of being audited.
When clipping is inactive for $\pi^*$, the gain over uniform auditing $\pi_{\mathrm{unif}}(Z)=\rho$ is
\begin{equation}
  V_{\mathrm{eff}}(\pi_{\mathrm{unif}}) - V_{\mathrm{oracle}}
  = \frac{\operatorname{Var}\{\sigma(Z)\}}{\rho},
  \quad V_{\mathrm{oracle}} = V_{\mathrm{eff}}(\pi^*),
  \label{eq:uniform-oracle-gap} 
\end{equation}
so greater heterogeneity in conditional uncertainty offers a larger potential gain.
The adaptive bound~\eqref{eq:adaptive-oracle-bound} follows from a pointwise Lagrangian inequality evaluated at $(Z_t,\pi_t)$, together with $\E[\pi^*(Z_t)]=\rho$, which uses only that $Z_t$ has the same distribution as $Z$.
Independence of $Z_t$ from $\mathcal F_{t-1}$ is needed only for the conditional version used in Proposition~\ref{prop:adaptive-variance-bound}.
The bound~\eqref{eq:adaptive-oracle-bound} shows that, if $\sigma$ were known, history dependence could not improve this variance criterion under the same expected budget.
Under the long-run expected budget of Section~\ref{sec:setup}, the corresponding bound holds for the lower limit of the average variance criterion.
The next result extends this bound, within the augmented estimating-equation class of Conditions~\ref{cond:basic}--\ref{cond:cov}, from the audit-dependent term to the asymptotic variance $V_{\mathrm{EL}}$ in Theorem~\ref{thm:adaptive-wilks}.

\begin{proposition}
\label{prop:adaptive-variance-bound}
Suppose that Conditions~\ref{cond:basic} and~\ref{cond:cov} hold, that $0<\underline\pi<\rho<1$ and $\Pp\{\sigma(Z)>0\}=1$, and that $\limsup_{n\to\infty}\frac1n\sum_{t=1}^n\E[\pi_t]\le\rho.$
Then
\[
V_{\EL}\ge V_{\mathrm{oracle}}=\Var\{m(Z)\}+\E\left[\frac{\sigma^2(Z)}{\pi^*(Z)}\right].
\]
\end{proposition}

The role of adaptation is to learn the oracle allocation from previously acquired labels.

\begin{corollary}
\label{cor:oracle-recovery}
Suppose that Condition~\ref{cond:basic} holds and that the audit probabilities and auxiliary vectors are given by \eqref{eq:learned-allocation}.
Assume that the equation $\E[\clip_{[\underline\pi,1]}\{\sigma(Z)/\tau\}]=\rho$ has a unique solution $\tau^*$, and that $\tau^*\in\operatorname{int}(T)$.
Write $\widehat\pi_n(z)=\clip_{[\underline\pi,1]}\{\widehat\sigma_n(z)/\widehat\tau_n\}$ for the audit rule learned from the first $n$ observations.
Assume also that $\|\widehat m_n-m\|_2\xrightarrow{p}0$, $\|\widehat\sigma_n-\sigma\|_2\xrightarrow{p}0$ and $\widehat\tau_n\xrightarrow{p}\tau^*$, and that the covariance matrix $\Omega^*$ of $G(\mu_0)$ under the fixed design $(\pi^*,(1,m)^\top)$ is positive definite.
Then $\|\widehat\pi_n-\pi^*\|_2\xrightarrow{p}0$. 
Consequently, Condition~\ref{cond:cov} holds with $\Omega=\Omega^*$, and hence $V_{\EL}=V_{\mathrm{oracle}}$.
Moreover, $\frac1n\sum_{t=1}^n\E[\pi_t]\to\rho$ and $\frac1n\sum_{t=1}^n\pi_t\xrightarrow{p}\rho$.
\end{corollary}

Together with Proposition~\ref{prop:adaptive-variance-bound}, Corollary~\ref{cor:oracle-recovery} shows that the learned design attains the smallest asymptotic variance $V_{\EL}$ among history-dependent designs satisfying Conditions~\ref{cond:basic} and~\ref{cond:cov}.
Both $\pi^*$ and $V_{\mathrm{oracle}}$ depend on $\underline\pi$, so the optimality statement is relative to this floor.
Sufficient conditions for prediction consistency and threshold consistency are given in Propositions~\ref{prop:ipw-consistency} and~\ref{prop:threshold-consistency} in Section~\ref{app:subsection:A.7} of the appendix.
Neither sample splitting nor an independent calibration sample is required: the nuisance functions are fitted on past observations and evaluated before the current label is revealed, and the threshold is calibrated using only the unlabeled stream $Z_1,\ldots,Z_n$.
Section~\ref{subsec:theory-cs} shows that the same variance governs evidence accumulation in the confidence sequence.

\subsection{Predictable Empirical-Likelihood $e$-Processes}
\label{sec:auto:4.3}
\label{subsec:theory-cs}

The inferential question is whether multiple moment conditions can be combined using only past observations without losing their first-order profile information.
We first establish exact anytime validity, which requires neither covariance stabilization nor consistent prediction.
We then study two predictable empirical-likelihood $e$-processes: a vector version that mixes the empirical-likelihood dual over all $q$ multipliers (Theorem~\ref{thm:vector-bridge}), and the profiled version of Section~\ref{subsec:AAELI}, which bets along a predictable estimate of the efficient direction (Theorem~\ref{thm:profile-bridge}).
The vector version recovers the raw empirical-likelihood statistic with a $q\log n$ correction, whereas the profiled version retains the profile statistic with only a $\log n$ correction.
For the profiled continuous-mixture construction, the efficient variance $V_{\EL}$ also governs local evidence accumulation.

\begin{theorem}
\label{thm:anytime-validity}
Suppose that Condition~\ref{cond:basic} holds.
Let $\widehat a_{t-1}(\mu)$ be $\mathcal F_{t-1}$-measurable directions satisfying $\|\widehat a_{t-1}(\mu)\|\le C_a$ for every $t\ge1$ and $\mu\in\Theta$.
Let $\varpi$ be a prespecified probability distribution supported on the safe interval $I_\eta$, and define $E_n(\mu)$ by \eqref{eq:profile-eprocess}.
Then $E_n(\mu_0)$ is a nonnegative martingale with $E_0(\mu_0)=1$, and $\Pp\left(\mu_0\in\mathcal C_n\text{ for all }n\ge1\right)\ge1-\alpha.$
Consequently, $\Pp(\mu_0\in\mathcal C_N)\ge1-\alpha$ for every almost surely finite stopping time $N$.
\end{theorem}

Because coverage holds simultaneously over all $n$, the guarantee $\Pp(\mu_0\in\mathcal C_N)\ge1-\alpha$ extends to any almost surely finite random time $N$, even one that is not a stopping time, such as a stopping rule revised after inspecting the data.

Before profiling, consider mixing the empirical-likelihood dual objective directly over all $q$ multipliers.
Fix $\eta\in(0,1)$ and let $\Lambda_\eta=\{\lambda\in\R^q:\|\lambda\|\le\eta/C_G\}$, so that $1+\lambda^\top G_t(\mu)\ge1-\eta>0$ for all $t$, $\mu\in\Theta$, and $\lambda\in\Lambda_\eta$.
For a prespecified probability measure $\nu$ on $\Lambda_\eta$, define
\begin{equation}
E_n^{\mathrm{vec}}(\mu)=\int_{\Lambda_\eta}\prod_{t=1}^n\{1+\lambda^\top G_t(\mu)\}\,\nu(d\lambda),
\quad
T_n^{\mathrm{vec}}(\mu)=2\sup_{\lambda\in\Lambda_\eta}\sum_{t=1}^n\log\{1+\lambda^\top G_t(\mu)\}.
\label{eq:vector-eprocess}
\end{equation}
The supremum in $T_n^{\mathrm{vec}}$ is the empirical-likelihood dual problem restricted to $\Lambda_\eta$.

\begin{theorem}
\label{thm:vector-bridge}
Under Condition~\ref{cond:basic}, $E_n^{\mathrm{vec}}(\mu_0)$ is a unit-mean nonnegative martingale, and
\[
E_n^{\mathrm{vec}}(\mu)=\prod_{t=1}^n\bigl\{1+\bar\lambda_{t-1}(\mu)^\top G_t(\mu)\bigr\},
\quad
\bar\lambda_{t-1}(\mu)=\frac{\int_{\Lambda_\eta}\lambda\prod_{s<t}\{1+\lambda^\top G_s(\mu)\}\,\nu(d\lambda)}{E_{t-1}^{\mathrm{vec}}(\mu)},
\]
so the mixture is a predictable empirical-likelihood bettor with $\mathcal F_{t-1}$-measurable multiplier $\bar\lambda_{t-1}(\mu)\in\Lambda_\eta$.
Consequently, $\mathcal C_n^{\mathrm{vec}}=\{\mu\in\Theta:E_n^{\mathrm{vec}}(\mu)<1/\alpha\}$ satisfies $\Pp(\mu_0\in\mathcal C_n^{\mathrm{vec}}\text{ for all }n\ge1)\ge1-\alpha$.
If Condition~\ref{cond:cov} holds and $\nu$ has a Lebesgue density $\varphi_\nu$ on $\Lambda_\eta$ that is continuous and positive at zero, then $T_n^{\mathrm{vec}}(\mu_0)=-2\log\mathcal R_n(\mu_0)+o_p(1)$ and
\begin{equation}
\begin{aligned}
2\log E_n^{\mathrm{vec}}(\mu_0)+q\log n-C_{\mathrm{vec}}&=-2\log\mathcal R_n(\mu_0)+o_p(1),\\
C_{\mathrm{vec}}&=2\log \varphi_\nu(0)+q\log(2\uppi)-\log\det\Omega.
\end{aligned}
\label{eq:vector-bridge}
\end{equation}
\end{theorem}

By Theorem~\ref{thm:adaptive-wilks} and Lemma~\ref{lem:local-el} of the appendix, $-2\log\mathcal R_n(\mu_0)=\mathbb G_n^\top\Omega^{-1}\mathbb G_n+o_p(1)\xrightarrow{d}\chi_q^2$.
Thus the vector $e$-process accumulates evidence against the full moment system, including $r$ auxiliary equations whose expectations do not depend on $\mu$.

We next impose additional conditions to characterize the local information structure of the betting process.
At $\mu_0$, the $e$-process bets along $\widehat a_{t-1}(\mu_0)$, a predictable estimate of $a^*=\Omega^{-1}D/\{D^\top\Omega^{-1}D\}^{1/2}$ with $D=-e_1$, for which $(a^*)^\top G_t(\mu_0)=-\{A_t(\mu_0)-(b^*)^\top B_t\}/V_{\EL}^{1/2}$ is the standardized efficient combination.
Write $T_{n,\mathrm{prof}}(\mu)=-2\log\frac{\mathcal R_n(\mu)}{\mathcal R_n(\widehat\mu_n)}$ for the profile empirical-likelihood statistic, and recall the scalar empirical likelihood ratio $\widetilde{\mathcal R}_n(\mu)$ defined in Section~\ref{subsec:AAELI}.

\begin{theorem}
\label{thm:profile-bridge}
Suppose that Conditions~\ref{cond:basic} and~\ref{cond:cov} hold, that the directions $\widehat a_{t-1}(\mu)$ are constructed by \eqref{eq:predictable-direction} with $\varepsilon_t\downarrow0$, and that $C_a>\|a^*\|$.
Then
\begin{equation}
-2\log\widetilde{\mathcal R}_n(\mu_0)=T_{n,\mathrm{prof}}(\mu_0)+o_p(1).
\label{eq:profile-equivalence}
\end{equation}
If $\varpi$ has a density $\varphi_\varpi$ on $I_\eta$ that is continuous and positive at zero, then
\begin{equation}
2\log E_n(\mu_0)+\log n-C_{\mathrm{prof}}=T_{n,\mathrm{prof}}(\mu_0)+o_p(1),
\label{eq:profile-bridge}
\end{equation}
with $C_{\mathrm{prof}}=2\log\varphi_\varpi(0)+\log(2\uppi)$.
\end{theorem}

The first expansion shows that learning the direction from past observations does not alter the first-order profile statistic at $\mu_0$.
Since $\|a^*\|^2=(1+\|b^*\|^2)/V_{\EL}$ and $V_{\EL}\ge V_{\mathrm{eff}}(\pi)\ge\Var(Y)$ (Theorem~\ref{thm:eff-decomp} and Proposition~\ref{prop:adaptive-variance-bound}), we have $\|a^*\|^2\le(1+\|b^*\|^2)/\Var(Y)$.
For $h_t=(1,\widehat m_{t-1}(Z_t))^\top$ with $\widehat m_{t-1}$ close to $m$, $b^*\approx(-\mu_0,1)^\top$, so the condition is most demanding when $\Var(Y)$ is small.
Radial truncation preserves the direction of the estimated projection but changes its normalization and may affect finite-sample performance.
The second identifies the $\log n$ penalty induced by smooth scalar mixing.

\begin{remark}
\label{rem:profiling-dimension}
Theorems~\ref{thm:vector-bridge} and~\ref{thm:profile-bridge} give $2\log E_n^{\mathrm{vec}}(\mu_0)+q\log n=-2\log\mathcal R_n(\mu_0)+O_p(1)$ and $2\log E_n(\mu_0)+\log n=T_{n,\mathrm{prof}}(\mu_0)+O_p(1).$
With $\mathcal I=D^\top\Omega^{-1}D$, let $P=\Omega^{-1}-\Omega^{-1}DD^\top\Omega^{-1}/\mathcal I$, so that $PD=0$ and $(\mathbb G_n+Du)^\top\Omega^{-1}(\mathbb G_n+Du)
=\{(a^*)^\top(\mathbb G_n+Du)\}^2+\mathbb G_n^\top P\mathbb G_n .$
The second term does not depend on the local parameter $u$, although the auxiliary moments contribute to determining the efficient direction $a^*$ and the information $\mathcal I$.
The vector $e$-process mixes over all $q$ directions and incurs a smooth-mixture correction of $q\log n$, whereas the profiled $e$-process bets only along the predictable estimate of $a^*$ and incurs a correction of $\log n$.
At the level of the local quadratic approximation, empirical-likelihood profiling corresponds to reducing the betting mixture to the efficient scalar direction, which is why the confidence sequence \eqref{eq:confidence-sequence} is built from the profiled construction.
\end{remark}

\begin{remark}
\label{rem:projection-value}
For a fixed direction $a$ with $a^\top D\neq0$, define $V(a)=\frac{a^\top\Omega a}{(a^\top D)^2}$.
This is the variance-to-sensitivity ratio of the projected moment equation.
The Cauchy--Schwarz inequality gives $V(a)\ge\{D^\top\Omega^{-1}D\}^{-1}=V_{\EL}$ with equality if and only if $a$ is proportional to $\Omega^{-1}D$.
Indeed, $(a^\top D)^2=\{(\Omega^{1/2}a)^\top(\Omega^{-1/2}D)\}^2\le(a^\top\Omega a)(D^\top\Omega^{-1}D).$
Theorem~\ref{thm:profile-bridge} shows that the optimal projection can be learned predictably while retaining the first-order profile statistic.
No prescribed convergence rate for the direction estimator is required.
The gain is the removal of information loss from an inefficient prescribed linear combination, while maintaining exact anytime validity throughout learning.
\end{remark}

The same predictable construction also characterizes evidence against deterministic candidate means approaching $\mu_0$.
The required score approximations are derived in Section~\ref{app:subsection:A.11} of the appendix from the conditions of Theorem~\ref{thm:profile-bridge}.

\begin{corollary}
\label{cor:alternatives}
Suppose that the conditions of Theorem~\ref{thm:profile-bridge} hold and $\varpi$ has a density $\varphi_\varpi$ on $I_\eta$, continuous and positive at zero.
Let $(\mu_n)\subseteq\Theta$ be a deterministic sequence.
\begin{enumerate}
\item If $\mu_n=\mu_0+u/\sqrt n$ for a fixed $u\in\R$, then $2\log E_n(\mu_n)+\log n-C_{\mathrm{prof}}\xrightarrow{d}\chi_1^2(u^2/V_{\EL}).$
\item If $\mu_n=\mu_0+\Delta_n$ with $\Delta_n\to0$ and $n\Delta_n^2/\log n\to\infty$, then $2\log E_n(\mu_n)=\frac{n\Delta_n^2}{V_{\EL}}\{1+o_p(1)\}.$
\end{enumerate}
\end{corollary}

Under the learned design of Corollary~\ref{cor:oracle-recovery}, these expansions hold with $V_{\EL}=V_{\mathrm{oracle}}$.
In computation we replace the continuous mixture by a finite mixture of predictable betting strategies.
Each component keeps every betting factor positive, so the finite mixture remains an exactly valid $e$-process.
The next result shows that no valid $e$-value can accumulate evidence faster than the rate in Corollary~\ref{cor:alternatives}, whatever audit rule or estimating equation it uses.

\begin{proposition}
\label{prop:evidence-bound}
Suppose that Condition~\ref{cond:basic}~(a)--(c) and the assumptions of Theorem~\ref{thm:oracle-annotation} hold under $P_0$.
Write $V=V_{\mathrm{oracle}}$ and define, for sufficiently small $|\delta|$, the local submodel $\frac{dP_\delta}{dP_0}(z,y)=1+\frac{\delta}{V}\left\{m(z)-\mu_0+\frac{y-m(z)}{\pi^*(z)}\right\}.$
Then $\mu(P_\delta)=\mu_0+\delta$.
Use the same audit and learning rules under all $P_\delta$, with the same independent initial information.
Assume $\limsup_{n\to\infty}\frac1n\sum_{t=1}^n\E_0[\pi_t]\le\rho.$
Let $\mathcal E_n(\mu)$ be nonnegative, $\mathcal F_n$-measurable statistics.
Assume that there exists $\delta_0>0$ such that $\E_{P_\delta}[\mathcal E_n(\mu_0+\delta)]\le1$ for every $n\ge1$ and every $|\delta|\le\delta_0$ with $\mu_0+\delta\in\Theta$.
Then, for every deterministic sequence $\delta_n\ne0$ with $\delta_n\to0$ and $\mu_0+\delta_n\in\Theta$ eventually:
(i) $\limsup_{n\to\infty}\frac{\E_0[\log\mathcal E_n(\mu_0+\delta_n)]}{n\delta_n^2/2}\le\frac1V$;
(ii) if, in addition, $n\delta_n^2\to\infty$, then, for every $\epsilon>0$, $\limsup_{n\to\infty}P_0\left\{\frac{2\log\mathcal E_n(\mu_0+\delta_n)}{n\delta_n^2}\ge\frac{1+\epsilon}{V}\right\}\le\frac1{1+\epsilon}$.
\end{proposition}

By part~(ii), if the normalized log-evidence $2\log\mathcal E_n(\mu_0+\delta_n)/(n\delta_n^2)$ converges in probability to a constant, that constant is at most $1/V_{\mathrm{oracle}}$.
Thus $V_{\mathrm{oracle}}$ is an information bound beyond the augmented estimating-equation class.
The AAELI $e$-process is valid under every $P_\delta$ (Theorem~\ref{thm:anytime-validity}), and under the conditions of Corollaries~\ref{cor:oracle-recovery} and~\ref{cor:alternatives} it satisfies $2\log E_n(\mu_0+\delta_n)/(n\delta_n^2)\xrightarrow{p}1/V_{\mathrm{oracle}}$ whenever $\delta_n\to0$ and $n\delta_n^2/\log n\to\infty$.
Hence the continuous-mixture AAELI construction attains the bound in Proposition~\ref{prop:evidence-bound}.

\section{Synthetic Studies} 
\label{sec:auto:5}
\label{sec:synthetic}
\subsection{Data Generation Process}
\label{sec:auto:5.1}
We consider $K=3$ LLM judges scoring a stream of $n=10{,}000$ items, with a target human audit rate $\rho=0.2$, corresponding to a nominal target of $n\rho=2{,}000$ human labels.
For an adaptive audit rule, the finite-sample expected number of labels is $\E[\sum_{t=1}^n R_t]=\sum_{t=1}^n\E[\pi_t]$, which need not equal $n\rho$.
To evaluate adaptive allocation of human audits, we construct a population with heterogeneous outcome uncertainty.
For each item, we generate observed features $X\sim N(0,I_5)$ and an unobserved latent factor $H\sim N(0,1)$.
The human outcome follows $Y\mid X,H\sim\mathrm{Bernoulli}\!\left(\operatorname{logit}^{-1}\{\zeta(X,H)\}\right)$ and $\zeta(X,H)=\kappa(X_5)(0.8X_1-0.6X_2+0.4X_3+H)$.
We set $\kappa(X_5)=0.3$ when $X_5\le\Phi^{-1}(0.2)$ and $\kappa(X_5)=4$ otherwise.
The smaller coefficient pushes outcome probabilities toward $0.5$, so 20\% of the population has higher conditional uncertainty.
For each judge $k=1,\ldots,K$, we generate the logit $L_k=d_k^\top X+c_kH+U+e_k$ and the prediction $J_k=\operatorname{logit}^{-1}(L_k)$, where $U\sim N(0,0.3^2)$ is noise shared across judges and $e_k\overset{\mathrm{i.i.d.}}{\sim}N(0,0.7^2)$ is judge-specific noise.
The judge-specific latent-factor loadings are $(c_1,c_2,c_3)=(0.6,0.8,1.0)$, and the feature coefficients are $d_1=(0.30,-0.20,0.10,0.05,0)^\top$, $d_2=(0.20,-0.10,0.25,0,0.10)^\top$, and $d_3=(0.35,-0.15,0,0.10,0)^\top$, so that the judges differ in how strongly they track the human-relevant latent factor and which features they emphasize.
The pre-audit information consists of the observed features and LLM predictions, $Z=(X^\top,J_1,\ldots,J_K)^\top$.
Our goal is to estimate the population mean human outcome $\mu=\mathbb{E}[Y]$.

\subsection{Baselines} 
\label{sec:auto:5.2}
We consider the following baselines: 
PPI \citep{waudby2024time,zrnic2024active} audits items uniformly at rate $\rho$ and builds anytime-valid PPI confidence sequences.
ASI \citep{zrnic2024active} adaptively allocates human labels based on predictive uncertainty and estimates the target mean using an inverse-probability-weighted AIPW estimator.
RASI \citep{li2026robust} robustifies ASI against predictor misspecification by interpolating between uncertainty-based and uniform sampling.
FTRL \citep{sfyraki2026revisiting} selects a history-dependent labeling probability
via follow-the-regularized-leader under a budget constraint.
Since RASI and FTRL only provide fixed-time confidence intervals, we extend them to
anytime-valid confidence sequences by applying the betting construction \citep{zrnic2024active}.
All methods that learn from audited labels use a random forest \citep{breiman2001random} as the prediction model.
Implementation details are given in Section~\ref{app:section:B} of the appendix.

\subsection{Main Results}
\label{sec:auto:5.3}

\begin{figure*}[htbp]
\centering
\includegraphics[width=0.7\textwidth]{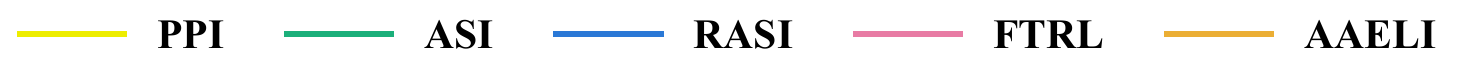}
\subfloat[Confidence Sequences]{%
    \includegraphics[width=0.33\textwidth]{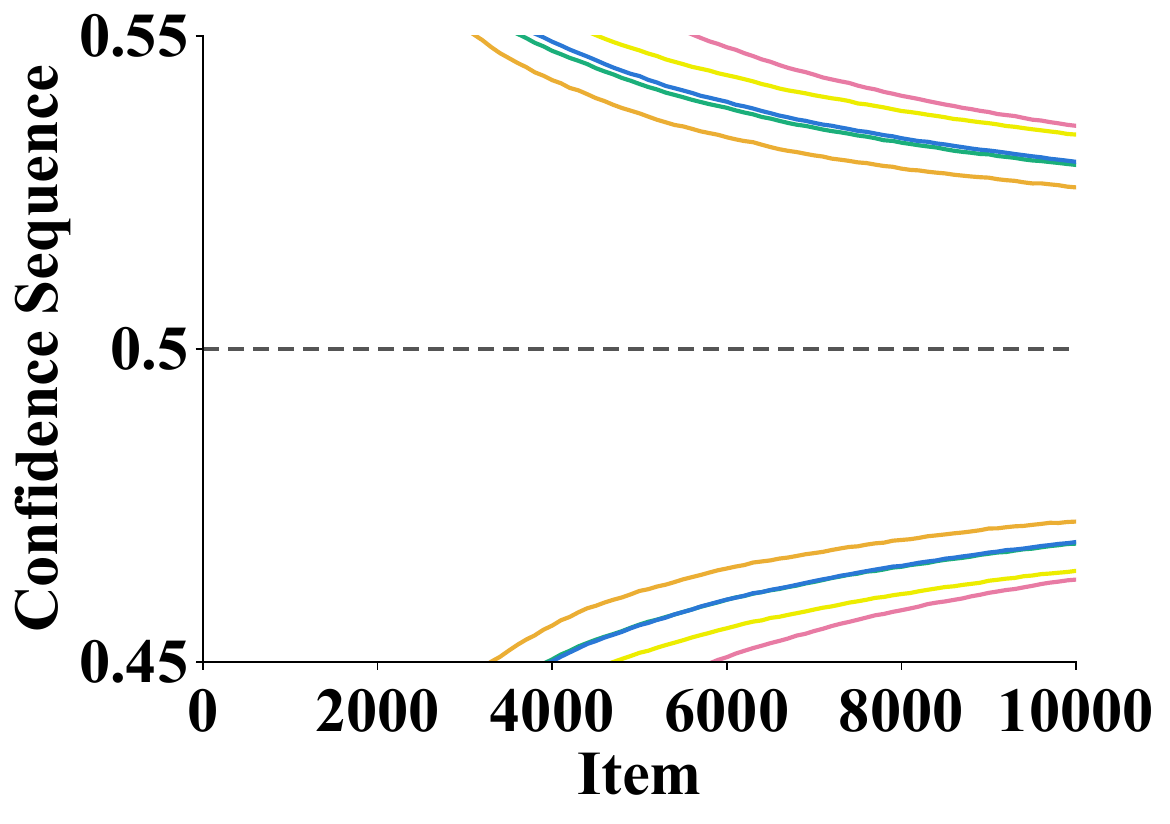}
}
\subfloat[Anytime Coverage]{%
    \includegraphics[width=0.33\textwidth]{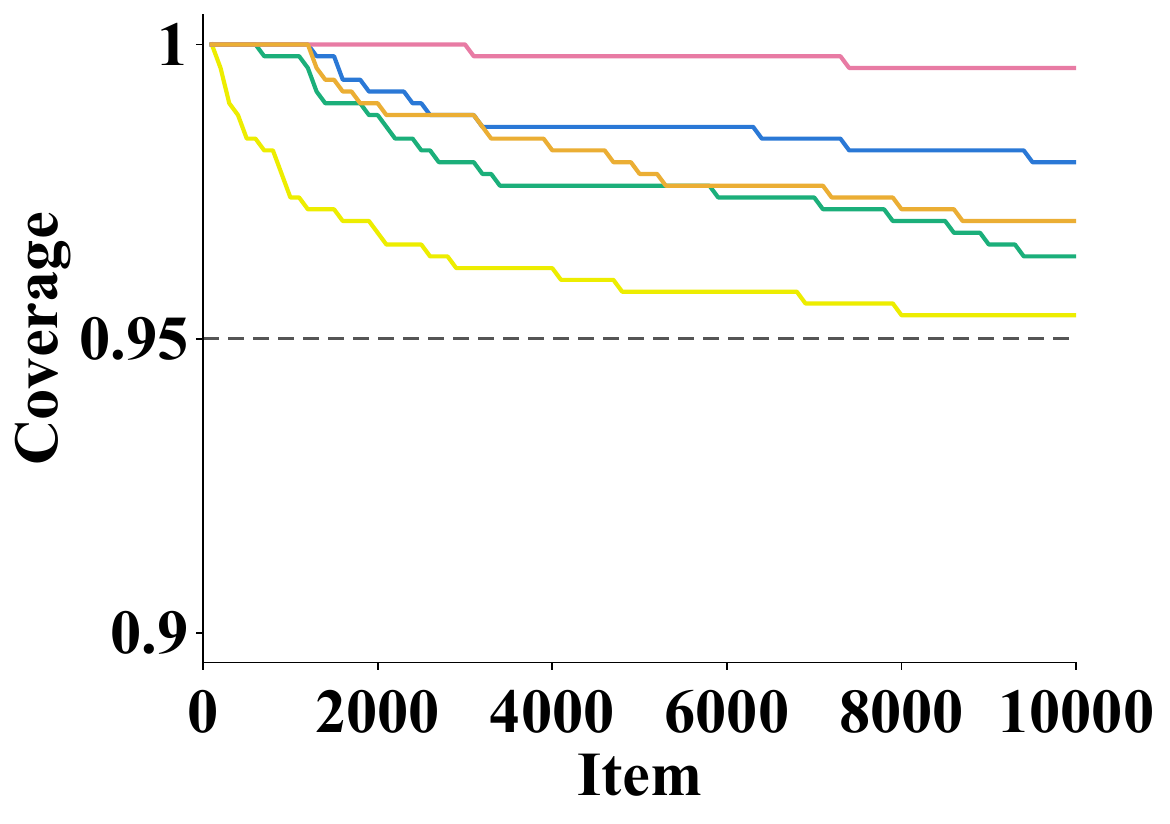}
}
\subfloat[Width]{%
    \includegraphics[width=0.33\textwidth]{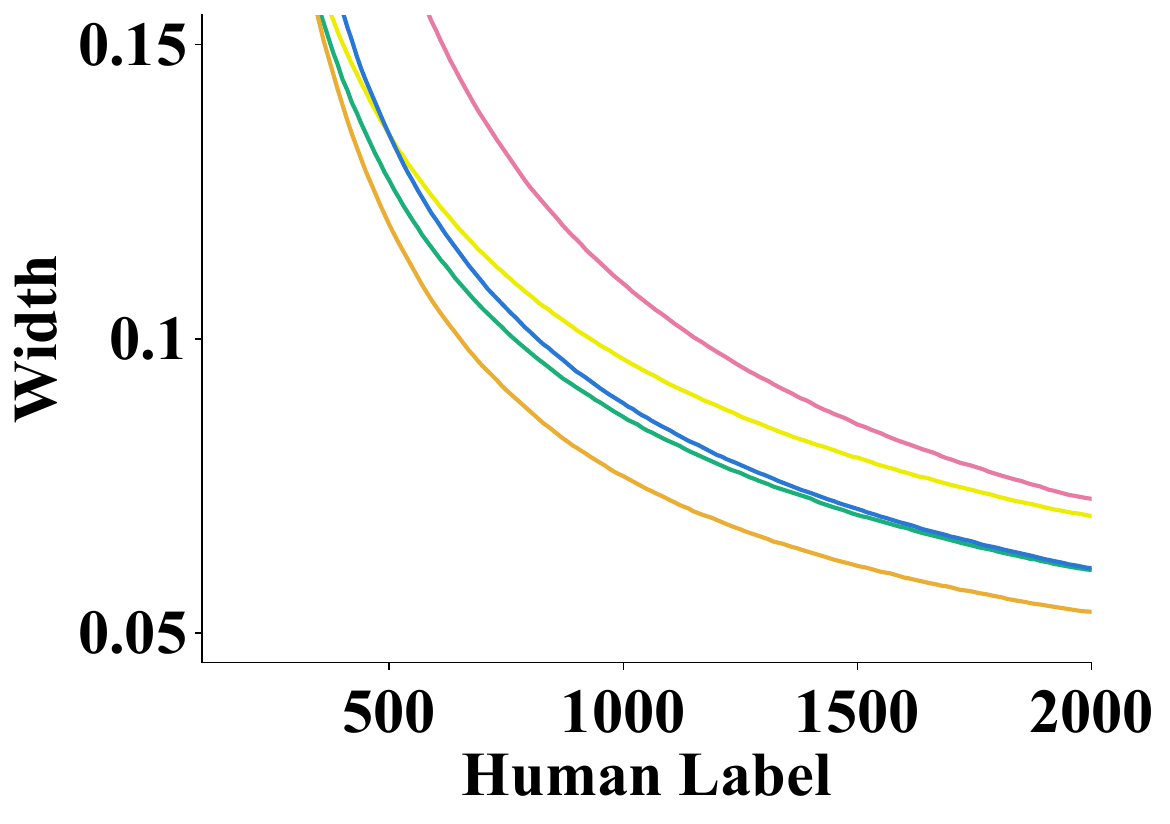}
}
\caption{Anytime-valid inference with $n=10{,}000$ items and budget $\rho=0.2$, averaged over 500 replications.
Left: Average confidence sequence bounds.
The dotted line marks the true mean $\mu=0.5$.
Middle: Percentage of replications whose confidence sequence covers the true mean at every item up to the current one.
Right: Average interval width vs.\ the number of human labels.}
\label{figure:1}
\end{figure*}

\textit{AAELI Maintains Anytime-Valid Coverage with Fewer Human Labels.}
Figure~\ref{figure:1} shows the average confidence-sequence bounds (left) and anytime coverage (middle) over the item stream, and the average interval width against the number of human labels (right).
The results show that the confidence sequences of all methods contract symmetrically around the true mean $\mu=0.5$, with AAELI yielding the tightest bounds throughout the stream (e.g., $[0.472, 0.526]$ at the final item versus $[0.469, 0.529]$ for ASI).
The middle panel reports the empirical anytime coverage over the stream.
The empirical anytime coverage of all methods, including ours, remains above the nominal $95\%$ level (dashed line) throughout the evaluated horizon.
The right panel reports the average interval width as a function of the number of human labels.
Our method attains the narrowest confidence sequences at every budget, and thus requires the fewest human labels to reach any given width.
For example, reaching a width of $0.1$ requires on average 620 human labels with our method, compared with 734 for ASI, 781 for RASI, 880 for PPI, and 1,121 for FTRL, a saving of 16\%--45\%.
Relative to the strongest baseline, ASI, the savings grow as the target width shrinks, from 18\% at width $0.09$ to 21\%, 23\%, and 24\% at widths $0.08$, $0.07$, and $0.06$, respectively.

\textit{Fixed-Budget Comparison.}
We further compare the methods at a fixed budget, where each method processes the entire stream of $n=10{,}000$ items and constructs a fixed-sample $95\%$ confidence interval once at the end.
This removes the anytime-validity correction and isolates the gains from the audit rule and the auxiliary information.
We also include balanced active inference (BAI) \citep{chen2026balanced}, which applies only to fixed-budget inference and is therefore omitted from the anytime valid comparisons.
Table~\ref{table1} reports the results for $\rho\in\{0.05,0.10,0.15,0.20\}$.
All methods attain coverage close to the nominal level.
AAELI yields the narrowest intervals at every budget: at $\rho=0.2$, its average width is $0.034$, compared with $0.036$ for ASI and RASI, $0.037$ for PPI and FTRL, and $0.040$ for BAI.
Relative to the strongest baseline, the width reduction ranges from $3.9\%$ to $5.6\%$ across budgets.
These results show that the advantage of AAELI is not specific to the anytime-valid construction: the learned audit rule and the empirical-likelihood combination of the augmented moments also improve fixed-sample efficiency.

\begin{table}[htbp]
    \centering
\caption{Fixed-budget inference on synthetic data with $n=10{,}000$ items and human audit budgets $\rho\in\{0.05,0.10,0.15,0.20\}$, averaged over 500 replications. Each method uses its own audit rule and constructs a fixed-sample $95\%$ confidence interval once, at $n=10{,}000$. Coverage is the proportion of replications whose interval contains $\mu_0=0.5$, and Width is the average interval width. Subscripts give Monte Carlo standard errors. The smallest width for each budget is shown in bold.}
    \resizebox{\textwidth}{!}{%
    \begin{tabular}{cccccccc} \toprule
         $\rho$ & Metric & PPI & ASI & RASI & FTRL & BAI & AAELI \\ \midrule
         \multirow{2}{*}{0.05} & Coverage & 0.946$_{\pm 0.010}$ & 0.942$_{\pm 0.010}$ & 0.946$_{\pm 0.010}$ & 0.948$_{\pm 0.010}$ & 0.950$_{\pm 0.010}$ & 0.946$_{\pm 0.011}$\\
                               & Width $\downarrow$  & 0.074$_{\pm 0.002}$ & 0.074$_{\pm 0.003}$ & 0.076$_{\pm 0.003}$ & 0.077$_{\pm 0.002}$ & 0.080$_{\pm 0.004}$ & \textbf{0.070}$_{\pm 0.003}$ \\ \midrule
         \multirow{2}{*}{0.1}  & Coverage& 0.952$_{\pm 0.010}$ & 0.954$_{\pm 0.009}$ & 0.942$_{\pm 0.010}$ & 0.940$_{\pm 0.011}$ & 0.940$_{\pm 0.011}$ & 0.956$_{\pm 0.009}$ \\
                               & Width $\downarrow$  & 0.052$_{\pm 0.001}$ & 0.051$_{\pm 0.001}$ & 0.051$_{\pm 0.001}$ & 0.053$_{\pm 0.001}$ & 0.056$_{\pm 0.002}$ & \textbf{0.049}$_{\pm 0.002}$ \\ \midrule
         \multirow{2}{*}{0.15} & Coverage & 0.938$_{\pm 0.011}$ & 0.944$_{\pm 0.010}$ & 0.942$_{\pm 0.010}$ & 0.946$_{\pm 0.010}$ & 0.950$_{\pm 0.010}$ & 0.948$_{\pm 0.011}$\\
                               & Width $\downarrow$  & 0.042$_{\pm 0.001}$ & 0.041$_{\pm 0.001}$ & 0.041$_{\pm 0.001}$ & 0.043$_{\pm 0.001}$ & 0.045$_{\pm 0.002}$ & \textbf{0.039}$_{\pm 0.001}$ \\ \midrule
         \multirow{2}{*}{0.2}  & Coverage & 0.952$_{\pm 0.010}$ & 0.956$_{\pm 0.009}$ & 0.952$_{\pm 0.010}$ & 0.962$_{\pm 0.009}$ & 0.960$_{\pm 0.009}$ & 0.954$_{\pm 0.009}$ \\
                               & Width $\downarrow$  & 0.037$_{\pm 0.001}$ & 0.036$_{\pm 0.001}$ & 0.036$_{\pm 0.001}$ & 0.037$_{\pm 0.001}$ & 0.040$_{\pm 0.001}$ & \textbf{0.034}$_{\pm 0.001}$ \\ \bottomrule
    \end{tabular}}
    \label{table1}
\end{table}

\begin{figure*}[htbp]
\centering
\includegraphics[width=0.95\textwidth]{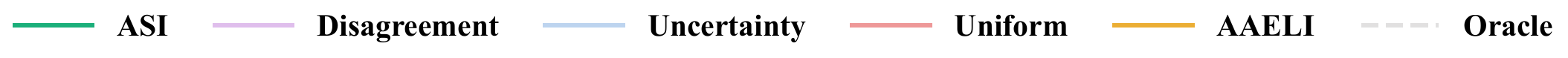}
\subfloat[Confidence Sequences]{%
    \includegraphics[width=0.33\textwidth]{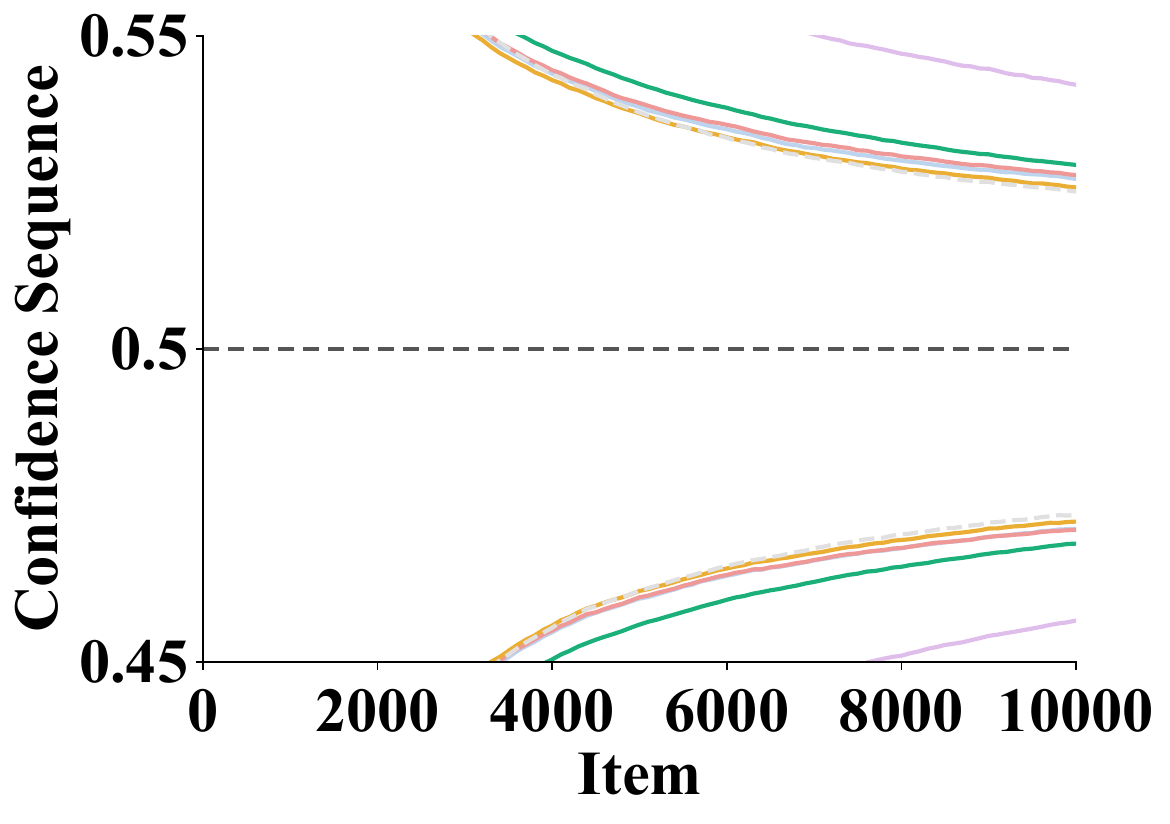}
}
\subfloat[Anytime Coverage]{%
    \includegraphics[width=0.33\textwidth]{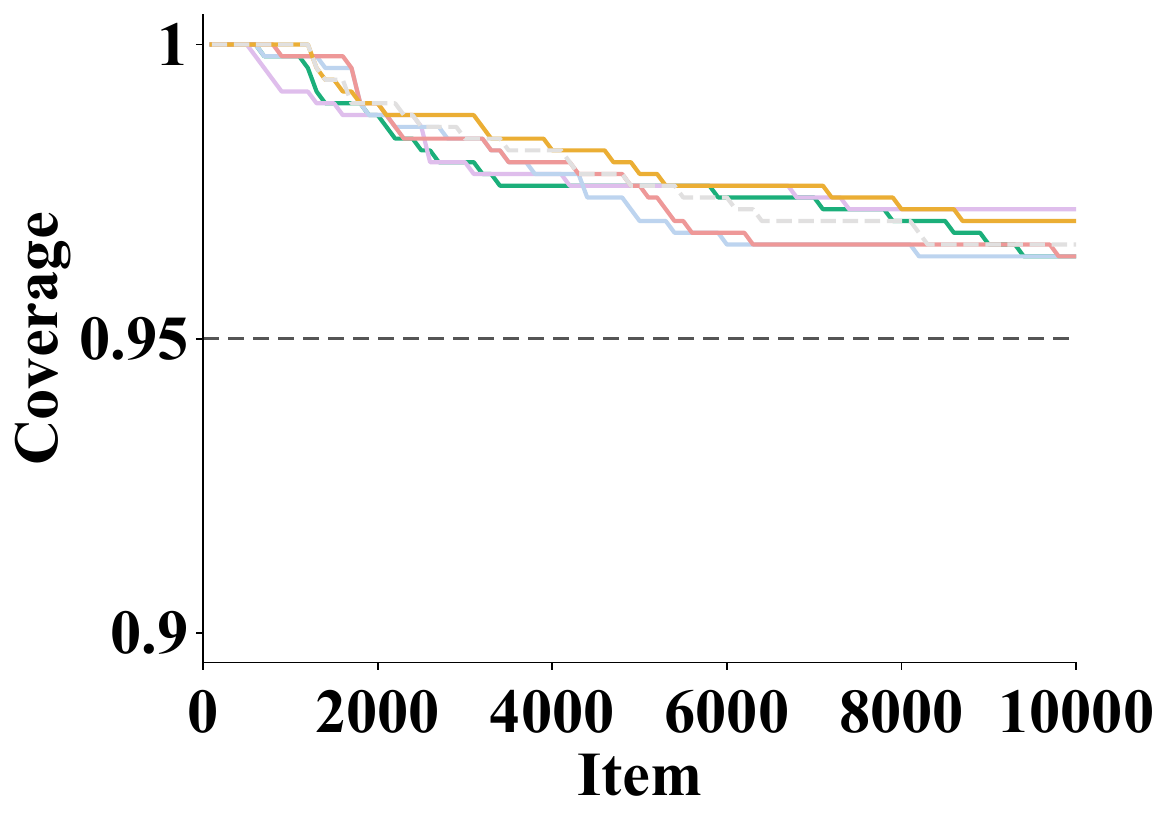}
}
\subfloat[Width]{%
    \includegraphics[width=0.33\textwidth]{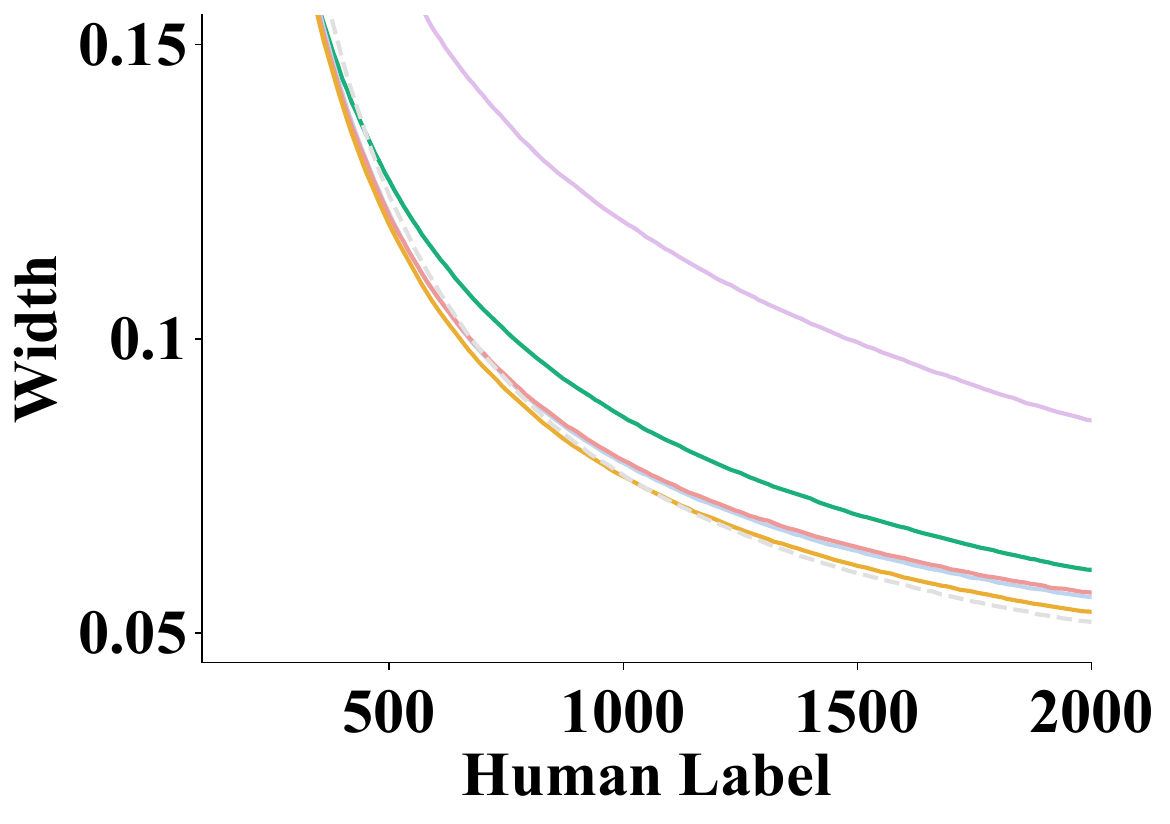}
}
\caption{Ablation on the human audit policy. 
Uniform, Uncertainty, Disagreement, and Oracle use the same AAELI inference procedure as AAELI, but replace its learned audit rule with uniform auditing, auditing by the uncertainty of the LLM judges, auditing by the disagreement among the judges, and the oracle rule with known $\sigma(Z)$, respectively.
ASI is the strongest baseline.}
\label{figure:a1}
\end{figure*}

\textit{Ablation Study on Adaptive Auditing.}
We examine the effect of adaptive auditing by fixing the inference procedure to AAELI and replacing its learned audit rule with four alternatives.
Let $\bar J_t=K^{-1}\sum_{k=1}^KJ_{tk}$ denote the average judge score.
Uniform audits every instance with probability $\pi_t=\rho$.
Uncertainty uses the score $s_t=\bar J_t(1-\bar J_t)$, which is largest when the judges are on average undecided.
Disagreement uses the score $s_t=K^{-1}\sum_{k=1}^K(J_{tk}-\bar J_t)^2$, the variance of the judge scores across judges.
For both rules, $\pi_t=\clip_{[\underline\pi,1]}(s_t/\tau)$, where $\tau$ is calibrated on unlabeled instances so that the average audit probability equals $\rho$.
Oracle uses the rule \eqref{eq:oracle-allocation} with known $\sigma(Z)$.
Figure~\ref{figure:a1} shows that AAELI nearly matches the oracle and saves 10.4\% of human labels relative to uniform auditing at width $0.06$, whereas auditing by judge uncertainty performs similarly to uniform auditing and auditing by judge disagreement is markedly worse.
This confirms the oracle rule \eqref{eq:oracle-allocation} in Theorem~\ref{thm:oracle-annotation}, which allocates human audits in proportion to the conditional standard deviation $\sigma(Z)$ of the human label rather than the uncertainty or disagreement of the LLM judges.

\begin{figure*}[htbp]
\centering
\includegraphics[width=0.45\textwidth]{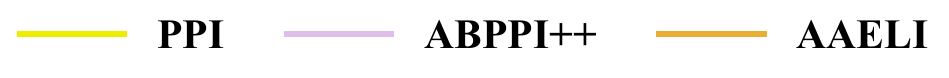}
\subfloat[Confidence Sequences]{%
    \includegraphics[width=0.33\textwidth]{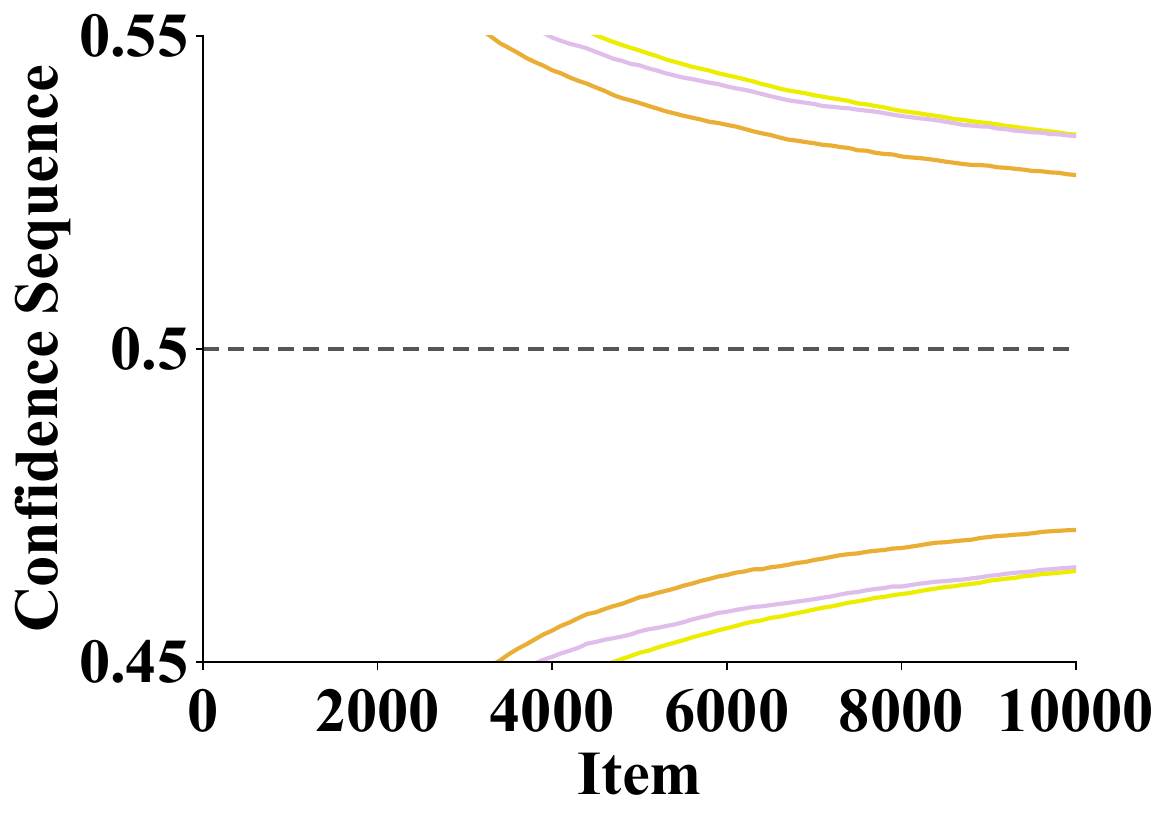}
}
\subfloat[Anytime Coverage]{%
    \includegraphics[width=0.33\textwidth]{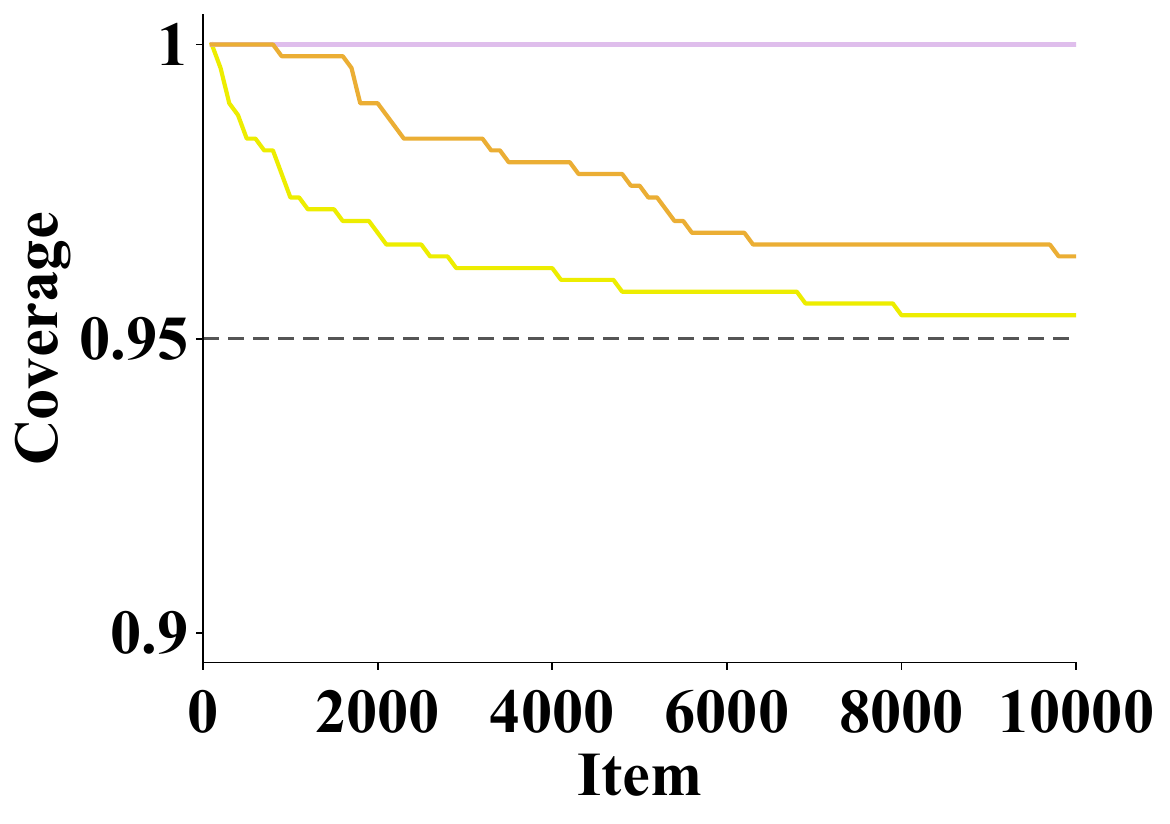}
}
\subfloat[Width]{%
    \includegraphics[width=0.33\textwidth]{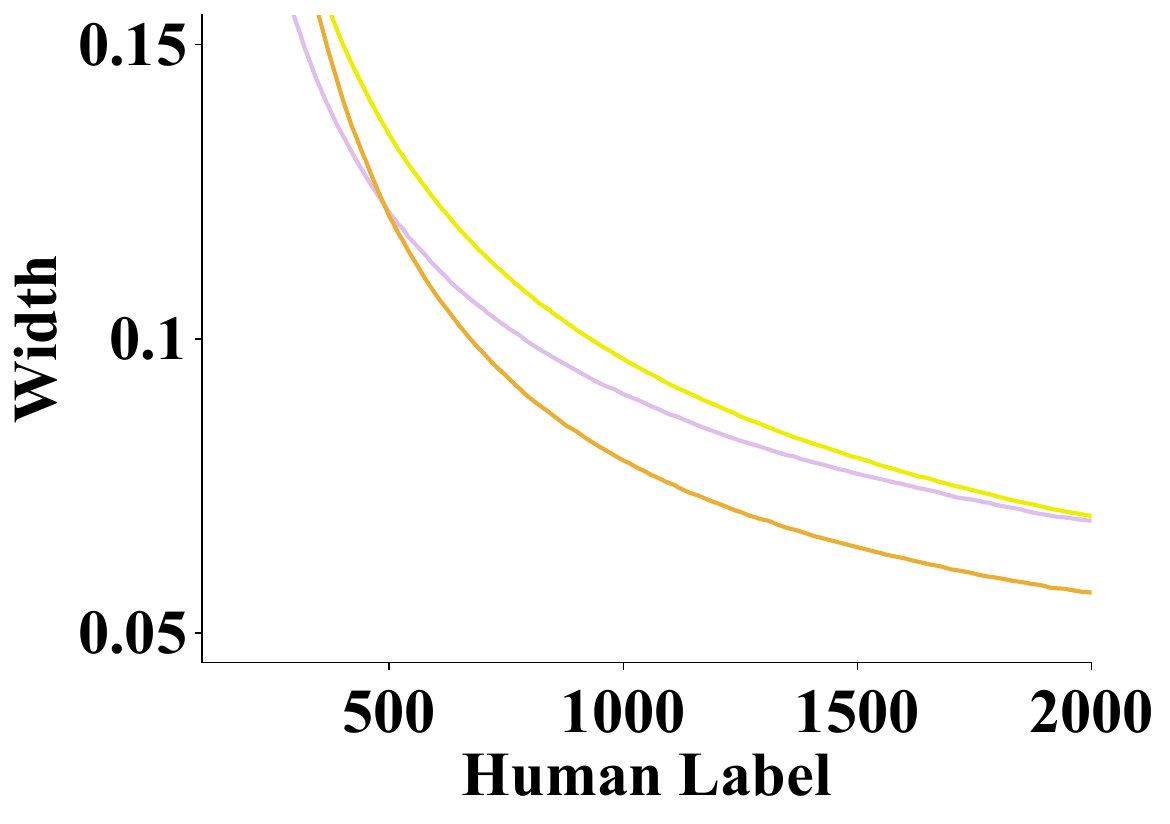}
}
\caption{Ablation on the empirical likelihood inference procedure.
PPI, ABPPI++, and AAELI use the same uniform auditing, but construct confidence sequences by betting on the PPI estimator \citep{waudby2024time,zrnic2024active}, by anytime-valid Bayes-assisted prediction-powered inference with the PPI++ estimator \citep{kilian2026anytime,angelopoulos2026ppi++}, and by the AAELI inference procedure.}
\label{figure:a2}
\end{figure*}

\textit{Ablation Study on Empirical Likelihood Inference.}
We examine the effect of the empirical-likelihood inference procedure by fixing the audit policy to uniform auditing.
We compare confidence sequences built by betting on the PPI estimator (PPI) \citep{waudby2024time,zrnic2024active}, by anytime-valid Bayes-assisted prediction-powered inference with the PPI++ estimator (ABPPI++) \citep{kilian2026anytime,angelopoulos2026ppi++}, and by the AAELI inference procedure (AAELI).
Figure~\ref{figure:a2} shows that, at width $0.07$, AAELI requires on average 1{,}186 human labels, saving 36.4\% relative to PPI and 29.8\% relative to ABPPI++.
Moreover, the average widths of PPI and ABPPI++ do not reach $0.06$ within the budget, whereas that of AAELI does.
Since all three procedures share the same uniform audit rule, these differences reflect both how each procedure uses the auxiliary information and its betting construction: PPI and ABPPI++ rely on a single prescribed or power-tuned combination of the judge predictions, whereas AAELI learns the variance-minimizing combination of the augmented moments.
The matched comparison below, in which the betting construction is held fixed, isolates the contribution of the learned combination.

\begin{figure*}[htbp]
\centering
\includegraphics[width=0.6\textwidth]{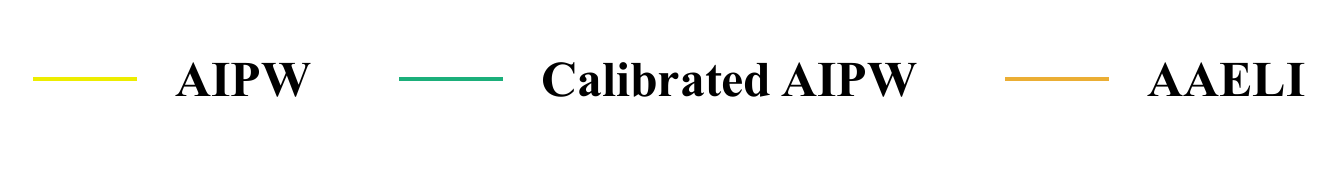}
\subfloat[Confidence Sequences]{%
    \includegraphics[width=0.33\textwidth]{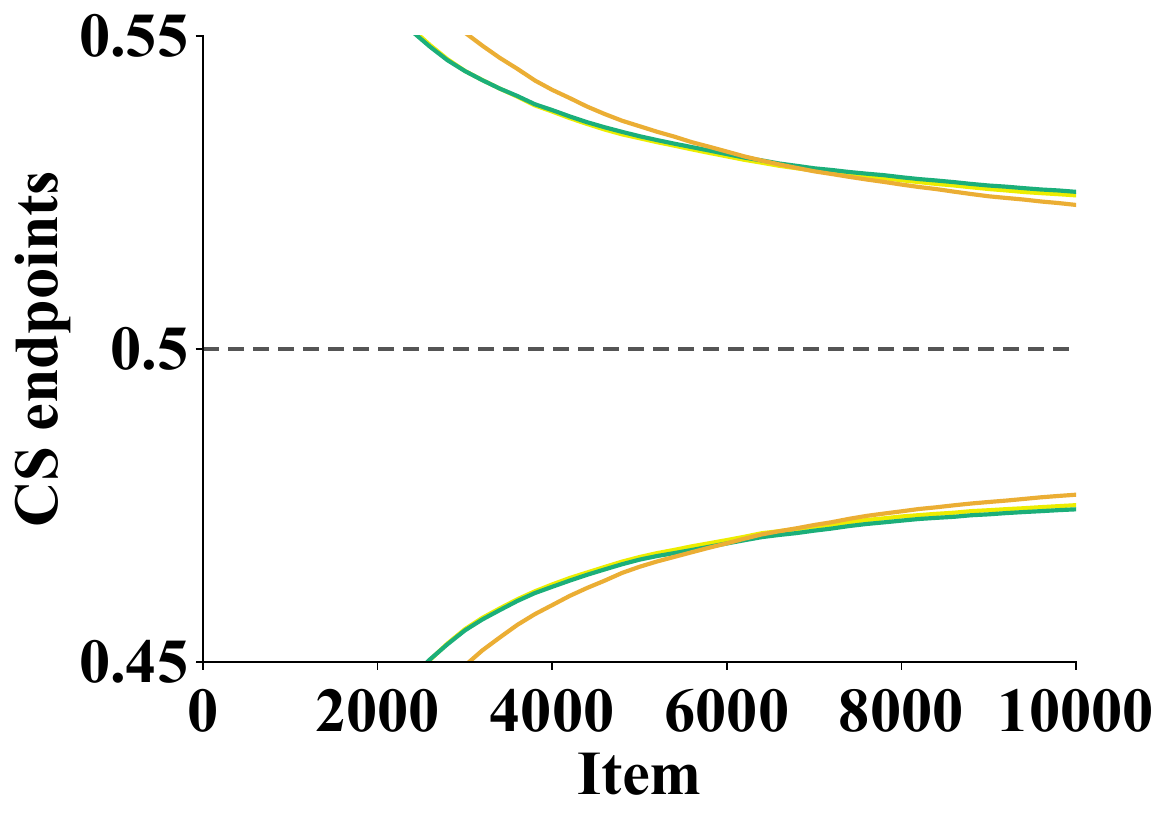}
}
\subfloat[Anytime Coverage]{%
    \includegraphics[width=0.33\textwidth]{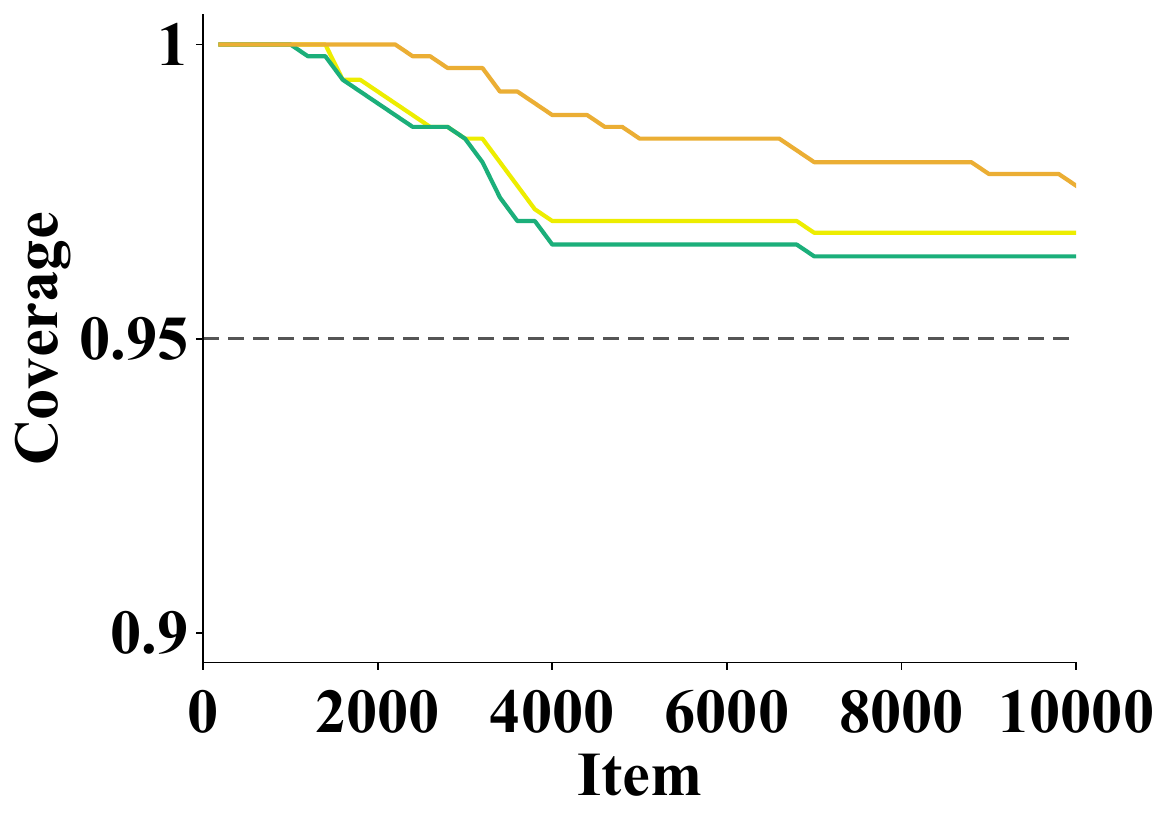}
}
\subfloat[Width]{%
    \includegraphics[width=0.33\textwidth]{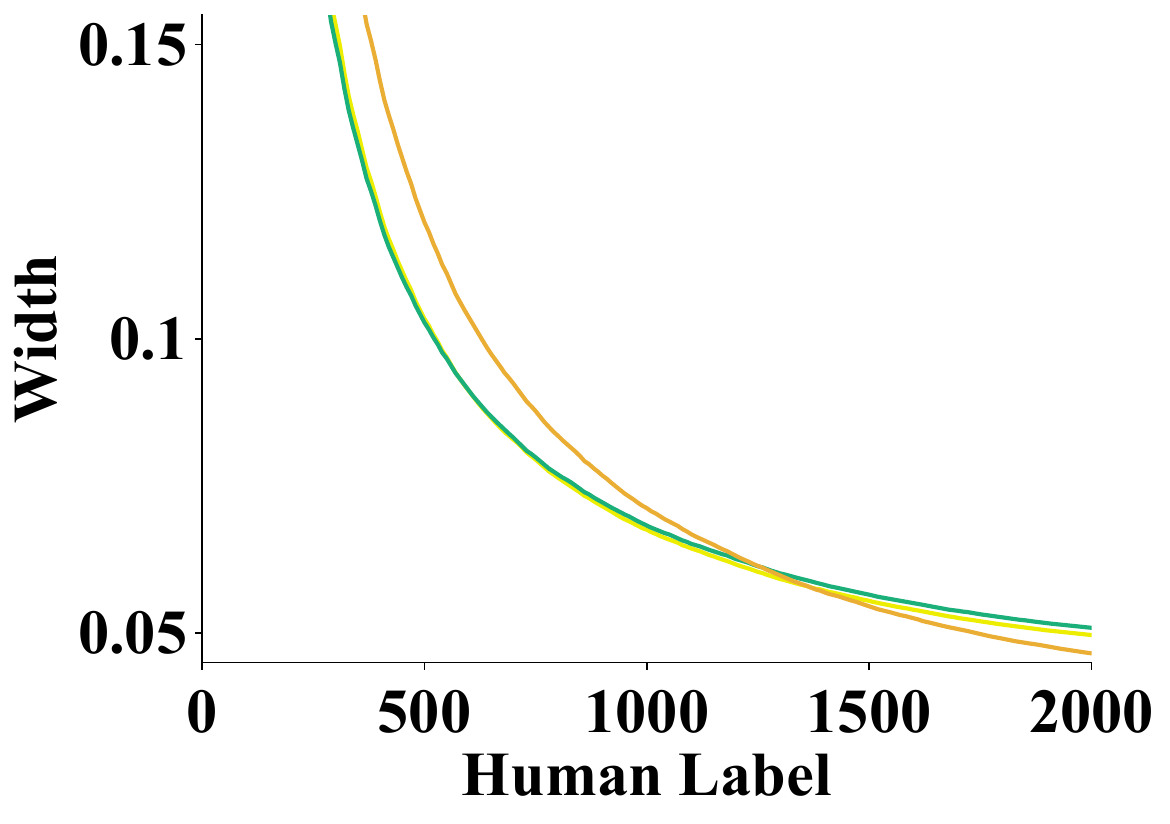}
}
\caption{Ablation on the learned moment combination with unmodified random-forest predictions.}
\label{fig:ablation-aipw-calibrated}
\end{figure*}

\begin{figure*}[htbp]
\centering
\includegraphics[width=0.6\textwidth]{Figure/Legend3.pdf}
\subfloat[Confidence Sequences]{%
    \includegraphics[width=0.33\textwidth]{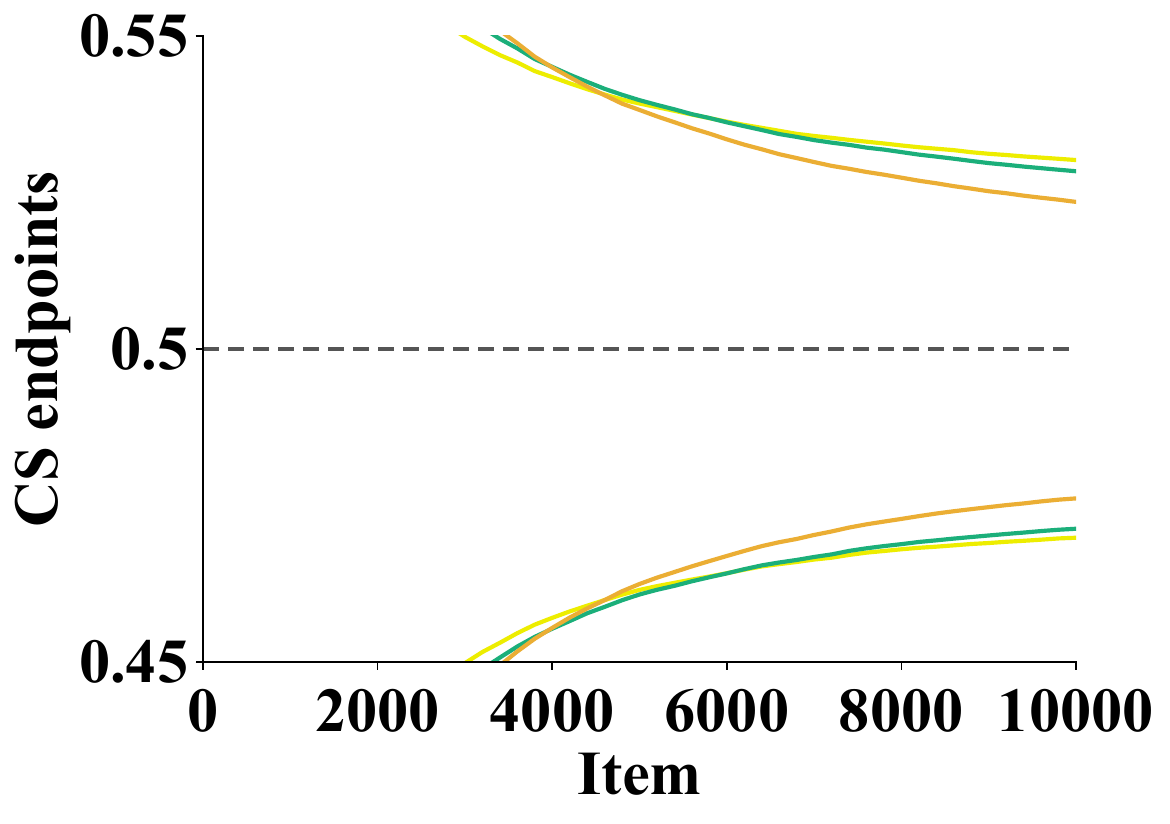}
}
\subfloat[Anytime Coverage]{%
    \includegraphics[width=0.33\textwidth]{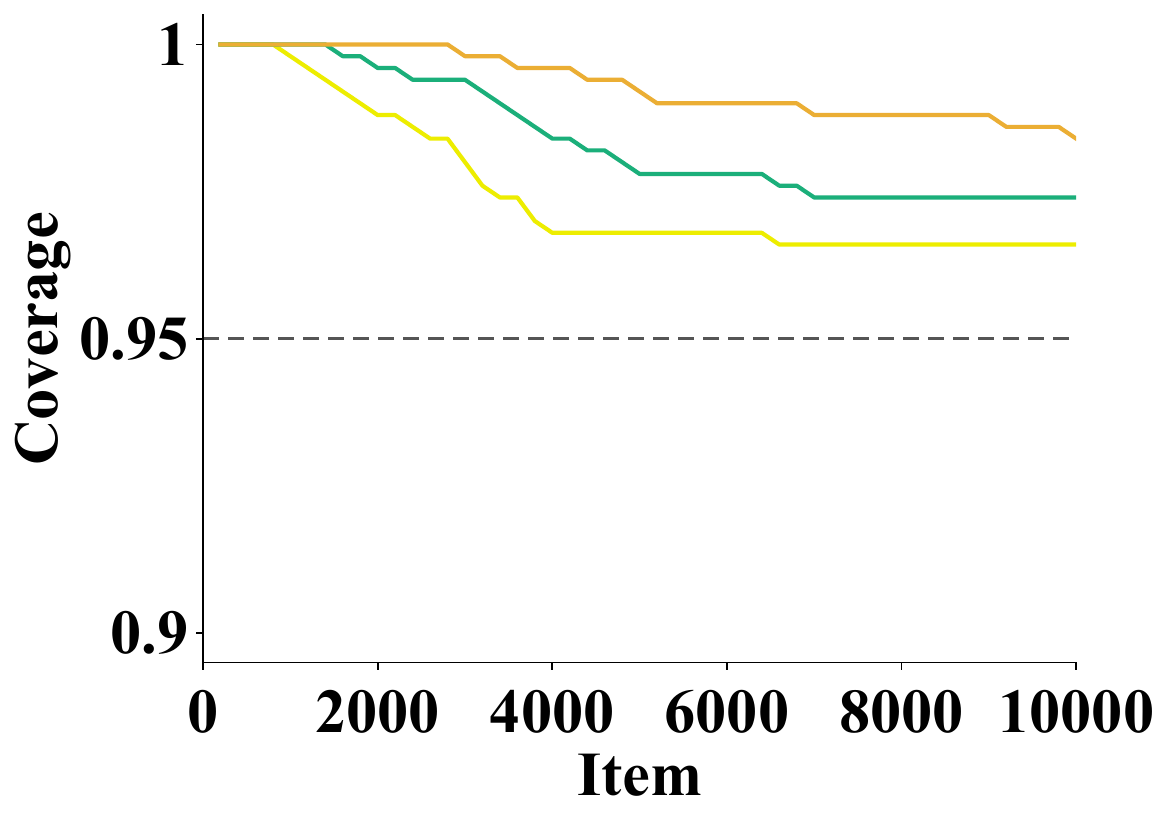}
}
\subfloat[Width]{%
    \includegraphics[width=0.33\textwidth]{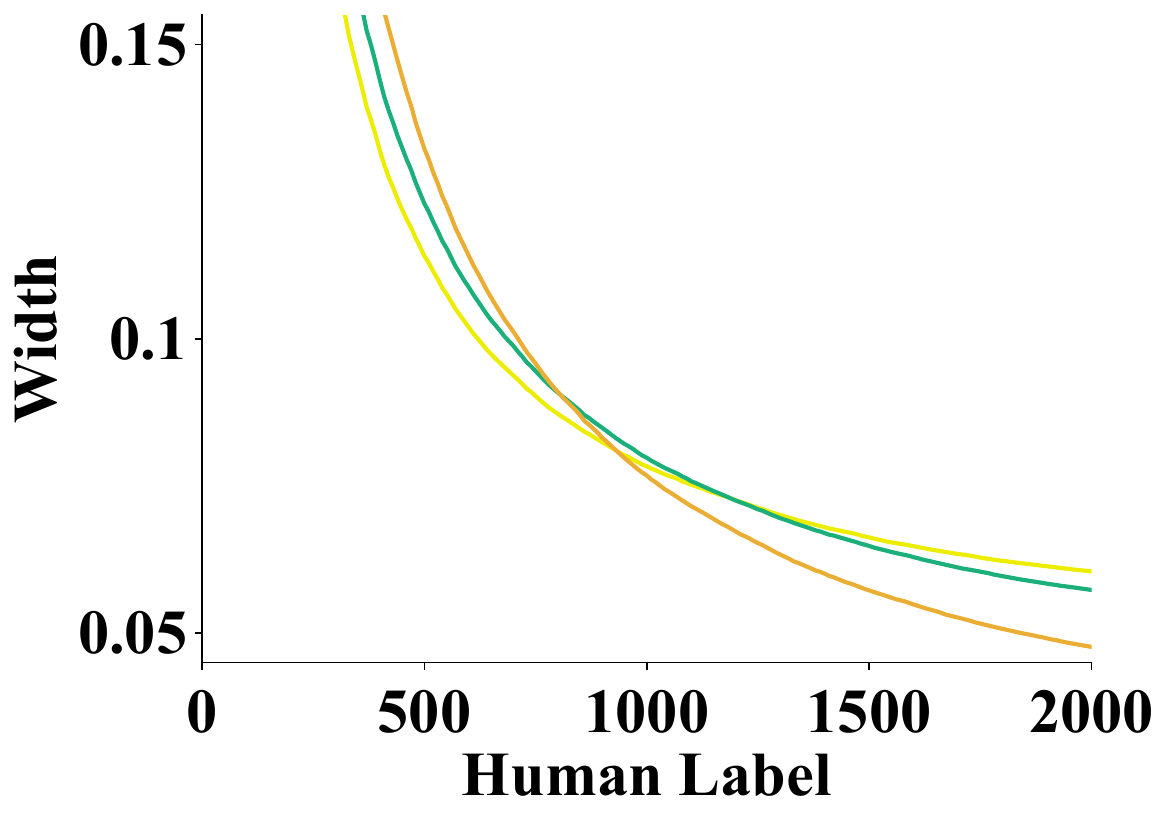}
}
\caption{Ablation on the learned moment combination under affine miscalibration of random-forest predictions.}
\label{fig:ablation-aipw-affine}
\end{figure*}

\textit{Ablation on the Learned Moment Combination.}
We compare AAELI with two matched baselines: betting on the standard AIPW score with prescribed coefficients, and betting on an AIPW score whose predictor is affinely calibrated by IPW ridge regression using only past audited observations.
To examine predictor miscalibration, we transform the prequential random-forest predictions as $f_t=0.15+0.60\widehat m_{t-1}(Z_t)$.
This transformation is applied identically to all three methods and leaves the audit design and random-forest training unchanged.
All methods use uniform auditing, identical audited labels and predictor outputs, the same equally weighted betting grid, and running intersections.
Scores are scaled using past observations, and safe betting bounds are obtained by worst-case bounding.
The resulting scales and effective betting fractions may differ across methods.
AAELI estimates its projection direction separately for each candidate mean from past second moments.
Across 500 replications, the three methods perform comparably with well-calibrated predictions (Figure~\ref{fig:ablation-aipw-calibrated}), although AAELI requires fewer human labels to reach narrower target widths.
Under affine predictor miscalibration (Figure~\ref{fig:ablation-aipw-affine}), AAELI reaches the target average confidence-sequence widths with fewer human labels than standard and past-data calibrated AIPW betting.
The results support the finite-sample performance of the implemented procedure in this miscalibrated-predictor setting.

\begin{figure*}[htbp]
\centering
\includegraphics[width=0.7\textwidth]{Figure/Legend.pdf}
\subfloat[Confidence Sequences]{%
    \includegraphics[width=0.33\textwidth]{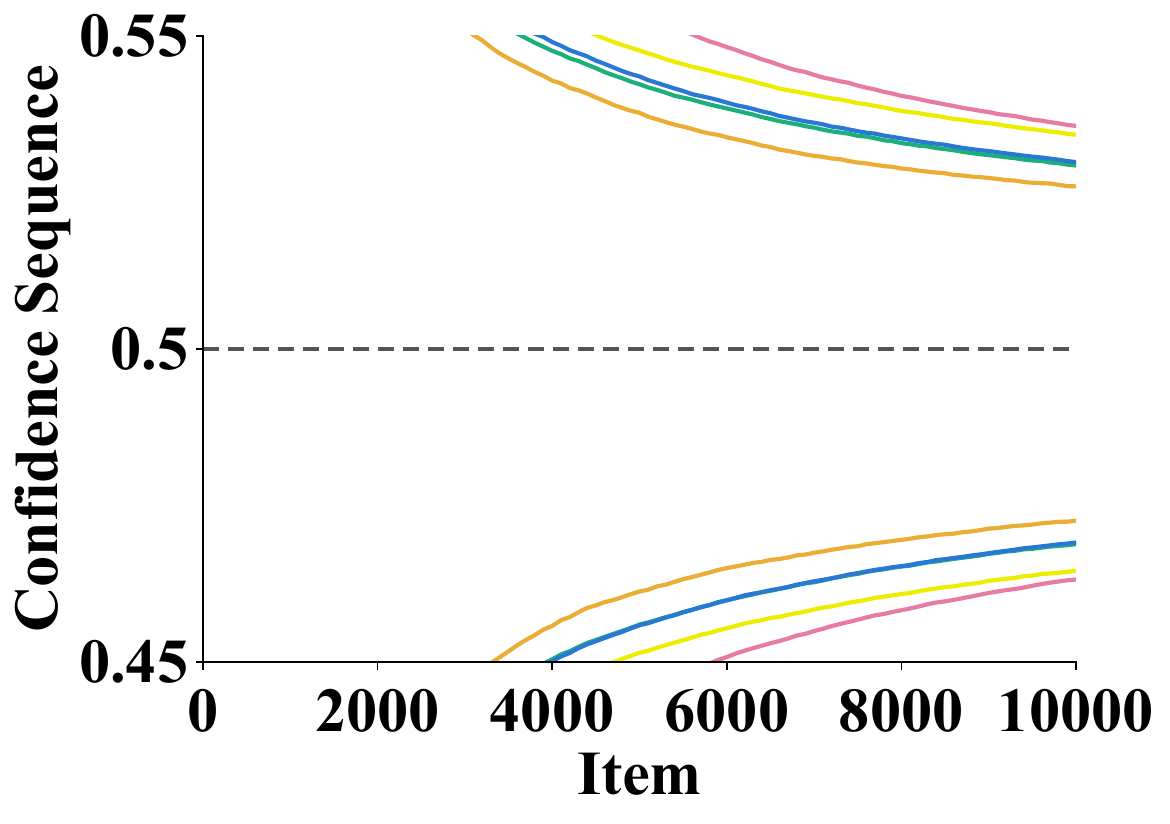}
}
\subfloat[Anytime Coverage]{%
    \includegraphics[width=0.33\textwidth]{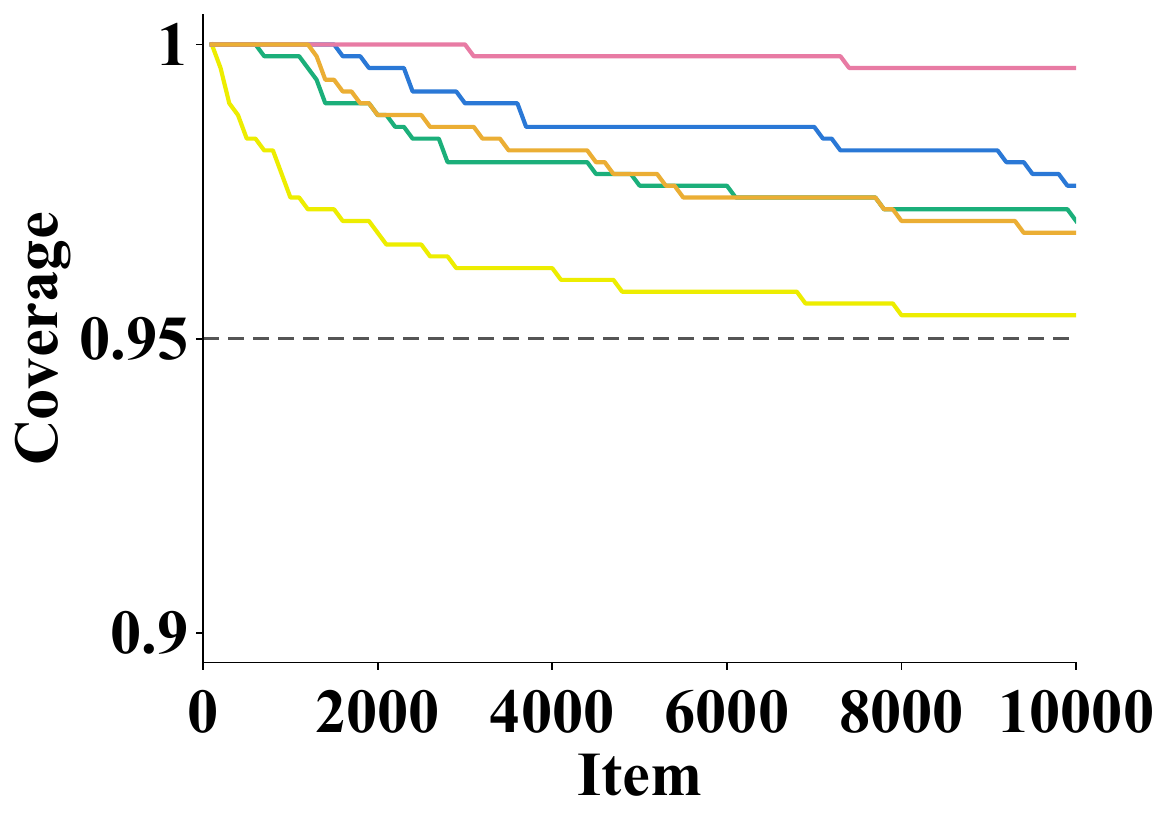}
}
\subfloat[Width]{%
    \includegraphics[width=0.33\textwidth]{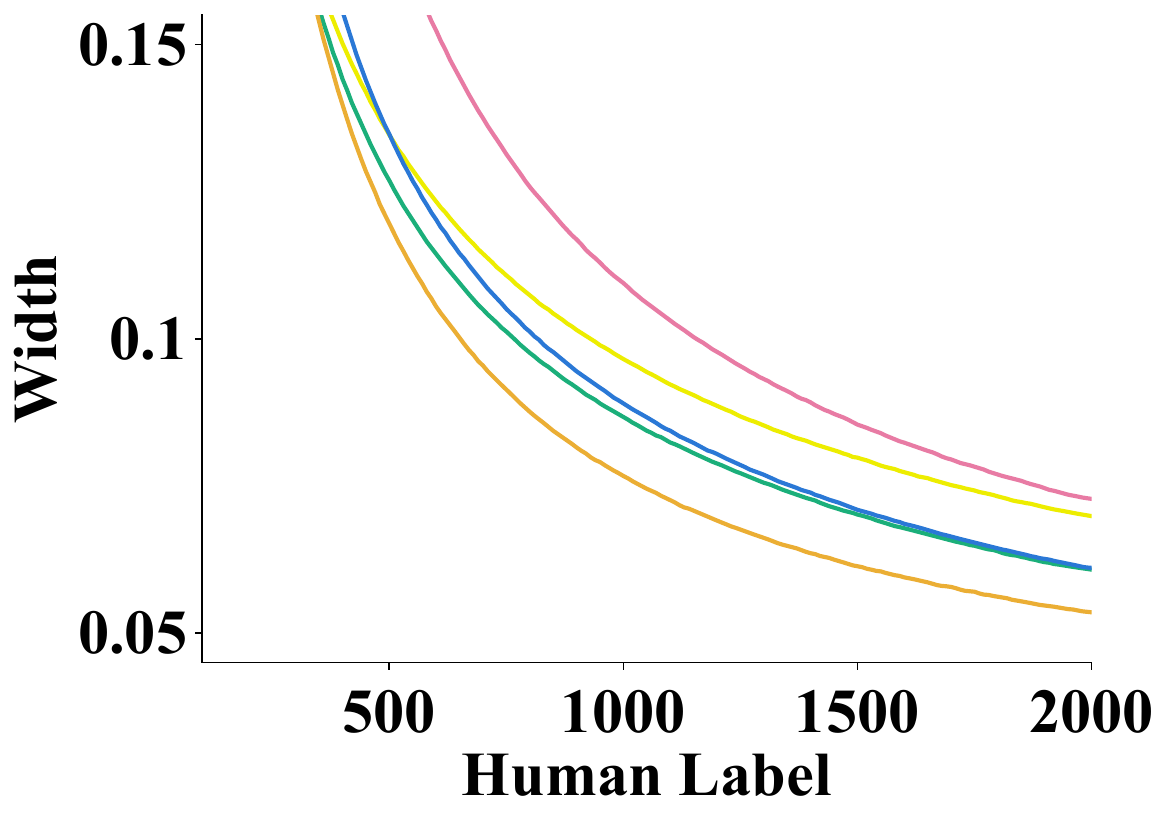}
}
\caption{Robustness to biased LLM judges.}
\label{figure:a3}
\end{figure*}

\begin{figure*}[htbp]
\centering
\includegraphics[width=0.7\textwidth]{Figure/Legend.pdf}
\subfloat[Confidence Sequences]{%
    \includegraphics[width=0.33\textwidth]{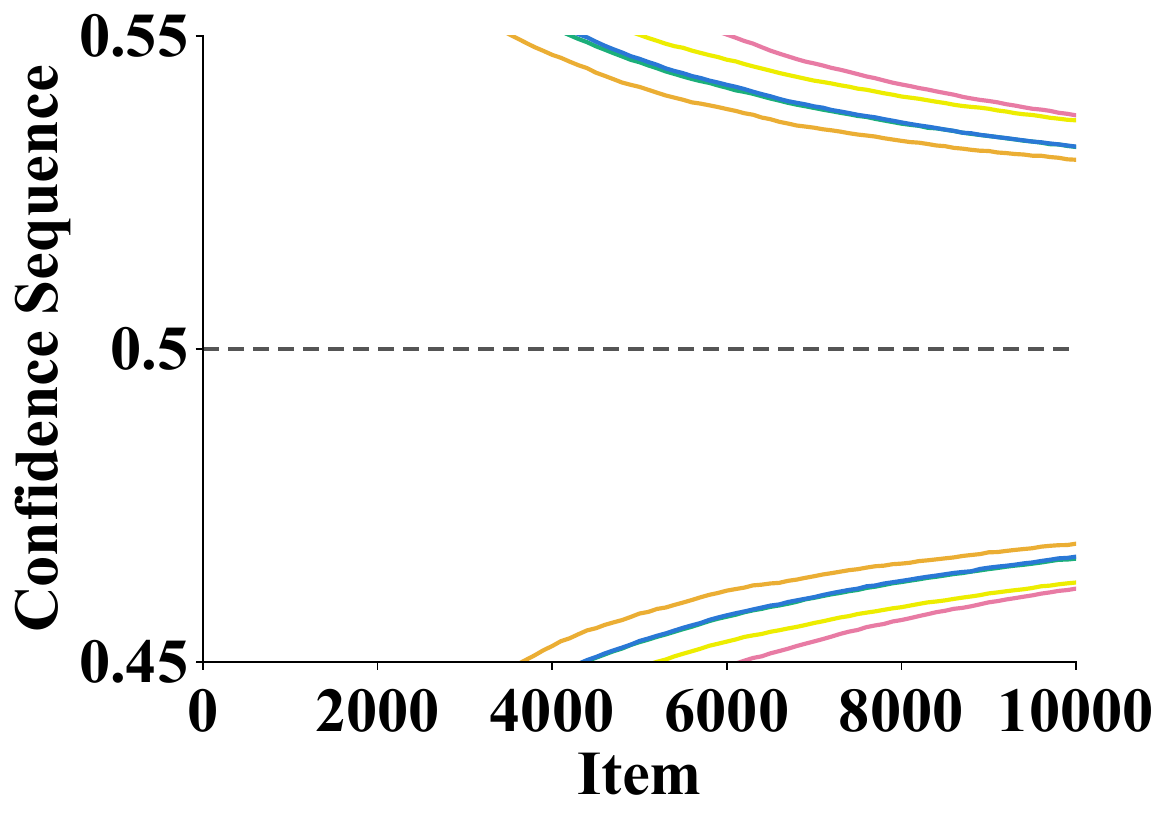}
}
\subfloat[Anytime Coverage]{%
    \includegraphics[width=0.33\textwidth]{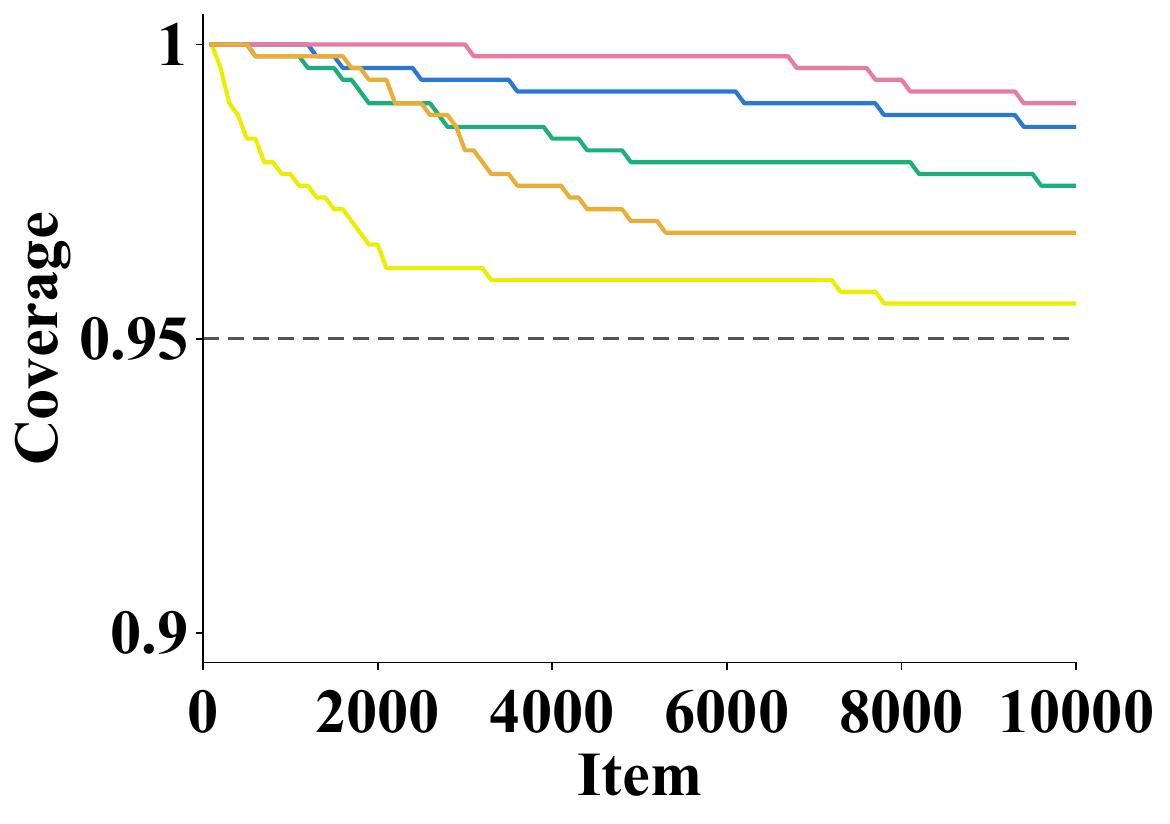}
}
\subfloat[Width]{%
    \includegraphics[width=0.33\textwidth]{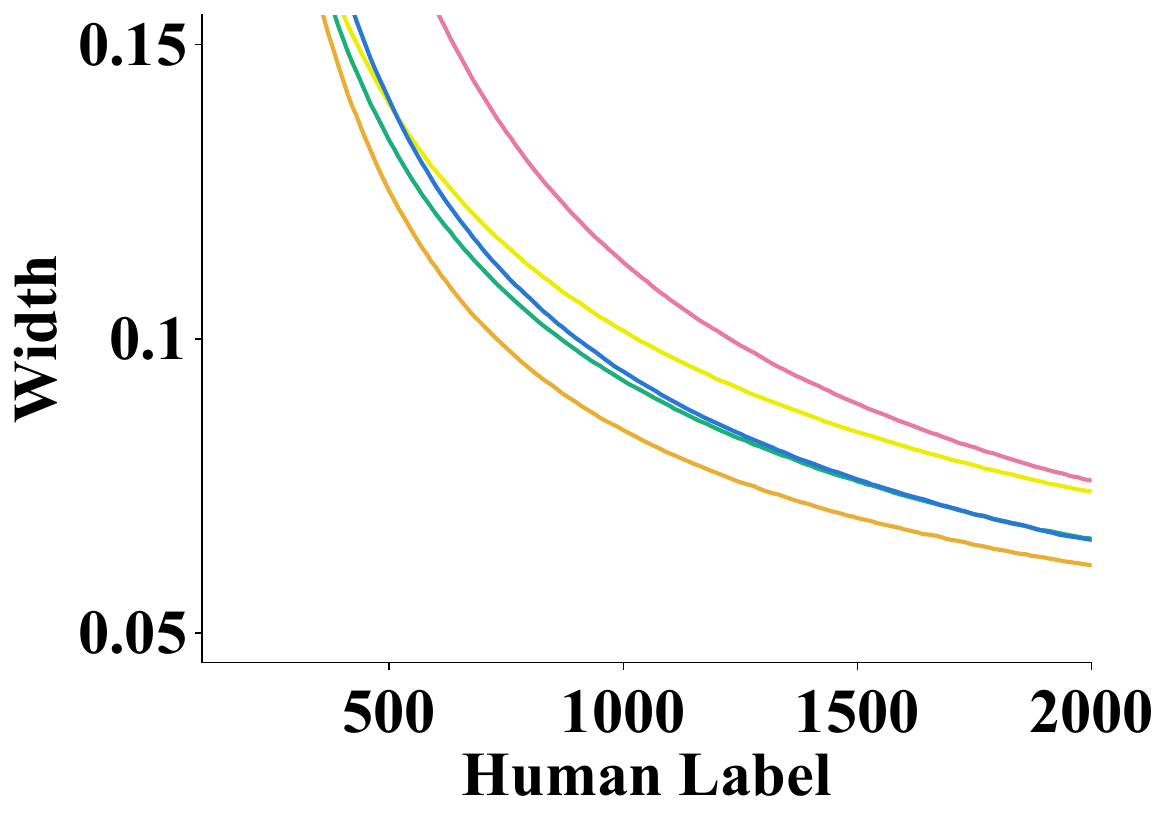}
}
\caption{Robustness to highly correlated LLM judges.}
\label{figure:a4}
\end{figure*}

\begin{figure*}[htbp]
\centering
\includegraphics[width=0.7\textwidth]{Figure/Legend.pdf}
\subfloat[Confidence Sequences]{%
    \includegraphics[width=0.33\textwidth]{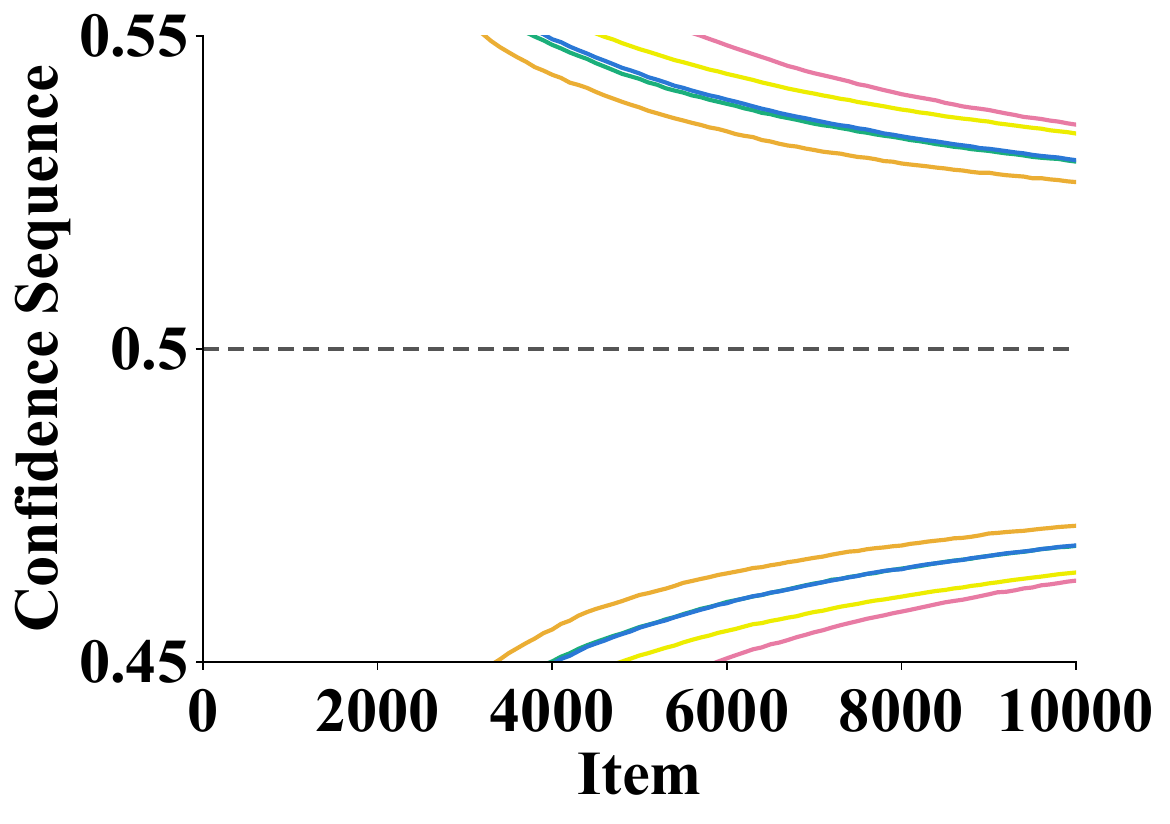}
}
\subfloat[Anytime Coverage]{%
    \includegraphics[width=0.33\textwidth]{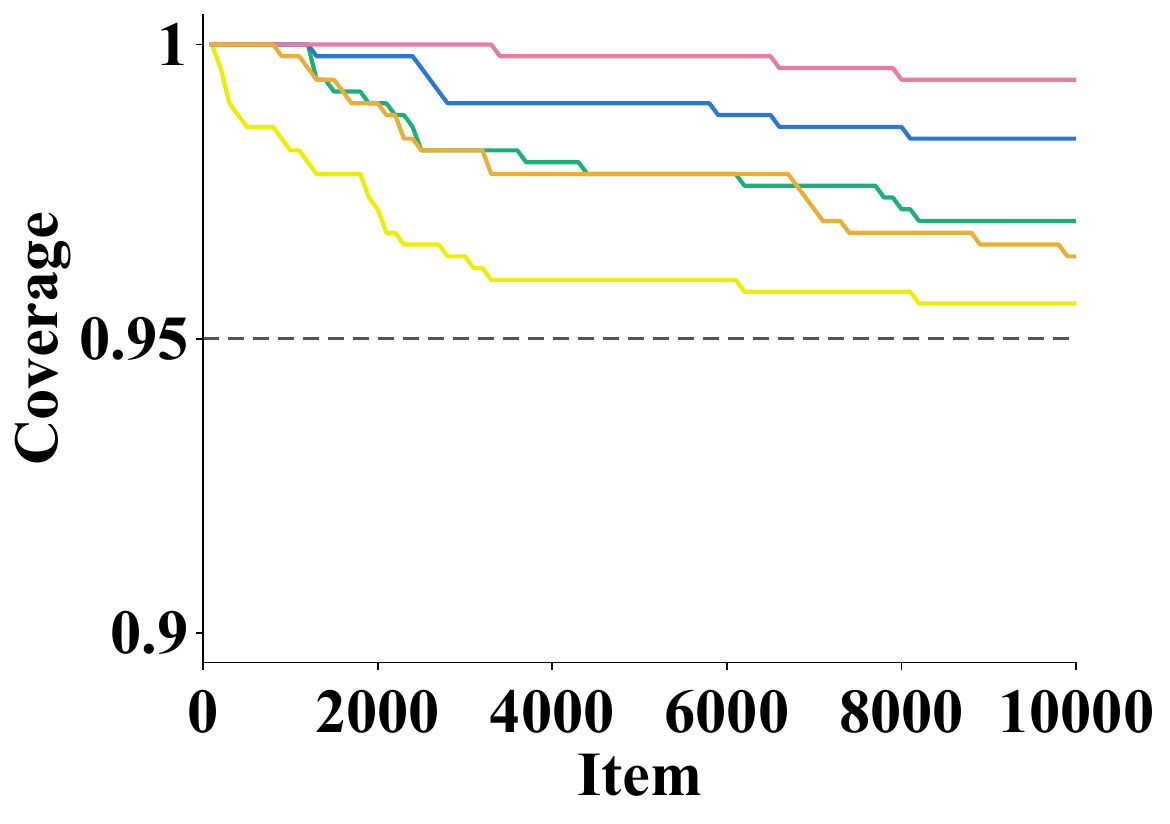}
}
\subfloat[Width]{%
    \includegraphics[width=0.33\textwidth]{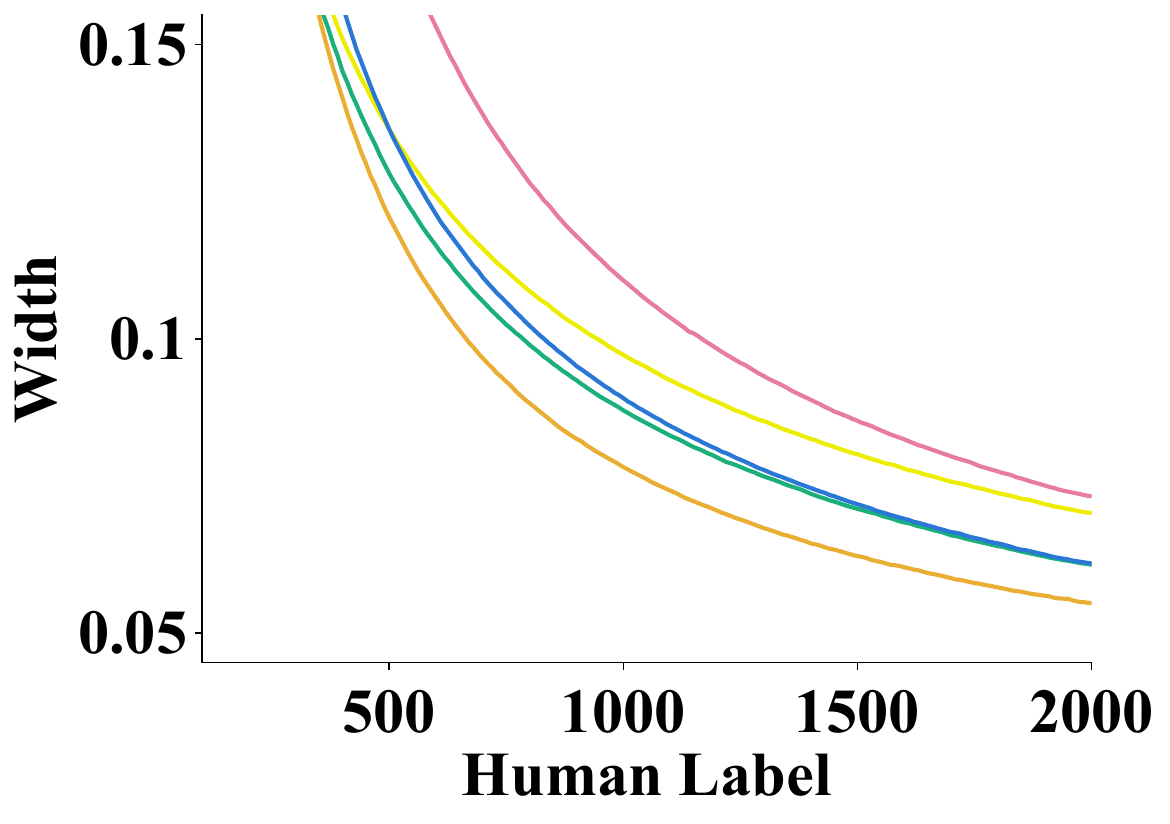}
}
\caption{Robustness to shared judge failure.}
\label{figure:a5}
\end{figure*}

\textit{Robustness to Imperfect LLM Judges.}
In this section, we also examine whether AAELI remains valid and efficient when the LLM judges are imperfect. 
We consider three departures from the baseline setting.
First, biased judges have all their logits shifted upward by 1.5. Second, correlated judges share noise with standard deviation 1.5 instead of 0.3, which raises the correlation of their errors from about 0.16 to 0.82. 
Third, under shared judge failure, the loadings of all judges on the latent factor are reversed on a common subset of about 8\% of the instances.
Figures~\ref{figure:a3}--\ref{figure:a5} show that AAELI maintains anytime validity in all three settings and continues to require the fewest human labels.
This is consistent with Proposition~\ref{prop:estimating-equations}: validity relies on the randomized audits rather than on the accuracy of the judges, and the judges affect only efficiency, through the information they carry about the human label and how well $m(Z)$ is learned (Theorem~\ref{thm:eff-decomp} and Remark~\ref{rem:information}).

Moreover, we conduct a series of robustness studies under diverse experimental conditions.
First, we vary the target human audit rate over $\rho\in\{0.05,0.10,0.15\}$, corresponding to nominal targets of 500, 1{,}000, and 1{,}500 human labels, respectively.
Figures~\ref{figure:a6}--\ref{figure:a8} show that AAELI maintains anytime validity and requires the fewest human labels across all budgets, with larger relative savings at smaller budgets.
Second, we replace the random forest \citep{breiman2001random} used by all learning-based methods with a multilayer perceptron \citep{rumelhart1986learning} with two hidden layers and a support vector regression \citep{drucker1996support} with an RBF kernel.
Figures~\ref{figure:a9} and~\ref{figure:a10} show that the same conclusions hold under both learners.

\section{Real Data Application}
\label{sec:auto:6}
\label{sec:real}
\subsection{Application to LLM--Human Alignment Measurement}
\label{sec:auto:6.1}
Large language models increasingly serve as reward models for reinforcement learning \citep{gunjal2026rubrics} and as judges of model outputs \citep{zheng2023judging}, creating a growing need to reliably measure how well their judgments align with human preferences.
Specifically, let $J_t$ and $H_t$ denote the LLM judge's choice and the human preference on item $t$, respectively.
We define $Y_t=\mathbf{1}\{J_t=H_t\}$ and quantify alignment with human preferences by the agreement rate $\mu_0=\mathbb{E}[Y_t]$.
Since $Y_t$ is observed only when item $t$ is audited by a human, our method applies directly and estimates $\mu_0$ from limited human labels.
We evaluate the alignment of DeepSeek-V4-Pro-0813 \citep{xu2026deepseek} with human preferences on Chatbot Arena \citep{chiang2024chatbot}, using the outputs of three low-cost judges, GPT-6-Luna \citep{openai2026gpt6}, Claude-Haiku-5.5 \citep{anthropic2026haiku55}, and DeepSeek-V4.1-Flash-1017 \citep{xu2026deepseek}.
The corpus consists of $N = 10{,}000$ aligned battles, and each replicate stream is drawn uniformly at random with replacement from the corpus, so that $\mu_0$ is the corpus agreement rate.

\begin{figure*}[htbp]
\centering
\includegraphics[width=0.7\textwidth]{Figure/Legend.pdf}
\subfloat[Confidence Sequences]{%
    \includegraphics[width=0.34\textwidth]{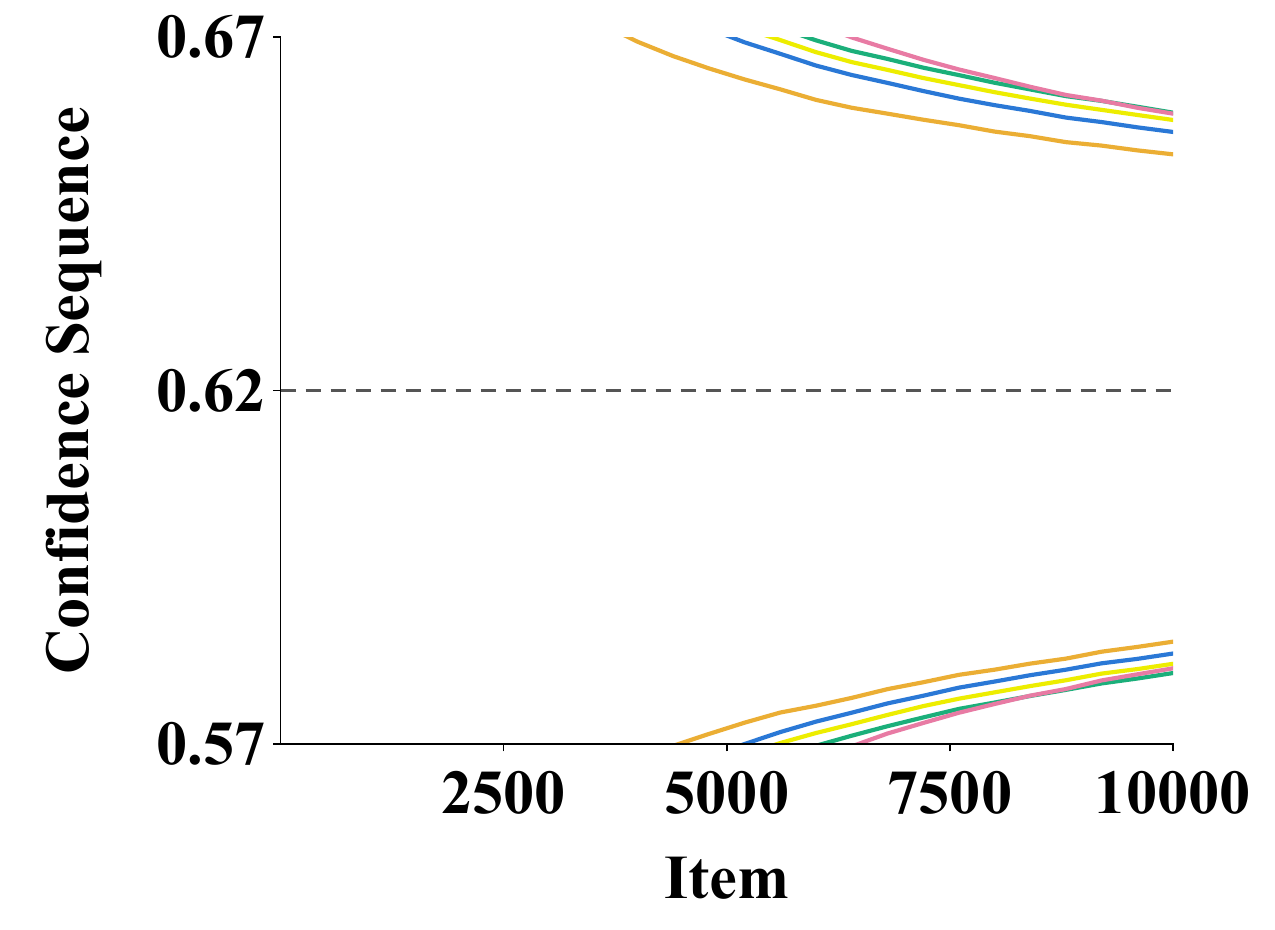}
}
\subfloat[Anytime Coverage]{%
    \includegraphics[width=0.34\textwidth]{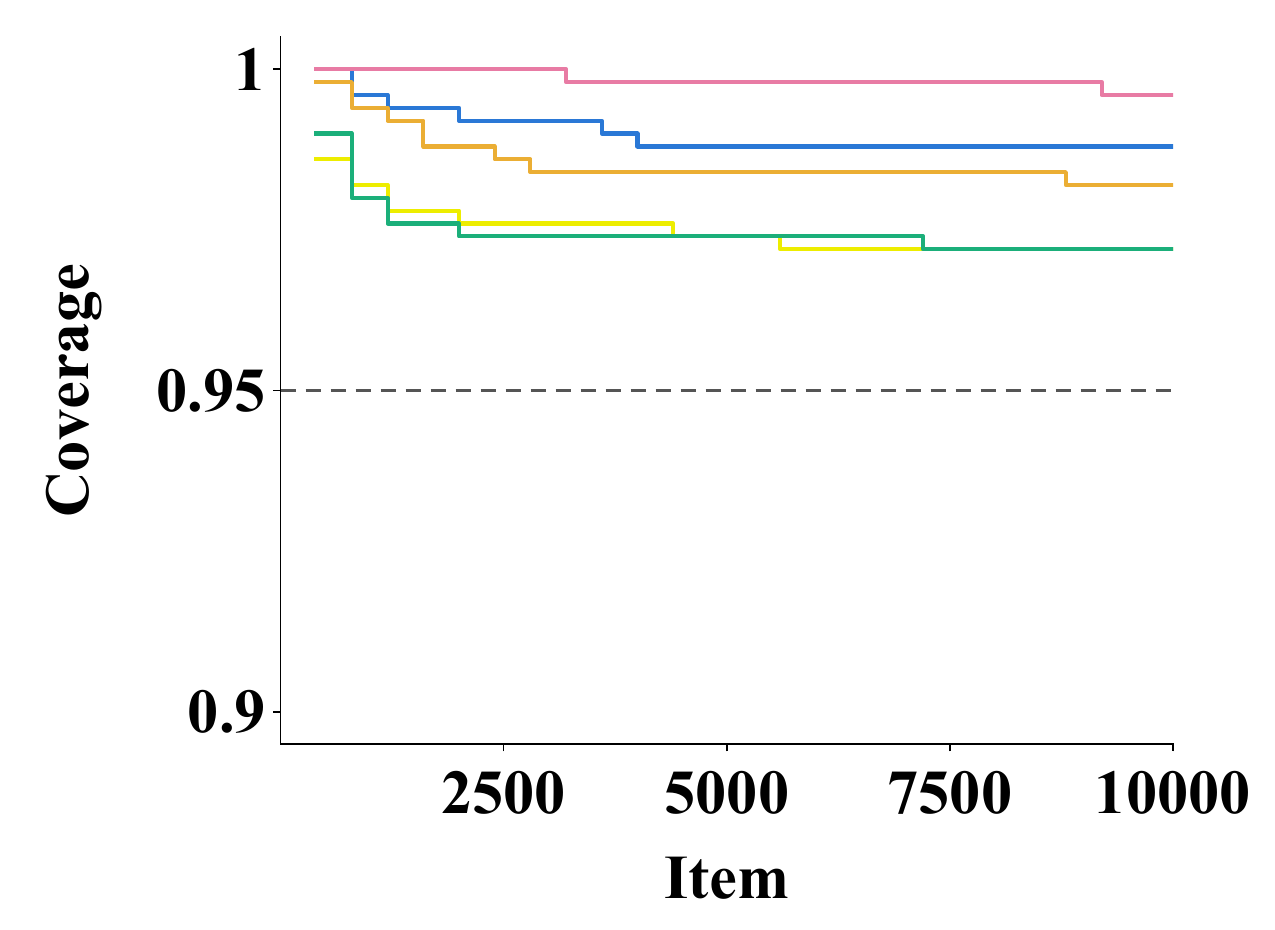}
}
\subfloat[Width]{%
    \includegraphics[width=0.34\textwidth]{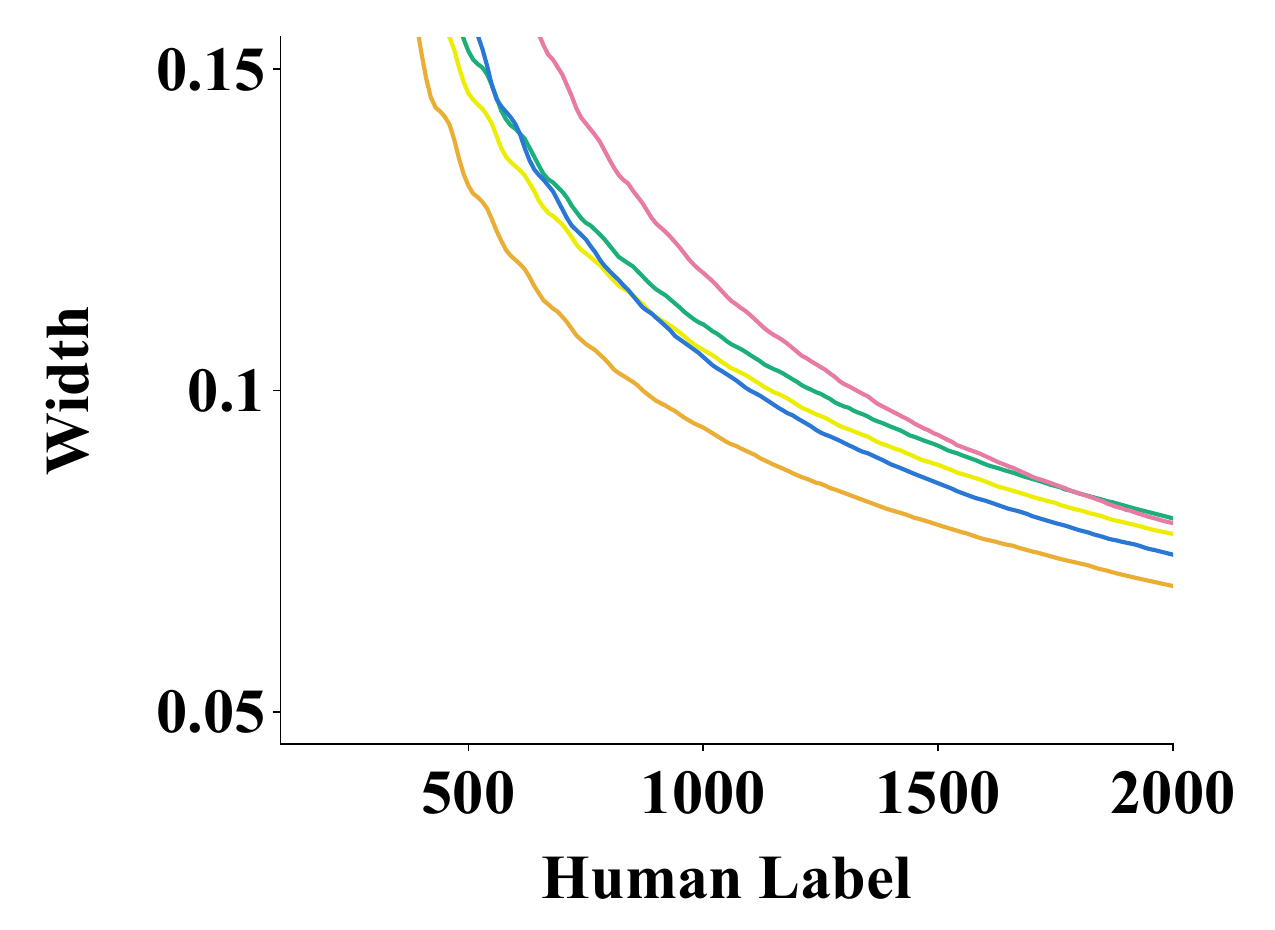}
}
\caption{Anytime-valid inference on Chatbot Arena with human budget $\rho=0.2$.}
\label{figure:2}
\end{figure*}

Figure~\ref{figure:2} shows the results on Chatbot Arena: confidence-sequence bounds (left), anytime coverage (middle), and interval width versus human labels (right).
The left panel shows that all confidence sequences contract around the true agreement rate $\mu\approx0.62$, with ours yielding the tightest bounds throughout the stream.
The middle panel shows that every anytime-valid procedure maintains coverage above the nominal level $1-\alpha=0.95$ uniformly over the stream.
The right panel shows that AAELI reaches any given confidence-sequence width with fewer human labels than every baseline.
For example, AAELI needs about 870 human labels to reach width $w=0.10$, whereas the strongest baseline at this width, RASI, needs about 1{,}100, a saving of roughly 21\%.
Relative to the strongest baseline at each width, the number of labels saved grows as the target width shrinks, from about 80 at $w=0.15$ to 170 at $w=0.12$ and 260 at $w=0.08$, corresponding to relative savings of 15\%--23\%.
Moreover, AAELI is the only method that reaches width $w=0.07$ within the budget, requiring about 1{,}970 human labels.
Figure~\ref{figure:chatbox1} shows that AAELI remains the most label-efficient method with GPT-6-Luna as the only LLM judge.

\subsection{Application to Political Articles Analysis}
\label{sec:auto:6.2}
In this section, we aim to study the mean political leaning of news articles. 
Monitoring political leaning is of practical interest for news aggregators and media monitoring, yet obtaining human labels for every article is costly. 
We use the AllSides dataset \citep{baly2020we}, combining its validation and test sets for a total of 8{,}296 articles.
Each article has a human-annotated political leaning $H_t \in \{\textit{left}, \textit{center}, \textit{right}\}$, which we encode as $Y_t = 0$, $1/2$, and $1$, respectively. 
The inferential target is the mean political leaning, $\mu_0 = \mathbb{E}[Y_t] = \tfrac{1}{2}\mathbb{P}(H_t = \textit{center}) + \mathbb{P}(H_t = \textit{right})$. 
Since $\mu_0 - \tfrac{1}{2} = \tfrac{1}{2}\{\mathbb{P}(H_t = \textit{right}) - \mathbb{P}(H_t = \textit{left})\}$, the value $\mu_0 = 1/2$ indicates equal proportions of left- and right-leaning articles, while $\mu_0 > 1/2$ ($\mu_0 < 1/2$) indicates a net right (left) leaning.
Under the resampling protocol, $\mu_0$ is the mean of $Y_t$ over the full corpus, which gives $\mu_0 = 0.52$ and indicates a slight net right leaning.
We use the outputs of the same three low-cost judges as auxiliary information in $Z_t$.

\begin{figure*}[htbp]
\centering
\includegraphics[width=0.7\textwidth]{Figure/Legend.pdf}
\subfloat[Confidence Sequences]{%
    \includegraphics[width=0.34\textwidth]{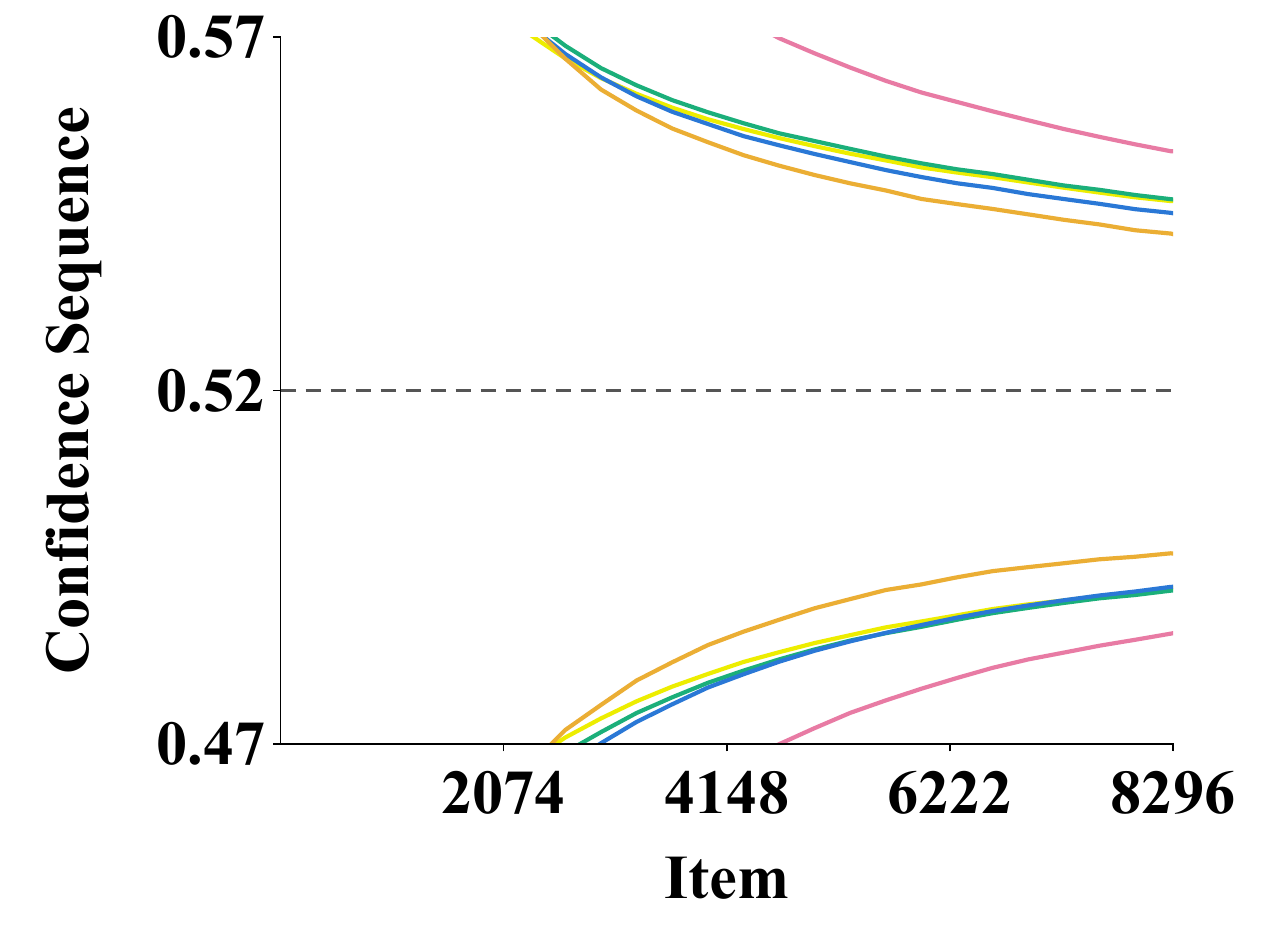}
}
\subfloat[Anytime Coverage]{%
    \includegraphics[width=0.34\textwidth]{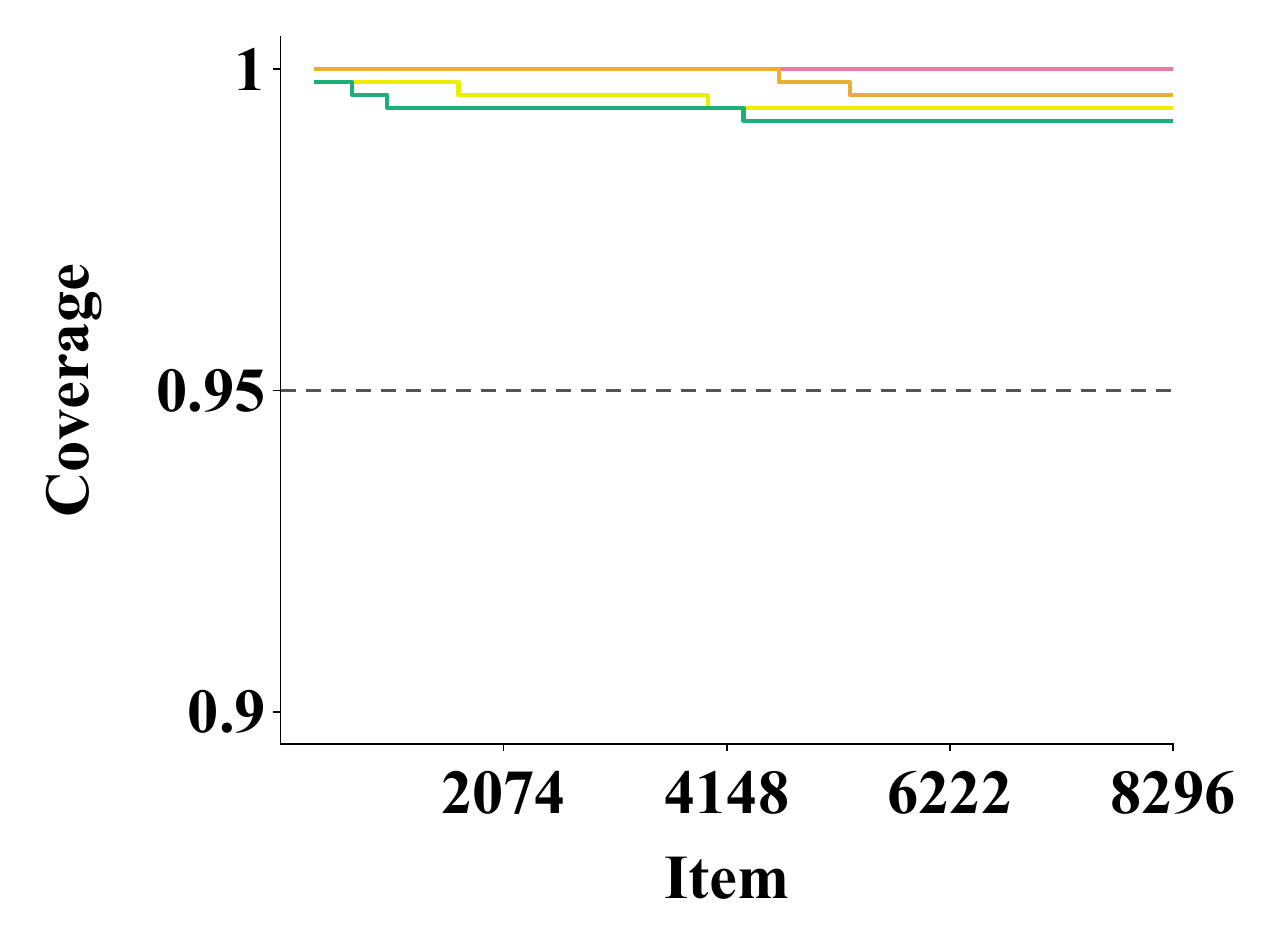}
}
\subfloat[Width]{%
    \includegraphics[width=0.34\textwidth]{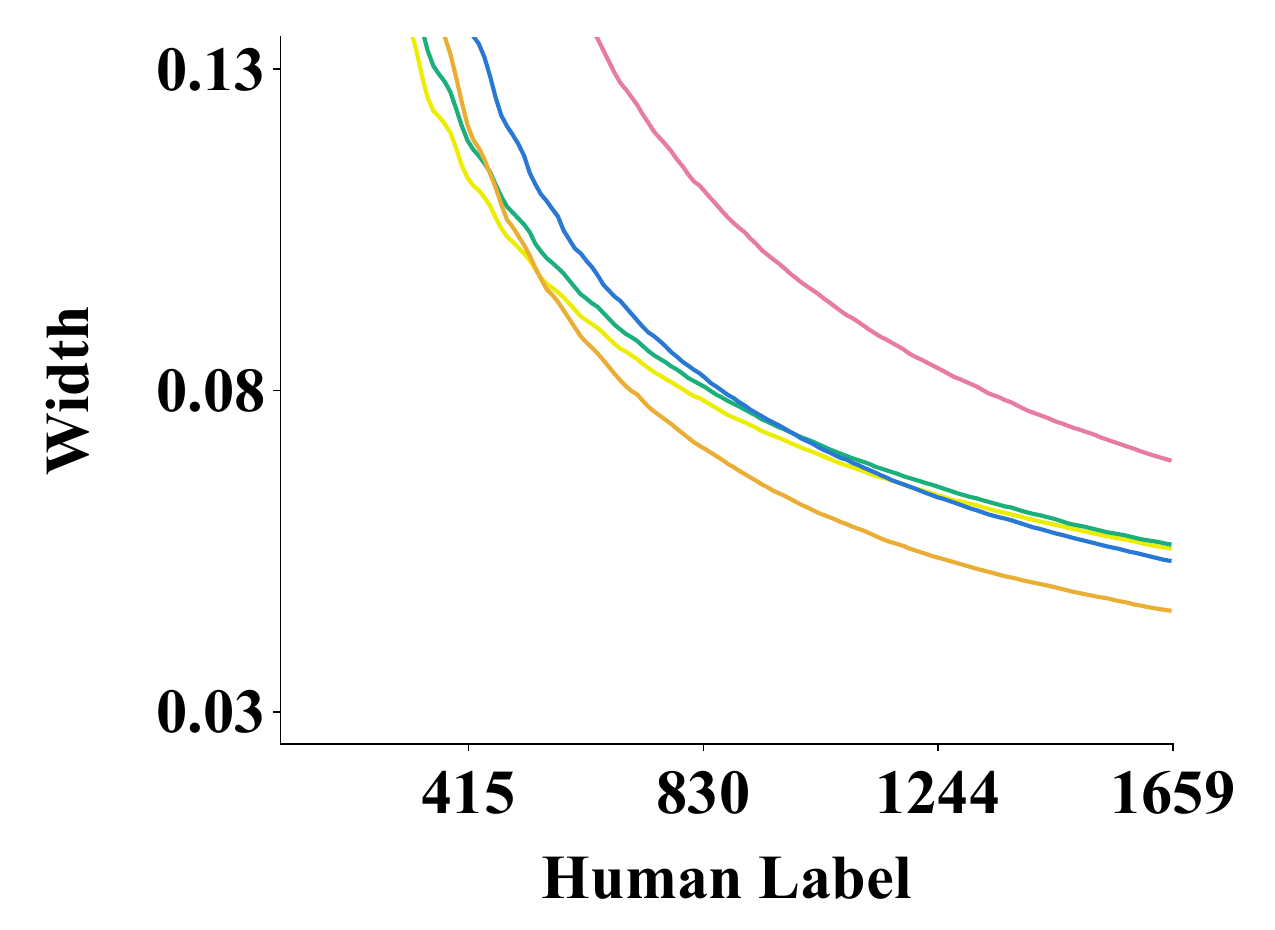}
}
\caption{Anytime-valid inference on AllSides with human budget $\rho=0.2$.}
\label{figure:3}
\end{figure*}

Figure~\ref{figure:3} shows that AAELI maintains anytime coverage above the nominal $95\%$ level.
It is also the most label-efficient, and its advantage grows as the target width becomes more stringent: to reach $\operatorname{Width}=0.10$, AAELI needs on average 483 human labels, whereas the strongest baseline, RASI, needs 566, a saving of 14.7\%.
This saving increases to 18.5\%, 20.2\%, 20.3\%, and 23.1\% at $\operatorname{Width}=0.09$, $0.08$, $0.07$, and $0.06$, respectively; at $\operatorname{Width}=0.06$, AAELI needs 1{,}032 labels compared with 1{,}342 for RASI.
Moreover, AAELI attains the narrowest width at the end of the stream, about $0.044$, compared with $0.049$ for RASI and $0.056$--$0.078$ for the other baselines.
With GPT-6-Luna as the only LLM judge (Figure~\ref{figure:allsides1}), AAELI requires the fewest human labels.

\subsection{Application to Misinformation Prevalence Estimation}
\label{sec:auto:6.3}
Next, we consider estimating the prevalence of misinformation in news headlines.
We use the Misinfo dataset \citep{gabriel2022misinfo}, combining its validation and test sets.
Let $H_t \in \{\textit{misinfo}, \textit{real}\}$ denote the human-annotated label of headline $t$, and define $Y_t = \mathbf{1}\{H_t = \textit{misinfo}\}$.
The target is the prevalence of misinformation, $\mu_0 = \mathbb{E}[Y_t] = \mathbb{P}(H_t = \textit{misinfo})$.
We consider three low-cost LLM judges, namely GPT-6-Luna \citep{openai2026gpt6}, Claude-Haiku-5.5 \citep{anthropic2026haiku55}, and DeepSeek-V4.1-Flash-1017 \citep{xu2026deepseek}.
Under the resampling protocol, $\mu_0$ is the mean of $Y_t$ over the full corpus, which gives $\mu_0 = 0.55$.

\begin{figure*}[htbp]
\centering
\includegraphics[width=0.7\textwidth]{Figure/Legend.pdf}
\subfloat[Confidence Sequences]{%
    \includegraphics[width=0.34\textwidth]{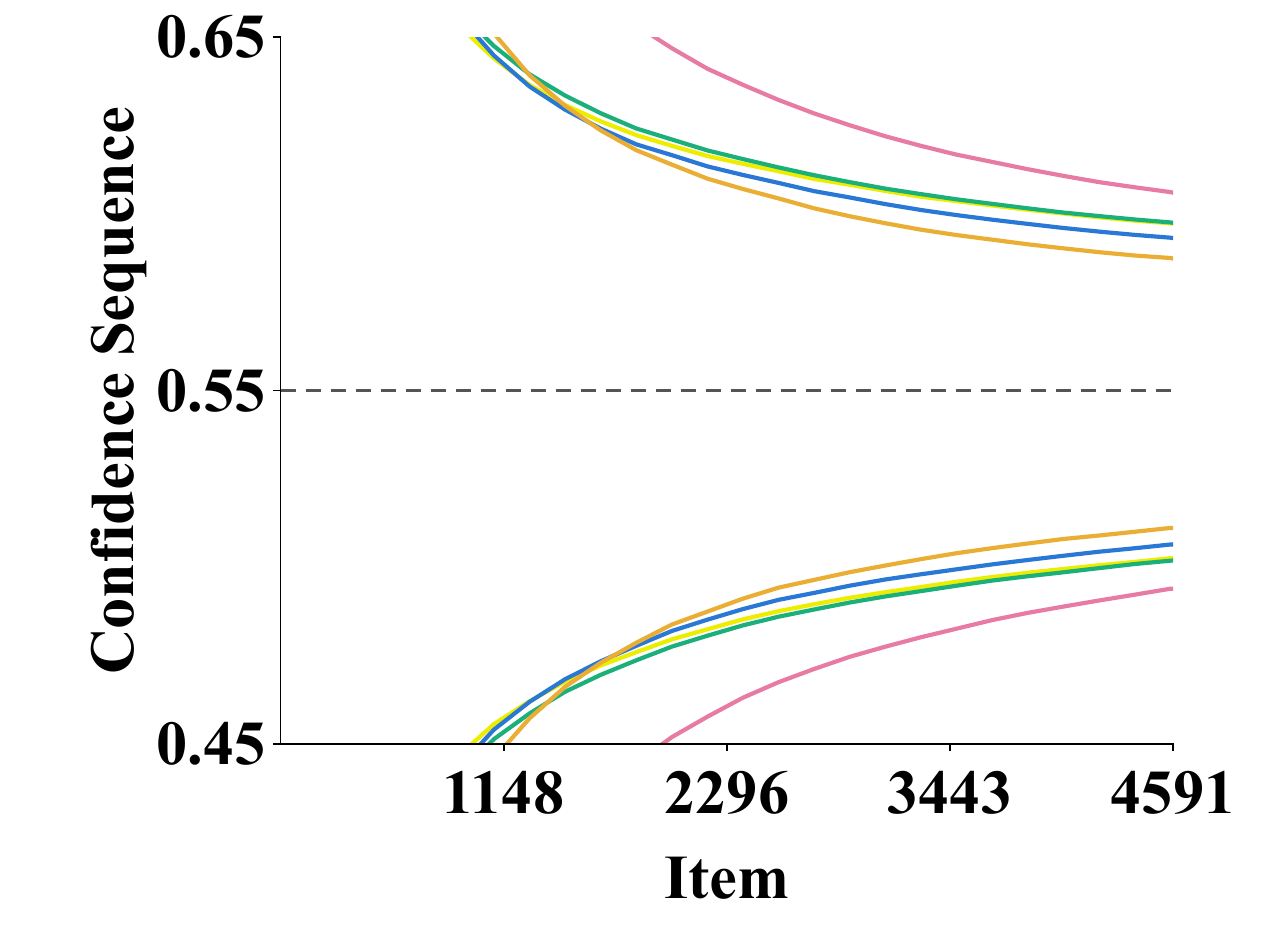}
}
\subfloat[Anytime Coverage]{%
    \includegraphics[width=0.34\textwidth]{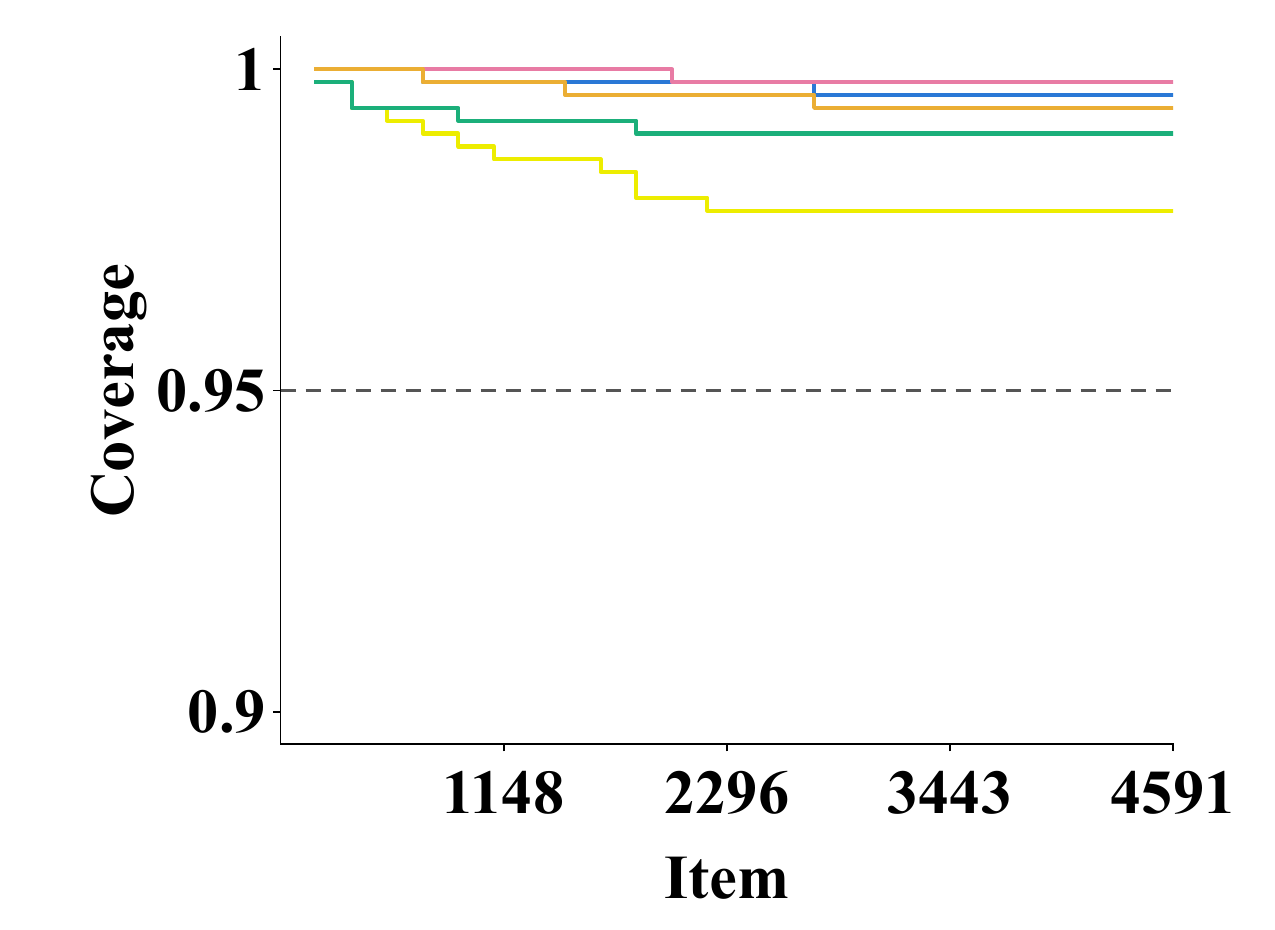}
}
\subfloat[Width]{%
    \includegraphics[width=0.34\textwidth]{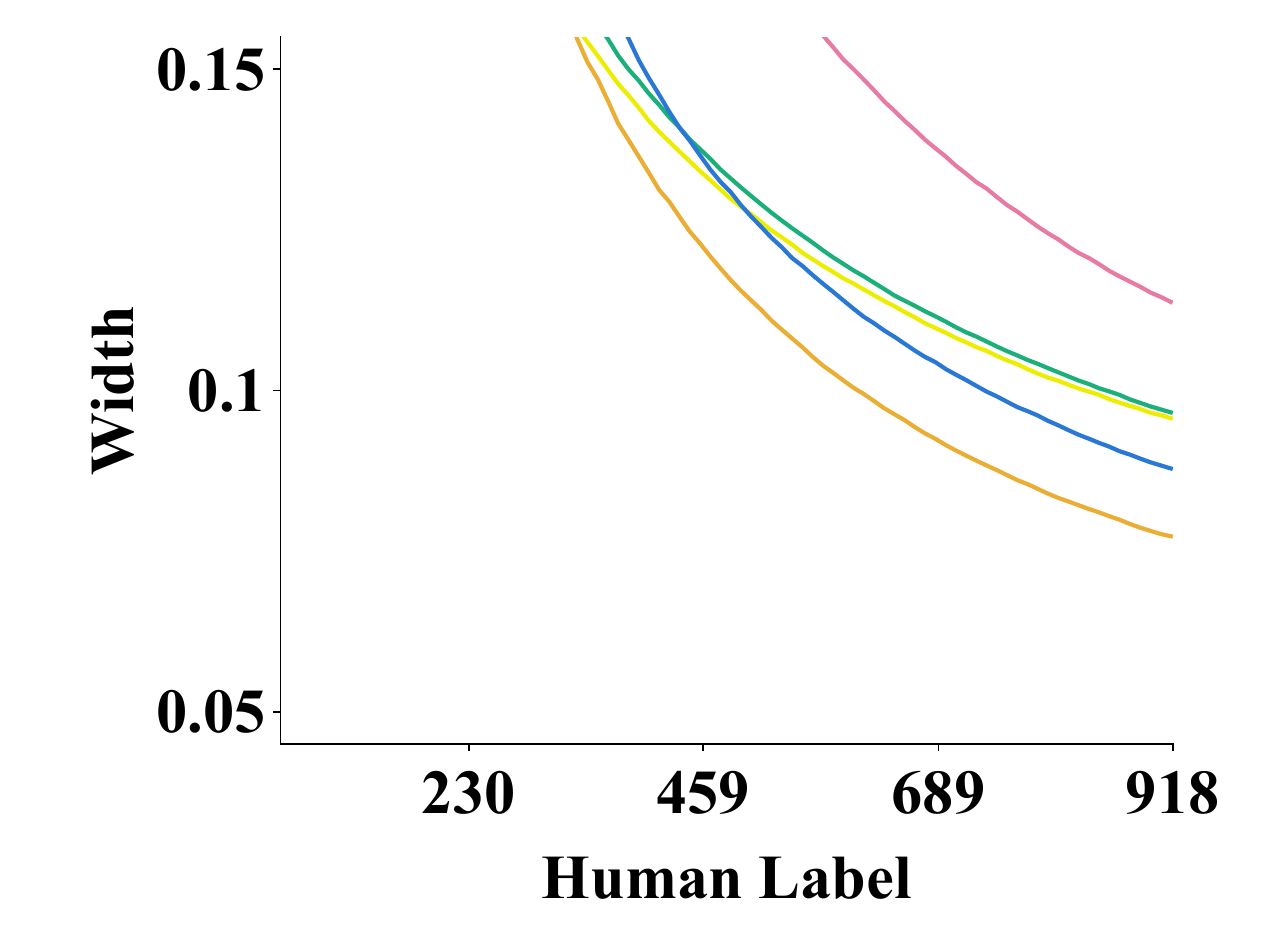}
}
\caption{Anytime-valid inference on Misinfo with human budget $\rho=0.2$.}
\label{figure:4}
\end{figure*}

Figure~\ref{figure:4} shows the results on the Misinfo data.
All methods maintain anytime coverage above the nominal $95\%$ level throughout the stream.
AAELI requires the fewest human labels, saving 17.0\% relative to the strongest baseline at $\operatorname{Width}=0.10$ (616 versus 742 for RASI).
This advantage grows as the target width becomes more stringent, increasing from 9.0\% at $\operatorname{Width}=0.14$ to 13.7\%, 15.3\%, and 16.4\% at $\operatorname{Width}=0.13$, $0.12$, and $0.11$, respectively.
Moreover, AAELI is the only method whose width falls below $0.08$ within the budget, reaching about $0.076$ at the end of the stream, compared with $0.087$--$0.112$ for the baselines.
Figure~\ref{figure:misinfo1} further shows that AAELI remains the most label-efficient method even with GPT-6-Luna as the sole LLM judge.

\section{Conclusion}
\label{sec:auto:7}
\label{sec:conclusion}
In this article, we have developed AAELI, a method that uses LLM judgments to guide adaptive human auditing and constructs anytime-valid confidence sequences for the population mean of human labels. 
Our contributions are threefold. 
First, we construct estimating equations generated by the randomized audit design and develop empirical-likelihood theory for the resulting martingale estimating equations, so that inference remains valid without any unbiasedness, calibration, or independence assumption on the LLM judges. 
Second, mixing the empirical-likelihood dual objective over its multipliers yields exactly anytime-valid predictable empirical-likelihood $e$-processes.
The vector version recovers the raw empirical-likelihood statistic with a $q\log n$ correction, while predictable projection preserves the first-order profile empirical-likelihood statistic with only a $\log n$ correction, so that profiling corresponds to dimension reduction in predictable betting.
The same efficient variance governs local evidence accumulation for the continuous-mixture construction.
Third, adapting Neyman-type allocation and AIPW efficiency theory, we show that the judges affect only efficiency, through the information they add to the pre-audit variables and through how well the auxiliary vector represents the conditional human mean.
The oracle audit probability is proportional to the conditional standard deviation of the human label and no history-dependent rule attains a smaller asymptotic variance for augmented estimating equations with the same audit-probability floor and budget.
This rule can be learned from selectively audited data without sample splitting.
Empirically, AAELI reduces the number of human labels needed to reach a target width by roughly 10\%--25\% relative to the strongest baseline.

Looking ahead, the AAELI framework can be extended in several directions.
First, the estimating equations can accommodate targets beyond the population mean, such as quantiles of human scores and multiway model ranking.
Second, the current analysis assumes a stationary evaluation population and noise-free human labels. 
Handling distribution shift over time and noisy or inconsistent human annotations would broaden its applicability.
Third, the current analysis treats LLM judgments as freely available for every instance.
Incorporating heterogeneous costs for querying multiple LLM judges would lead to a joint allocation of LLM queries and human audits.


\clearpage
\bibliographystyle{abbrvnat}
\bibliography{reference}

\clearpage
\appendix
\etocdepthtag.toc{mtappendix}
\etocsettagdepth{mtmain}{none}
\etocsettagdepth{mtappendix}{subsection}
\renewcommand{\contentsname}{Appendix}
\tableofcontents
\setcounter{figure}{0}
\setcounter{table}{0}
\setcounter{equation}{0}
\renewcommand{\thefigure}{S\arabic{figure}}
\renewcommand{\thetable}{S\arabic{table}}
\renewcommand{\theequation}{S\arabic{equation}}
\section{Proofs}
\label{app:section:A}

\subsection{Proof of Proposition~\ref{prop:estimating-equations}}
\label{app:subsection:A.1}

\begin{lemma}
\label{lem:standing}
Under Condition~\ref{cond:basic} \textup{(a)--(c)}, the following statements hold for every $t\ge1$:
\begin{enumerate}
\renewcommand{\labelenumi}{(\roman{enumi})}
\item
$(Z_t,Y_t)$ is independent of $\mathcal F_{t-1}$.

\item $\E[R_t\mid\mathcal P_t,Y_t]=\pi_t$ and $\E[R_t\mid\mathcal P_t]=\pi_t$.

\item
For every $\mathcal P_t$-measurable $h_t$ satisfying $\E\|h_t\|<\infty$, the vector $G_t(\mu)$ is $\mathcal F_t$-measurable and integrable for every fixed $\mu\in\R$.
\end{enumerate}
If Condition~\ref{cond:basic} \textup{(d)} also holds, then there is a deterministic constant $C_G<\infty$ such that
\[
\sup_{t\ge1}\sup_{\mu\in\Theta}\|G_t(\mu)\|\le C_G \qquad\text{almost surely}.
\]
\end{lemma}

\begin{proof}[Proof of Lemma~\ref{lem:standing}]

For \textup{(i)}, define $\mathcal H_t=\sigma\bigl(\mathcal F_0,\{Z_s,Y_s,U_s:1\le s\le t\}\bigr)$ and $\mathcal H_0=\mathcal F_0.$
We show inductively that $\mathcal F_t\subseteq\mathcal H_t$.
If $\mathcal F_{t-1}\subseteq\mathcal H_{t-1}$, then $\mathcal P_t\subseteq\sigma(\mathcal H_{t-1},Z_t)$.
Thus $\pi_t$, $R_t=\mathbf{1}\{U_t\le\pi_t\}$, and $R_tY_t$ are $\mathcal H_t$-measurable, which proves the inclusion.
By the i.i.d.\ sampling assumption and independence of the auxiliary randomization, $(Z_t,Y_t)$ is independent of $\mathcal H_{t-1}$ and hence of $\mathcal F_{t-1}$.

For \textup{(ii)}, $U_t$ is independent of $\sigma(\mathcal H_{t-1},Z_t,Y_t)$ and therefore of $\sigma(\mathcal P_t,Y_t)$.
Since $\pi_t$ is $\mathcal P_t$-measurable,
\[
\E[R_t\mid\mathcal P_t,Y_t]=\Pp(U_t\le\pi_t\mid\mathcal P_t,Y_t)=\pi_t.
\]
The tower property then gives
\[
\E[R_t\mid\mathcal P_t]=\E\!\left[\E[R_t\mid\mathcal P_t,Y_t]\middle|\mathcal P_t\right]=\pi_t.
\]

For \textup{(iii)}, write $A_t(\mu)=\frac{R_tY_t-\mu R_t}{\pi_t}$ and $B_t=\left(\frac{R_t}{\pi_t}-1\right)h_t$.
Since $\mathcal P_t\subseteq\mathcal F_t$, both components are $\mathcal F_t$-measurable.
Moreover,
\[
|A_t(\mu)|\le\frac{1+|\mu|}{\underline\pi},
\qquad
\|B_t\|\le(1+\underline\pi^{-1})\|h_t\|,
\]
which proves integrability.

Finally, if \textup{(d)} holds, then $\Theta\subseteq[0,1]$ implies $|A_t(\mu)|\le\underline\pi^{-1}$ uniformly in $t$ and $\mu\in\Theta$.
Consequently, one may take
\[
C_G=\left\{\underline\pi^{-2}+(1+\underline\pi^{-1})^2C_h^2\right\}^{1/2}.
\]
\end{proof}

\begin{proof}[Proof of Proposition~\ref{prop:estimating-equations}]
By Lemma~\ref{lem:standing}  \textup{(iii)}, $G_t(\mu)$ is $\mathcal F_t$-measurable and integrable.
Lemma~\ref{lem:standing} \textup{(ii)} gives
\[
\E[A_t(\mu)\mid\mathcal P_t,Y_t]=\frac{Y_t-\mu}{\pi_t}\E[R_t\mid\mathcal P_t,Y_t]=Y_t-\mu.
\]
Using the tower property and the independence established in Lemma~\ref{lem:standing} \textup{(i)}, we obtain
\[
\E[A_t(\mu)\mid\mathcal F_{t-1}]=\E[Y_t-\mu\mid\mathcal F_{t-1}]=\mu_0-\mu.
\]
Likewise, since $h_t$ and $\pi_t$ are $\mathcal P_t$-measurable, Lemma~\ref{lem:standing}  \textup{(ii)} implies
\[
\E[B_t\mid\mathcal P_t]=h_t\left\{\frac{\E[R_t\mid\mathcal P_t]}{\pi_t}-1\right\}=0_r.
\]
Another application of the tower property yields $\E[B_t\mid\mathcal F_{t-1}]=0_r$.
Combining the two components gives
\[
\E[G_t(\mu)\mid\mathcal F_{t-1}]=\begin{pmatrix}\mu_0-\mu\\0_r\end{pmatrix}.
\]
This vector equals zero if and only if $\mu=\mu_0$, so the conditional moment equation uniquely identifies $\mu_0$.
At $\mu_0$, integrability, adaptedness, and the conditional mean-zero identity show that $\{G_t(\mu_0)\}_{t\ge1}$ is a martingale difference sequence with respect to $(\mathcal F_t)_{t\ge0}$.
\end{proof}

\subsection{Proof of Theorem~\ref{thm:adaptive-wilks}}
\label{app:subsection:A.2}
\label{app:adaptive-el-proofs}

Throughout this subsection, assume Conditions~4.1 and~4.2.
Write $q=r+1$, let $e_1$ be the first standard basis vector of $\R^q$, and set $D=-e_1$.
Define $W_n=\frac1n\sum_{t=1}^n\frac{R_t}{\pi_t}$, $
\mathbb G_n=\frac1{\sqrt n}\sum_{t=1}^nG_t(\mu_0)$, $
\overline G_n(\mu)=\frac1n\sum_{t=1}^nG_t(\mu)$, $
\widehat\Sigma_n(\mu)=\frac1n\sum_{t=1}^nG_t(\mu)G_t(\mu)^\top.$
For $u\in\R$, write $\mu_n(u)=\mu_0+u/\sqrt n$.
All matrix norms below are operator norms.
By Lemma~\ref{lem:standing}, there is a deterministic constant $C_G<\infty$ such that
\[
\sup_{t\ge1}\sup_{\mu\in\Theta}\|G_t(\mu)\|\le C_G
\qquad\text{almost surely}.
\]

\begin{lemma}
\label{lem:appendix-martingale-limits}
Let $W_n\xrightarrow{p}1$, $\mathbb G_n\xrightarrow{d} N(0,\Omega)$ and $\widehat\Sigma_n(\mu_0)\xrightarrow{p}\Omega.$
Moreover, for every fixed $M<\infty$,
\[
\sup_{|u|\le M}\left\|\sqrt n\,\overline G_n(\mu_n(u))-\mathbb G_n-Du\right\|\xrightarrow{p}0
\]
and
\[
\sup_{|u|\le M}\left\|\widehat\Sigma_n(\mu_n(u))-\Omega\right\|\xrightarrow{p}0.
\]
\end{lemma}

\begin{proof}[Proof of Lemma~\ref{lem:appendix-martingale-limits}]
By Lemma~\ref{lem:standing}\textup{(ii)} and the tower property, $\E\left[\frac{R_t}{\pi_t}-1\middle|\mathcal F_{t-1}\right]=0.$
The summands are uniformly bounded.
Orthogonality of martingale differences therefore gives $\E[(W_n-1)^2]\le\frac{C}{n}$ for a deterministic constant $C<\infty$. 
Hence $W_n\xrightarrow{p}1$.

By Proposition~\ref{prop:estimating-equations}, $G_t(\mu_0)$ is a martingale difference sequence.
Its deterministic uniform bound implies the conditional Lindeberg condition for the normalized sums.
Condition~\ref{cond:cov} and the martingale central limit theorem consequently yield $\mathbb G_n\xrightarrow{d} N(0,\Omega).$

For the second moments, each entry of
\[
G_t(\mu_0)G_t(\mu_0)^\top
-\E[G_t(\mu_0)G_t(\mu_0)^\top
\mid\mathcal F_{t-1}]
\]
is a uniformly bounded martingale difference.
Its average converges to zero in $L^2$.
Since $q$ is fixed, Condition~\ref{cond:cov} implies $\widehat\Sigma_n(\mu_0)\xrightarrow{p}\Omega.$

For the local expansions, the definition of $G_t$ gives the exact identity
\[
G_t(\mu_n(u))=G_t(\mu_0)-\frac{u}{\sqrt n}\frac{R_t}{\pi_t}e_1.
\]
Consequently,
\[
\sqrt n\,\overline G_n(\mu_n(u))=\mathbb G_n-uW_ne_1=\mathbb G_n+Du-u(W_n-1)e_1.
\]
Taking the supremum over $|u|\le M$ proves the first uniform expansion.

For the second expansion, put $d_t(u)=-\frac{u}{\sqrt n}\frac{R_t}{\pi_t}e_1.$
Then $\sup_{|u|\le M}\|d_t(u)\|\le\frac{M}{\underline\pi\sqrt n}.$
Expanding the outer product and using $\|G_t(\mu_0)\|\le C_G$ gives
\[
\sup_{|u|\le M}\left\|\widehat\Sigma_n(\mu_n(u))-\widehat\Sigma_n(\mu_0)\right\|\le\frac{2C_GM}{\underline\pi\sqrt n}+\frac{M^2}{\underline\pi^2n}.
\]
Combining this deterministic bound with $\widehat\Sigma_n(\mu_0)\xrightarrow{p}\Omega$ proves the result.
\end{proof}

\begin{lemma}
\label{lem:local-convex-hull}
For every fixed $M<\infty$,
\[
\Pp\left(0\in\operatorname{int}\operatorname{conv}\{G_t(\mu_n(u)):1\le t\le n\}\text{ for every }|u|\le M\right)\to1.
\]
Consequently, uniformly for $|u|\le M$, the empirical-likelihood constraints are feasible and the dual solution exists uniquely in the interior of its domain with probability tending to one.
\end{lemma}

\begin{proof}[Proof of Lemma~\ref{lem:local-convex-hull}]
Fix $M<\infty$.
Since $\mu_0$ lies in the interior of $\Theta$, $\mu_n(u)\in\Theta$ for every $|u|\le M$ and all sufficiently large $n$.
Lemma~\ref{lem:appendix-martingale-limits} gives
\[
\sup_{|u|\le M}\|\overline G_n(\mu_n(u))\|=O_p(n^{-1/2})
\]
and
\[
\sup_{|u|\le M}\|\widehat\Sigma_n(\mu_n(u))-\Omega\|\xrightarrow{p}0.
\]
Choose $c\in(0,\lambda_{\min}(\Omega))$. 
With probability tending to one,
\[
\inf_{|u|\le M}\lambda_{\min}\{\widehat\Sigma_n(\mu_n(u))\}\ge c,\qquad C_G\sup_{|u|\le M}\|\overline G_n(\mu_n(u))\|<c.
\]

On this event, suppose that zero is not in the interior of the convex hull for some $|u|\le M$.
The separating-hyperplane theorem gives a unit vector $v$ such that
\[
v^\top G_t(\mu_n(u))\ge0\qquad\text{for every }1\le t\le n.
\]
Since $v^\top G_t(\mu_n(u))\le C_G$,
\[
\{v^\top G_t(\mu_n(u))\}^2\le C_Gv^\top G_t(\mu_n(u)).
\]
Averaging yields
\[
c\le v^\top\widehat\Sigma_n(\mu_n(u))v\le C_Gv^\top\overline G_n(\mu_n(u))\le C_G\|\overline G_n(\mu_n(u))\|<c,
\]
a contradiction.

Thus zero lies in the interior of every stated convex hull on an event whose probability tends to one.
For each such $u$, there exist strictly positive feasible weights.
The empirical-likelihood maximum is therefore positive, and its maximizing weights are strictly positive.
The Lagrange multiplier equations give
\[
p_t=
\frac{1}{n\{1+\lambda_n(u)^\top G_t(\mu_n(u))\}},
\]
with all denominators positive.
Here the normalization multiplier equals $n$, as follows by multiplying the first-order equations by $p_t$ and summing over $t$.

Hence $\lambda_n(u)$ is an interior dual solution.
Positive definiteness of $\widehat\Sigma_n(\mu_n(u))$ implies that the vectors $G_t(\mu_n(u))$ span $\R^q$.
The dual log criterion is therefore strictly concave on its domain, so the solution is unique.
\end{proof}

\begin{lemma}
\label{lem:local-el}
For every fixed $M<\infty$,
\[
\sup_{|u|\le M}\left|-2\log\mathcal R_n(\mu_n(u))-(\mathbb G_n+Du)^\top\Omega^{-1}(\mathbb G_n+Du)\right|\xrightarrow{p}0.
\]
\end{lemma}

\begin{proof}[Proof of Lemma~\ref{lem:local-el}]
Fix $M<\infty$ and write
\[
g_t(u)=G_t(\mu_n(u)),\qquad\overline g_n(u)=\overline G_n(\mu_n(u)).
\]
By Lemmas~\ref{lem:appendix-martingale-limits} and~\ref{lem:local-convex-hull}, with probability tending to one the unique interior dual solution $\lambda_n(u)$ exists for every $|u|\le M$, and
\[
\sup_{|u|\le M}\|\sqrt n\,\overline g_n(u)-\mathbb G_n-Du\|\xrightarrow{p}0,\qquad\sup_{|u|\le M}\|\widehat\Sigma_n(\mu_n(u))-\Omega\|\xrightarrow{p}0.
\]
Also, $\|g_t(u)\|\le C_G$ uniformly for all sufficiently large $n$.

We first establish
\[
\sup_{|u|\le M}\|\lambda_n(u)\|=O_p(n^{-1/2}).
\]
The dual equation is
\[
0=\frac1n\sum_{t=1}^n\frac{g_t(u)}{1+\lambda_n(u)^\top g_t(u)}.
\]
For $\lambda_n(u)\ne0$, write $\lambda_n(u)=av$, where $a=\|\lambda_n(u)\|$ and $\|v\|=1$.
Using $(1+x)^{-1}=1-x/(1+x)$ and taking the inner product with $v$ gives
\[
v^\top\overline g_n(u)=\frac a n\sum_{t=1}^n\frac{\{v^\top g_t(u)\}^2}{1+av^\top g_t(u)}\ge\frac{a\,v^\top\widehat\Sigma_n(\mu_n(u))v}{1+aC_G}.
\]
Therefore,
\[
a\left[\lambda_{\min}\{\widehat\Sigma_n(\mu_n(u))\}-C_G\|\overline g_n(u)\|\right]\le\|\overline g_n(u)\|.
\]
The expression in brackets is bounded away from zero uniformly over $|u|\le M$ with probability tending to one, whereas $\sup_{|u|\le M}\|\overline g_n(u)\|=O_p(n^{-1/2})$.
This proves the asserted bound, including the case $\lambda_n(u)=0$.

It follows that
\[
\sup_{|u|\le M}\max_{t\le n}|\lambda_n(u)^\top g_t(u)|=O_p(n^{-1/2}).
\]
In particular, with probability tending to one,
\[
\inf_{|u|\le M}\min_{t\le n}\{1+\lambda_n(u)^\top g_t(u)\}\ge\frac12.
\]
Expanding the dual equation using $(1+x)^{-1}=1-x+x^2/(1+x)$ yields
\[
0=\overline g_n(u)-\widehat\Sigma_n(\mu_n(u))\lambda_n(u)+\mathfrak r_n(u),
\]
where
\[
\mathfrak r_n(u)=\frac1n\sum_{t=1}^n\frac{g_t(u)\{\lambda_n(u)^\top g_t(u)\}^2}{1+\lambda_n(u)^\top g_t(u)}.
\]
The uniform denominator bound implies
\[
\sup_{|u|\le M}\|\mathfrak r_n(u)\|
\le
2C_G^3\sup_{|u|\le M}\|\lambda_n(u)\|^2=O_p(n^{-1}).
\]
Since the inverse second-moment matrices are uniformly bounded in probability, the quantities
\[
\delta_n(u)=\lambda_n(u)-\widehat\Sigma_n(\mu_n(u))^{-1}\overline g_n(u)
\]
satisfy $\sup_{|u|\le M}\|\delta_n(u)\|=O_p(n^{-1}).$

The dual representation and Taylor expansion give, uniformly for $|u|\le M$,
\[
\begin{aligned}
-2\log\mathcal R_n(\mu_n(u))
&=2\sum_{t=1}^n\log\{1+\lambda_n(u)^\top g_t(u)\}\\
&=2n\lambda_n(u)^\top\overline g_n(u)-n\lambda_n(u)^\top\widehat\Sigma_n(\mu_n(u))\lambda_n(u)+o_p(1).
\end{aligned}
\]
Indeed, the summed cubic remainder is bounded by a constant times
\[
nC_G^3\sup_{|u|\le M}\|\lambda_n(u)\|^3=O_p(n^{-1/2})=o_p(1).
\]
By the definition of $\delta_n(u)$,
\[
\begin{aligned}
&2n\lambda_n(u)^\top\overline g_n(u)-n\lambda_n(u)^\top\widehat\Sigma_n(\mu_n(u))\lambda_n(u)\\&=n\overline g_n(u)^\top\widehat\Sigma_n(\mu_n(u))^{-1}\overline g_n(u)-n\delta_n(u)^\top\widehat\Sigma_n(\mu_n(u))\delta_n(u).
\end{aligned}
\]
The final term is uniformly $O_p(n^{-1})$.
Consequently,
\[
-2\log\mathcal R_n(\mu_n(u))=n\overline g_n(u)^\top\widehat\Sigma_n(\mu_n(u))^{-1}\overline g_n(u)+o_p(1)
\]
uniformly for $|u|\le M$.
Substituting the uniform local mean and second-moment expansions proves the result.
\end{proof}

\begin{remark}
\label{rem:argmax-existence}
For each fixed $n$, the map $\mu\mapsto\mathcal R_n(\mu)$ is upper semicontinuous on $\Theta$.
Indeed, write $\mathbb S_n=\{p\in[0,1]^n:\sum_tp_t=1\}$ and $\Gamma=\{(p,\mu)\in\mathbb S_n\times\Theta:\sum_tp_tG_t(\mu)=0\}$.
Because $G_t(\mu)$ is affine in $\mu$, the set $\Gamma$ is closed, and $\mathbb S_n$ is compact.
The objective $p\mapsto\prod_tnp_t$ is continuous, so its maximum over the closed-graph, compact-valued correspondence $\mu\mapsto\{p:(p,\mu)\in\Gamma\}$ is upper semicontinuous in $\mu$, with the convention $\max\emptyset=0$.
Hence $\arg\max_{\mu\in\Theta}\mathcal R_n(\mu)$ is a nonempty closed subset of $\Theta$.
Its smallest element is a measurable choice of $\widehat\mu_n$.
\end{remark}

\begin{lemma}
\label{lem:localization}
Every measurable choice $\widehat\mu_n\in\arg\max_{\mu\in\Theta}\mathcal R_n(\mu)$ satisfies
\[
\sqrt n(\widehat\mu_n-\mu_0)=O_p(1).
\]
\end{lemma}

\begin{proof}[Proof of Lemma~\ref{lem:localization}]
Let $\overline A_n(\mu)=\frac1n\sum_{t=1}^nA_t(\mu).$
Since $\Theta\subseteq[0,1]$ and $\pi_t\ge\underline\pi$, setting $C_A=\underline\pi^{-1}$ gives
\[
\sup_{t\ge1}\sup_{\mu\in\Theta}|A_t(\mu)|\le C_A.
\]

For any $\mu$ with $\mathcal R_n(\mu)>0$, consider strictly positive feasible weights $(p_t)$.
For every $s$ such that $1+sA_t(\mu)>0$ for all $t$, the empirical-likelihood constraints imply
\[
\sum_{t=1}^np_t\{1+sA_t(\mu)\}=1.
\]
The arithmetic--geometric mean inequality gives
\[
\prod_{t=1}^n\left[np_t\{1+sA_t(\mu)\}\right]\le1.
\]
Maximizing over feasible weights therefore yields
\[
-2\log\mathcal R_n(\mu)\ge2\sum_{t=1}^n\log\{1+sA_t(\mu)\}.
\]

Choose $s=\frac{\overline A_n(\mu)}{2C_A^2}.$
Then $|sA_t(\mu)|\le1/2$.
Using $\log(1+x)\ge x-x^2$ for $|x|\le1/2$,
\[
\begin{aligned}
-2\log\mathcal R_n(\mu)
&\ge2ns\overline A_n(\mu)-2s^2\sum_{t=1}^nA_t(\mu)^2\\
&\ge2ns\overline A_n(\mu)-2ns^2C_A^2\\
&=\frac{n\overline A_n(\mu)^2}{2C_A^2}.
\end{aligned}
\]
When $\mathcal R_n(\mu)=0$, the same bound holds with the convention $-\log0=+\infty$.

By Lemma~\ref{lem:local-el} at $u=0$,
\[
-2\log\mathcal R_n(\mu_0)=\mathbb G_n^\top\Omega^{-1}\mathbb G_n+o_p(1)=O_p(1),
\]
and $\mathcal R_n(\mu_0)>0$ with probability tending to one.
Since $\widehat\mu_n$ maximizes $\mathcal R_n$,
\[
\frac{n\overline A_n(\widehat\mu_n)^2}{2C_A^2}\le-2\log\mathcal R_n(\widehat\mu_n)\le-2\log\mathcal R_n(\mu_0)=O_p(1).
\]
Hence $\overline A_n(\widehat\mu_n)=O_p(n^{-1/2})$.

Finally, the exact identity
\[
\overline A_n(\mu)=\overline A_n(\mu_0)-(\mu-\mu_0)W_n
\]
implies, on the event $W_n>0$,
\[
\widehat\mu_n-\mu_0=\frac{\overline A_n(\mu_0)-\overline A_n(\widehat\mu_n)}{W_n}.
\]
Lemma~\ref{lem:appendix-martingale-limits} gives $W_n\xrightarrow{p}1$ and
\[
\sqrt n\,\overline A_n(\mu_0)=e_1^\top\mathbb G_n=O_p(1).
\]
The claimed localization follows.
\end{proof}

\begin{proof}[Proof of Theorem~\ref{thm:adaptive-wilks}] 
Lemma~\ref{lem:local-convex-hull} establishes feasibility throughout every fixed $n^{-1/2}$-neighborhood of $\mu_0$ with probability tending to one.
By Lemma~\ref{lem:localization},
\[
\widehat u_n=\sqrt n(\widehat\mu_n-\mu_0)=O_p(1).
\]
Define $\mathcal I=D^\top\Omega^{-1}D>0$ and $\Lambda_n(u)=(\mathbb G_n+Du)^\top\Omega^{-1}(\mathbb G_n+Du).$
Lemma~\ref{lem:local-el} gives
\[
-2\log\mathcal R_n(\mu_n(u))=\Lambda_n(u)+o_p(1)
\]
uniformly for $u$ in compact subsets of $\R$.
The unique minimizer of $\Lambda_n$ is
\[
u_n^*=-\mathcal I^{-1}D^\top\Omega^{-1}\mathbb G_n,
\]
and completing the square yields
\[
\Lambda_n(u)=\Lambda_n(u_n^*)+\mathcal I(u-u_n^*)^2.
\]
By Lemma~\ref{lem:appendix-martingale-limits}, $\mathbb G_n=O_p(1)$ and hence $u_n^*=O_p(1)$.
Since $\mu_0$ is an interior point of $\Theta$, $\mu_n(u_n^*)\in\Theta$ with probability tending to one.

Both $\widehat u_n$ and $u_n^*$ are tight.
The uniform local expansion can therefore be evaluated at these random arguments by first restricting them to a fixed compact set and then letting that set increase.
Since $\widehat\mu_n$ maximizes $\mathcal R_n$,
\[
\Lambda_n(\widehat u_n)
\le\Lambda_n(u_n^*)+o_p(1).
\]
The completed-square identity implies
\[
\mathcal I(\widehat u_n-u_n^*)^2=o_p(1).
\]
Consequently,
\[
\sqrt n(\widehat\mu_n-\mu_0)=-\mathcal I^{-1}D^\top\Omega^{-1}\mathbb G_n+o_p(1).
\]

Write $b^*=\Omega_B^{-1}\gamma$ and $ V_{\EL}=\sigma_A^2-\gamma^\top\Omega_B^{-1}\gamma.$
Positive definiteness of $\Omega$ implies $V_{\EL}>0$.
Since $D=-e_1$, the block inverse formula gives $\mathcal I=V_{\EL}^{-1}$ and $-\mathcal I^{-1}D^\top\Omega^{-1}=\begin{pmatrix}1&-(b^*)^\top\end{pmatrix}.$
Thus
\[
\sqrt n(\widehat\mu_n-\mu_0)=\frac1{\sqrt n}\sum_{t=1}^n\{A_t(\mu_0)-(b^*)^\top B_t\}+o_p(1).
\]
By Lemma~\ref{lem:appendix-martingale-limits}, $\mathbb G_n\xrightarrow{d} N(0,\Omega)$.
The limiting variance is
\[
\mathcal I^{-2}D^\top\Omega^{-1}\Omega\Omega^{-1}D=\mathcal I^{-1}=V_{\EL},
\]
which proves
\[
\sqrt n(\widehat\mu_n-\mu_0) \xrightarrow{d}N(0,V_{\EL}).
\]

Finally, the local expansion and $\widehat u_n-u_n^*=o_p(1)$ imply
\[
\begin{aligned}
-2\log\frac{\mathcal R_n(\mu_0)}{\mathcal R_n(\widehat\mu_n)}
&=\Lambda_n(0)-\Lambda_n(\widehat u_n)+o_p(1)\\
&=\Lambda_n(0)-\Lambda_n(u_n^*)+o_p(1)\\
&=\frac{(D^\top\Omega^{-1}\mathbb G_n)^2}{D^\top\Omega^{-1}D}+o_p(1).
\end{aligned}
\]
Since
\[
\frac{D^\top\Omega^{-1}\mathbb G_n}{\sqrt{D^\top\Omega^{-1}D}}\xrightarrow{d} N(0,1),
\]
the profile empirical-likelihood statistic converges in distribution to $\chi_1^2$.
\end{proof}

\subsection{Proof of Theorem~\ref{thm:eff-decomp}}
\label{app:subsection:A.3}
\label{app:efficiency-calculations}

\begin{proof}[Proof of Theorem~\ref{thm:eff-decomp}]
Consider the fixed-design setting in Section~3.2.
Write $m(Z)=\E[Y\mid Z]$, $\sigma^2(Z)=\Var(Y\mid Z)$, $\bar m(Z)=m(Z)-\mu_0$ and $w(Z)=\frac{1-\pi(Z)}{\pi(Z)}.$
Recall that $A=\frac{R}{\pi(Z)}(Y-\mu_0)$ and $B=\left(\frac{R}{\pi(Z)}-1\right)h(Z).$
Since $Y$ and $h$ are bounded and $\pi(Z)\ge\underline\pi>0$, both $A$ and $B$ have finite second moments.
Under the fixed design and independent audit randomization, the estimating vectors are i.i.d.
Their covariance matrix is
\[
\Omega=
\begin{pmatrix}
\sigma_A^2 & \gamma^\top\\
\gamma & \Omega_B
\end{pmatrix},\qquad\sigma_A^2=\E[A^2],\qquad\gamma=\E[BA],\qquad\Omega_B=\E[BB^\top].
\]
Whenever $\Omega$ is positive definite and the remaining conditions of Theorem~\ref{thm:adaptive-wilks} hold, Condition~\ref{cond:cov} holds with this matrix, and the block-inverse formula identifies
\[
V_{\EL}=\sigma_A^2-\gamma^\top\Omega_B^{-1}\gamma
\]
as the asymptotic variance of the empirical-likelihood estimator.
The variance decomposition below remains valid even when $V_{\EL}=0$.

The audit mechanism gives $\E[R\mid Z,Y]=\pi(Z)$, so $\E[A]=0$ and $\E[B]=0$.
Since $R^2=R$,
\[
\E\left[\frac{R^2}{\pi(Z)^2}\middle|Z,Y\right]=\frac1{\pi(Z)}.
\]
Moreover,
\[
\E\left[\left(\frac{R}{\pi(Z)}-1\right)^2\middle|Z,Y\right]=w(Z)
\]
and
\[
\E\left[\frac{R}{\pi(Z)}\left(\frac{R}{\pi(Z)}-1\right)\middle|Z,Y\right]=w(Z).
\]
Consequently,
\[
\sigma_A^2=\E\left[\frac{(Y-\mu_0)^2}{\pi(Z)}\right],\qquad\Omega_B=\E[w(Z)h(Z)h(Z)^\top].
\]
For the cross covariance, iterated conditional expectation gives
\[
\begin{aligned}
\gamma
&=\E[w(Z)h(Z)(Y-\mu_0)]\\
&=\E\left[w(Z)h(Z)\E[Y-\mu_0\mid Z]\right]\\
&=\E[w(Z)h(Z)\bar m(Z)].
\end{aligned}
\]

For any $b\in\R^r$, define $\psi_b=A-b^\top B$.
Since $\E[\psi_b]=0$,
\[
\Var(\psi_b)=\sigma_A^2-2b^\top\gamma+b^\top\Omega_Bb.
\]
Positive definiteness of $\Omega_B$ implies that the unique minimizing coefficient is
\[
b^*=\Omega_B^{-1}\gamma.
\]
Completing the square yields
\[
\Var(\psi_b)=V_{\EL}+(b-b^*)^\top\Omega_B(b-b^*).
\]
In particular,
\[
V_{\EL}=\min_{b\in\R^r}\Var(\psi_b).
\]

To identify this minimum, define $\phi_{\mathrm{eff}}=\bar m(Z)+\frac{R}{\pi(Z)}\{Y-m(Z)\}.$
By iterated conditional expectation,
\[
\begin{aligned}
\E\left[\frac{R}{\pi(Z)}\{Y-m(Z)\}\middle|Z\right]
&=
\E\left[\frac{Y-m(Z)}{\pi(Z)}\E[R\mid Z,Y]\middle|Z\right]\\
&=\E[Y-m(Z)\mid Z]\\
&=0.
\end{aligned}
\]
Thus the two terms in $\phi_{\mathrm{eff}}$ are uncorrelated, and $\E[\phi_{\mathrm{eff}}]=0$.
Furthermore,
\[
\begin{aligned}
\E\left[\left\{\frac{R}{\pi(Z)}(Y-m(Z))\right\}^2\middle|Z\right]
&=\E\left[\{Y-m(Z)\}^2\E\left[\frac{R^2}{\pi(Z)^2}\middle|Z,Y\right]\middle|Z\right]\\
&=\E\left[\frac{\{Y-m(Z)\}^2}{\pi(Z)}\middle|Z\right]\\
&=\frac{\sigma^2(Z)}{\pi(Z)}.
\end{aligned}
\]
It follows that
\[
\Var(\phi_{\mathrm{eff}})=\Var\{m(Z)\}+\E\left[\frac{\sigma^2(Z)}{\pi(Z)}\right]=V_{\mathrm{eff}}(\pi).
\]

For any $b\in\R^r$, put $d_b(Z)=\bar m(Z)-b^\top h(Z)$.
Direct calculation gives
\[
\psi_b-\phi_{\mathrm{eff}}=\left(\frac{R}{\pi(Z)}-1\right)d_b(Z).
\]
This difference has conditional mean zero given $Z$, and
\[
\E[(\psi_b-\phi_{\mathrm{eff}})^2\mid Z]=w(Z)d_b(Z)^2.
\]
To establish orthogonality, first note that
\[
\E\left[\frac{R}{\pi(Z)}-1\middle|Z\right]=0.
\]
Also, by the tower property,
\[
\begin{aligned}
&\E\left[\frac{R}{\pi(Z)}\left(\frac{R}{\pi(Z)}-1\right)\{Y-m(Z)\}\middle|Z\right]\\
&\quad=\E\left[\{Y-m(Z)\}\E\left[\frac{R}{\pi(Z)}\left(\frac{R}{\pi(Z)}-1\right)\middle|Z,Y\right]\middle|Z\right]\\
&\quad=w(Z)\E[Y-m(Z)\mid Z]=0.
\end{aligned}
\]
Therefore,
\[
\begin{aligned}
&\E[\phi_{\mathrm{eff}} (\psi_b-\phi_{\mathrm{eff}})\mid Z]\\
&\quad=\bar m(Z)d_b(Z)\E\left[\frac{R}{\pi(Z)}-1\middle|Z\right]\\
&\qquad+d_b(Z)\E\left[\frac{R}{\pi(Z)}\left(\frac{R}{\pi(Z)}-1\right)\{Y-m(Z)\}\middle|Z\right]\\
&\quad=0.
\end{aligned}
\]
Hence $\phi_{\mathrm{eff}}$ and $\psi_b-\phi_{\mathrm{eff}}$ are uncorrelated, so
\[
\Var(\psi_b)=V_{\mathrm{eff}}(\pi)+\E\left[w(Z)\{\bar m(Z)-b^\top h(Z)\}^2\right].
\]
Minimizing over $b$ proves
\[
V_{\EL}=V_{\mathrm{eff}}(\pi)+\inf_{b\in\R^r}\E\left[\frac{1-\pi(Z)}{\pi(Z)}\{m(Z)-\mu_0-b^\top h(Z)\}^2\right],
\]
where the infimum is attained at $b^*=\Omega_B^{-1}\E[BA]$.

The decomposition also implies
\[
\begin{aligned}
V_{\EL}
&\ge V_{\mathrm{eff}}(\pi)\\
&=\Var\{m(Z)\}+\E\left[\frac{\sigma^2(Z)}{\pi(Z)}\right]\\
&\ge\Var\{m(Z)\}+\E[\sigma^2(Z)]\\
&=\Var(Y).
\end{aligned}
\]
Thus, if $\Var(Y)>0$, then $V_{\EL}>0$.
Since $\Omega_B$ is positive definite, the Schur complement criterion then implies that $\Omega$ is positive definite.

Finally, consider $h(Z)=(1,\widetilde m(Z))^\top$.
Taking $b=(-\mu_0,1)^\top$ gives
\[
\bar m(Z)-b^\top h(Z)=m(Z)-\widetilde m(Z).
\]
Since $0\le w(Z)\le\frac{1-\underline\pi}{\underline\pi}$, the decomposition yields
\[
\begin{aligned}
0
&\le V_{\EL}-V_{\mathrm{eff}}(\pi)\\
&\le\E\left[w(Z)\{m(Z)-\widetilde m(Z)\}^2\right]\\
&\le\frac{1-\underline\pi}{\underline\pi}\|\widetilde m-m\|_2^2,
\end{aligned}
\]
which establishes (16).
\end{proof}

\subsection{Proof of Theorem~\ref{thm:oracle-annotation}}
\label{app:subsection:A.4}
\label{app:optimal-annotation}

\begin{proof}[Proof of Theorem~\ref{thm:oracle-annotation}]
Write $\mathcal P_\rho=\left\{p:\mathcal Z\to[\underline\pi,1]\text{ measurable}:\E[p(Z)]\le\rho\right\}$ and $J(p)=\E\left[\frac{\sigma^2(Z)}{p(Z)}\right].$
We first establish the optimal fixed audit rule and then prove the lower bound for adaptive rules.
For $\tau>0$, define $p_\tau(z)=\clip_{[\underline\pi,1]}\left\{\frac{\sigma(z)}{\tau}\right\}$ and $\Psi(\tau)=\E[p_\tau(Z)]$.
The function $\Psi$ is nonincreasing.
For every fixed $z$, $p_\tau(z)$ is continuous in
$\tau>0$ and bounded by one.
Dominated convergence therefore implies that $\Psi$
is continuous on $(0,\infty)$.
Since $\sigma(Z)>0$ almost surely,
\[
\lim_{\tau\downarrow0}\Psi(\tau)=1,
\qquad
\lim_{\tau\uparrow\infty}\Psi(\tau)=\underline\pi.
\]
Because $\underline\pi<\rho<1$, there exists
$\tau^*>0$ such that $\Psi(\tau^*)=\rho$.
Set $\pi^*=p_{\tau^*}$.

For each $z$ with $\sigma(z)>0$, consider
\[
f_z(p)=\frac{\sigma^2(z)}{p}+(\tau^*)^2p,
\qquad p\in[\underline\pi,1].
\]
Its derivatives satisfy
\[
f_z'(p)=-\frac{\sigma^2(z)}{p^2}+(\tau^*)^2,
\qquad
f_z''(p)=\frac{2\sigma^2(z)}{p^3}>0.
\]
Thus $f_z$ is strictly convex and its unique minimizer
on $[\underline\pi,1]$ is
\[
p=
\clip_{[\underline\pi,1]}
\left\{\frac{\sigma(z)}{\tau^*}\right\}
=\pi^*(z).
\]
Consequently, for every $p\in\mathcal P_\rho$,
\[
\frac{\sigma^2(Z)}{p(Z)}+(\tau^*)^2p(Z)
\ge
\frac{\sigma^2(Z)}{\pi^*(Z)}
+(\tau^*)^2\pi^*(Z)
\qquad\text{almost surely}.
\]
Taking expectations and using $\E[\pi^*(Z)]=\rho$
gives
\[
J(p)-J(\pi^*)
\ge
(\tau^*)^2\{\rho-\E[p(Z)]\}
\ge0.
\]
Hence $\pi^*$ solves the fixed-design optimization
problem.

To prove uniqueness, suppose that $p\in\mathcal P_\rho$
satisfies $J(p)=J(\pi^*)$.
Define the nonnegative pointwise Lagrangian gap
\[
L_p(Z)
=
\frac{\sigma^2(Z)}{p(Z)}+(\tau^*)^2p(Z)
-\frac{\sigma^2(Z)}{\pi^*(Z)}
-(\tau^*)^2\pi^*(Z).
\]
Then
\[
0=J(p)-J(\pi^*)
=
\E[L_p(Z)]
+(\tau^*)^2\{\rho-\E[p(Z)]\}.
\]
Both terms on the right are nonnegative.
Therefore $\E[L_p(Z)]=0$, so $L_p(Z)=0$ almost surely.
Strict convexity on $\{\sigma>0\}$ implies $p(Z)=\pi^*(Z)$ almost surely.
Thus the optimal audit rule is almost surely unique, although its representing threshold need not be unique.

Fix $n\ge1$ and consider any admissible adaptive audit rule.
The pointwise Lagrangian inequality established above holds for every $p\in[\underline\pi,1]$.
Applying it with $z=Z_t$ and $p=\pi_t$ gives
\[
\frac{\sigma^2(Z_t)}{\pi_t}+(\tau^*)^2\pi_t\ge\frac{\sigma^2(Z_t)}{\pi^*(Z_t)}+(\tau^*)^2\pi^*(Z_t)\qquad\text{almost surely}.
\]
Taking expectations, averaging over $t$, and using $\E[\pi^*(Z_t)]=\rho$, we obtain
\[
\frac1n\sum_{t=1}^n\E\left[\frac{\sigma^2(Z_t)}{\pi_t}\right]\ge\E\left[\frac{\sigma^2(Z)}{\pi^*(Z)}\right]+(\tau^*)^2\left\{\rho-\frac1n\sum_{t=1}^n\E[\pi_t]\right\}.
\]
Under the finite-sample expected budget constraint, the last term is nonnegative.
This proves (19).

The same inequality also shows that
\[
\limsup_{n\to\infty}\frac1n\sum_{t=1}^n\E[\pi_t]\le\rho
\]
implies
\[
\liminf_{n\to\infty}\frac1n\sum_{t=1}^n\E\left[\frac{\sigma^2(Z_t)}{\pi_t}\right]\ge\E\left[\frac{\sigma^2(Z)}{\pi^*(Z)}\right].
\]

If clipping is inactive for $\pi^*$, then $\pi^*(Z)=\frac{\sigma(Z)}{\tau^*}.$
The budget identity implies $\tau^*=\frac{\E[\sigma(Z)]}{\rho},$ and therefore
\[
J(\pi^*)=\tau^*\E[\sigma(Z)]=\frac{\{\E[\sigma(Z)]\}^2}{\rho}.
\]
For uniform auditing, $\pi_{\mathrm{unif}}(Z)=\rho$, so
\[
J(\pi_{\mathrm{unif}})=\frac{\E[\sigma^2(Z)]}{\rho}.
\]
Since the term $\Var\{m(Z)\}$ in $V_{\mathrm{eff}}(\pi)$ does not depend on $\pi$,
\[
\begin{aligned}
V_{\mathrm{eff}}(\pi_{\mathrm{unif}})-V_{\mathrm{oracle}}
&=J(\pi_{\mathrm{unif}})-J(\pi^*)\\
&=\frac{\E[\sigma^2(Z)]-\{\E[\sigma(Z)]\}^2}{\rho}\\
&=\frac{\Var\{\sigma(Z)\}}{\rho},
\end{aligned}
\]
which establishes (20).
\end{proof}

\subsection{Proof of Proposition~\ref{prop:adaptive-variance-bound}}
\label{app:subsection:A.5}
\label{app:adaptive-variance-bound}

\begin{proof}[Proof of Proposition~\ref{prop:adaptive-variance-bound}]
For a symmetric positive semidefinite $q\times q$ matrix $M$, define $\mathcal V(M)=\inf_{b\in\R^r}(1,-b^\top)M(1,-b^\top)^\top.$
As an infimum of linear functions of $M$, $\mathcal V$ is concave.
Note that $V_{\EL}=\mathcal V(\Omega)$ uses a single coefficient $b$ for all $t$, even though $h_t$ and $\pi_t$ may vary with $t$.
Concavity is exactly what allows the time-varying per-period bounds below, each of which optimizes $b$ separately, to be transferred to this common-coefficient criterion.
In particular,
\[
\mathcal V\left(\frac1n\sum_{t=1}^nM_t\right)\ge\frac1n\sum_{t=1}^n\mathcal V(M_t).
\]
On positive definite matrices, $\mathcal V$ equals the Schur complement $M_{11}-M_{12}M_{22}^{-1}M_{21}$ and is continuous.

Write $\Omega_t=\E[G_t(\mu_0)G_t(\mu_0)^\top\mid\mathcal F_{t-1}]$ and $\bar m_t=m(Z_t)-\mu_0.$
For every $b\in\R^r$,
\[
A_t(\mu_0)-b^\top B_t
=\bar m_t+\frac{R_t}{\pi_t}\{Y_t-m(Z_t)\}
+\left(\frac{R_t}{\pi_t}-1\right)\{\bar m_t-b^\top h_t\}.
\]
The audit mechanism and independence of $(Z_t,Y_t)$ from $\mathcal F_{t-1}$ (Lemma~\ref{lem:standing}) imply that the three terms on the right have conditional mean zero given $\mathcal F_{t-1}$ and are pairwise conditionally uncorrelated.
Consequently,
\[
\begin{aligned}
(1,-b^\top)\Omega_t(1,-b^\top)^\top
=\Var\{m(Z)\}
+\E\left[\frac{\sigma^2(Z_t)}{\pi_t}\middle|\mathcal F_{t-1}\right]+\E\left[\frac{1-\pi_t}{\pi_t}\{\bar m_t-b^\top h_t\}^2\middle|\mathcal F_{t-1}\right].
\end{aligned}
\]
Taking the infimum over $b$ gives
\[
\mathcal V(\Omega_t)\ge\Var\{m(Z)\}
+\E\left[\frac{\sigma^2(Z_t)}{\pi_t}\middle|\mathcal F_{t-1}\right].
\]

Let $\tau^*>0$ be a threshold representing the oracle rule.
The pointwise Lagrangian inequality in the proof of Theorem~\ref{thm:oracle-annotation} gives
\[
\frac{\sigma^2(Z_t)}{\pi_t}+(\tau^*)^2\pi_t
\ge\frac{\sigma^2(Z_t)}{\pi^*(Z_t)}+(\tau^*)^2\pi^*(Z_t)
\qquad\text{almost surely}.
\]
Taking conditional expectations and using independence of $Z_t$ from $\mathcal F_{t-1}$ yields
\[
\mathcal V(\Omega_t)\ge V_{\mathrm{oracle}}
+(\tau^*)^2\bigl\{\rho-\E[\pi_t\mid\mathcal F_{t-1}]\bigr\}.
\]
Therefore, with $M_n=n^{-1}\sum_{t=1}^n\Omega_t$,
\[
\mathcal V(M_n)\ge V_{\mathrm{oracle}}
+(\tau^*)^2\left\{\rho-\frac1n\sum_{t=1}^n\E[\pi_t\mid\mathcal F_{t-1}]\right\},
\]
and taking expectations gives
\[
\E[\mathcal V(M_n)]\ge V_{\mathrm{oracle}}
+(\tau^*)^2\left\{\rho-\frac1n\sum_{t=1}^n\E[\pi_t]\right\}.
\]

By Condition~\ref{cond:cov}, $M_n\xrightarrow{p}\Omega$ with $\Omega$ positive definite, so continuity of $\mathcal V$ at $\Omega$ gives $\mathcal V(M_n)\xrightarrow{p}\mathcal V(\Omega)=V_{\EL}$.
The deterministic bound $\|G_t(\mu_0)\|\le C_G$ gives
\[
0\le\mathcal V(M_n)\le e_1^\top M_ne_1\le C_G^2,
\]
so the convergence also holds in $L^1$ and $\E[\mathcal V(M_n)]\to V_{\EL}$.
Taking $\liminf_{n\to\infty}$ on both sides and using the long-run expected budget gives
\[
V_{\EL}\ge V_{\mathrm{oracle}}
+(\tau^*)^2\left\{\rho-\limsup_{n\to\infty}\frac1n\sum_{t=1}^n\E[\pi_t]\right\}
\ge V_{\mathrm{oracle}}.
\]
\end{proof}

\subsection{Proof of Remark~\ref{rem:information}}
\label{app:subsection:A.6}
\label{app:information}

For $W\in\{Z,Z'\}$, write $\sigma_W^2(W)=\Var(Y\mid W)$ and
\[
V^*(W)=\inf\Bigl\{\Var(Y)+\E\Bigl[\sigma_W^2(W)\tfrac{1-\pi(W)}{\pi(W)}\Bigr]:\ \underline\pi\le\pi(W)\le1,\ \E[\pi(W)]\le\rho\Bigr\}.
\]
This equals $\inf_\pi V_{\mathrm{eff}}(\pi)$ computed with information $W$, because
$\Var\{m(W)\}+\E[\sigma_W^2(W)/\pi(W)]=\Var(Y)+\E[\sigma_W^2(W)\{1-\pi(W)\}/\pi(W)]$.
Any audit rule $\pi(Z)$ feasible under $Z$ is also feasible under $Z'$, and the law of total variance gives $\E[\sigma_{Z'}^2(Z')\mid Z]\le\sigma_Z^2(Z)$.
Since $\{1-\pi(Z)\}/\pi(Z)\ge0$ is a function of $Z$,
\[
\E\Bigl[\sigma_{Z'}^2(Z')\tfrac{1-\pi(Z)}{\pi(Z)}\Bigr]
\le\E\Bigl[\sigma_Z^2(Z)\tfrac{1-\pi(Z)}{\pi(Z)}\Bigr].
\]
Taking the infimum over $\pi(Z)$ gives $V^*(Z')\le V^*(Z)$.

\subsection{Proof of Corollary~\ref{cor:oracle-recovery}}
\label{app:subsection:A.7}
\label{app:oracle-recovery}

All function norms below are taken with respect to $\Pp_Z$, with fitted functions held fixed during integration.
Define $L(f)=\E[(Y-f(Z))^2]$ and $\widehat L_n(f)=\frac1n\sum_{t=1}^n\frac{R_t}{\pi_t}\{Y_t-f(Z_t)\}^2.$

\begin{proposition}[IPW prediction consistency]
\label{prop:ipw-consistency}
Let $\mathcal M_n$ be a deterministic class of measurable
functions taking values in $[0,1]$.
Suppose that $\widehat m_n\in\mathcal M_n$ satisfies $\widehat L_n(\widehat m_n)\le\inf_{f\in\mathcal M_n}\widehat L_n(f)+\upsilon_n$ and $\upsilon_n=o_p(1).$
If $a_n:=\inf_{f\in\mathcal M_n}\|f-m\|_2^2\to0$ and $\Delta_n:=\sup_{f\in\mathcal M_n}|\widehat L_n(f)-L(f)|=o_p(1),$ then $\|\widehat m_n-m\|_2\xrightarrow{p}0$.
\end{proposition}

\begin{proof}
For every deterministic $f$, randomized auditing and
independence of the current evaluation instance from the
past imply
\[
\E\left[
\frac{R_t}{\pi_t}\{Y_t-f(Z_t)\}^2
\middle|\mathcal F_{t-1}
\right]=L(f).
\]
The regression identity gives
\[
L(f)-L(m)=\|f-m\|_2^2.
\]
The approximate empirical-risk inequality therefore yields
\[
\begin{aligned}
L(\widehat m_n)
&\le \widehat L_n(\widehat m_n)+\Delta_n\\
&\le \inf_{f\in\mathcal M_n}\widehat L_n(f)
+\upsilon_n+\Delta_n\\
&\le \inf_{f\in\mathcal M_n}L(f)+\upsilon_n+2\Delta_n.
\end{aligned}
\]
Consequently,
\[
\|\widehat m_n-m\|_2^2
\le a_n+\upsilon_n+2\Delta_n=o_p(1).
\]
\end{proof}

For a finite class $\mathcal M_n$,
the centered IPW losses are martingale differences bounded
in absolute value by $\underline\pi^{-1}$.
Azuma--Hoeffding and a union bound imply that, with
probability at least $1-\delta$,
\[
\Delta_n
\le
\frac1{\underline\pi}
\sqrt{\frac{2\log(2|\mathcal M_n|/\delta)}n}.
\]
Thus $\log|\mathcal M_n|=o(n)$, vanishing approximation error, and $\upsilon_n=o_p(1)$ suffice for prediction consistency.

For general $Y\in[0,1]$, write
$s_2(z)=\E[Y^2\mid Z=z]$.
Estimate $s_2$ by the same IPW construction with response
$Y_t^2$, restricting both fitted regressions to $[0,1]$.
Set
\[
\widehat\sigma_n
=
\left\{[\widehat s_{2,n}-\widehat m_n^2]_+\right\}^{1/2}.
\]
Since the positive-part map is nonexpansive and
$(\sqrt a-\sqrt b)^2\le|a-b|$ for $a,b\ge0$,
\[
\|\widehat\sigma_n-\sigma\|_2^2
\le
\|\widehat s_{2,n}-s_2\|_2
+2\|\widehat m_n-m\|_2.
\]
Consistency of both regressions therefore implies
$\|\widehat\sigma_n-\sigma\|_2\xrightarrow{p}0$.
For binary responses, one may instead set $\widehat\sigma_n=\{\widehat m_n(1-\widehat m_n)\}^{1/2}.$
In this case,
\[
\|\widehat\sigma_n-\sigma\|_2^2
\le \|\widehat m_n-m\|_1
\le \|\widehat m_n-m\|_2.
\]

\begin{proposition}
\label{prop:threshold-consistency}
Suppose that $\|\widehat\sigma_n-\sigma\|_2\xrightarrow{p}0$ and that $\widehat\sigma_n\in\mathcal S_n$, where $\mathcal S_n$ is a deterministic class of measurable functions with values in $[0,1]$ such that
\[
\Gamma_n:=\sup_{f\in\mathcal S_n,\ \tau\in T}
\left|\frac1n\sum_{s=1}^n\operatorname{clip}_{[\underline\pi,1]}\Bigl\{\frac{f(Z_s)}{\tau}\Bigr\}
-\E\Bigl[\operatorname{clip}_{[\underline\pi,1]}\Bigl\{\frac{f(Z)}{\tau}\Bigr\}\Bigr]\right|=o_p(1).
\]
Use the threshold definition in (9).
Assume that $\Psi_\sigma(\tau)=\E\left[\operatorname{clip}_{[\underline\pi,1]}\left\{\frac{\sigma(Z)}\tau\right\}\right]$ has a unique solution $\tau^*$ to $\Psi_\sigma(\tau)=\rho$ on $T=[\tau_-,\tau_+]$, and that $\tau^*\in\operatorname{int}(T)$.
Then $\widehat\tau_n\xrightarrow{p}\tau^*$ and $\|\widehat\pi_n-\pi^*\|_2\xrightarrow{p}0.$
\end{proposition}
The condition on $\Gamma_n$ holds, for example, when $\mathcal S_n\equiv\mathcal S$ is $\Pp_Z$-Glivenko--Cantelli, because $(x,\tau)\mapsto\operatorname{clip}_{[\underline\pi,1]}(x/\tau)$ is continuous and bounded on $[0,1]\times T$ \citep{vandervaart2000preservation}.
It parallels the uniform condition on $\Delta_n$ in Proposition~\ref{prop:ipw-consistency}.

\begin{proof}
The clipping map is one-Lipschitz, so
\[
\sup_{\tau\in T}
|\Psi_{\widehat\sigma_n}(\tau)-\Psi_\sigma(\tau)|
\le
\tau_-^{-1}\|\widehat\sigma_n-\sigma\|_1=o_p(1).
\]
Because $Z_1,\ldots,Z_n$ are i.i.d.\ with distribution $\Pp_Z$ by Condition~\ref{cond:basic}(a)
and $\widehat\sigma_n\in\mathcal S_n$,
\[
\sup_{\tau\in T}|\widehat\Psi_n(\tau)-\Psi_{\widehat\sigma_n}(\tau)|\le\Gamma_n=o_p(1),
\]
where $\Psi_{\widehat\sigma_n}$ is $\Psi_\sigma$ with $\sigma$ replaced by the fixed function $\widehat\sigma_n$.
No independence between $\widehat\sigma_n$ and the calibration sample is needed.
Consequently,
\[
\sup_{\tau\in T}
|\widehat\Psi_n(\tau)-\Psi_\sigma(\tau)|=o_p(1).
\]

Continuity, compactness, and uniqueness imply that, for
every $\epsilon>0$ for which the indexing set is nonempty,
\[
\inf_{\substack{\tau\in T\\|\tau-\tau^*|\ge\epsilon}}
|\Psi_\sigma(\tau)-\rho|>0.
\]
Because $\widehat\tau_n$ minimizes the empirical discrepancy
and $\Psi_\sigma(\tau^*)=\rho$, uniform convergence implies
$\widehat\tau_n\xrightarrow{p}\tau^*$.

Finally, the Lipschitz property of clipping gives
\[
\|\widehat\pi_n-\pi^*\|_2
\le
\tau_-^{-1}\|\widehat\sigma_n-\sigma\|_2
+
\|\sigma\|_2
\left|
\widehat\tau_n^{-1}-(\tau^*)^{-1}
\right|,
\]
which converges to zero in probability.
\end{proof}

\begin{proof}[Proof of Corollary~\ref{cor:oracle-recovery}]
Write $p_n(z)=\widehat\pi_n(z)$, $p^*(z)=\pi^*(z)$, and $h^*(z)=\begin{pmatrix}1\\m(z)\end{pmatrix}.$
As in the learned construction, the predictors take values in $[0,1]$.
All $L^2$ norms below are with respect to the common distribution $\Pp_Z$, treating the fitted functions as fixed when integrating over $z$.
We prove convergence of the audit rules, covariance stabilization, and recovery of the oracle variance.

The clipping map is one-Lipschitz.
On the event $\widehat\tau_n\ge\tau^*/2$,
\[
\begin{aligned}
\|p_n-p^*\|_2
&\le
\left\|
\frac{\widehat\sigma_n}{\widehat\tau_n}
-\frac{\sigma}{\tau^*}
\right\|_2\\
&\le
\frac{2}{\tau^*}\|\widehat\sigma_n-\sigma\|_2
+
\|\sigma\|_2
\left|\frac1{\widehat\tau_n}-\frac1{\tau^*}\right|.
\end{aligned}
\]
Since $\widehat\tau_n\xrightarrow{p}\tau^*>0$, this event has
probability tending to one.
The assumed consistency of $\widehat\sigma_n$ therefore
implies
\[
\|p_n-p^*\|_2\xrightarrow{p}0.
\]

Define $e_n^m=\|\widehat m_n-m\|_2$ and $e_n^\pi=\|p_n-p^*\|_2$.
Both errors are uniformly bounded because $0\le\widehat m_n,m\le1$ and $\underline\pi\le p_n,p^*\le1$.
Their convergence to zero in probability consequently gives
\[
\E[e_n^m]\to0,
\qquad
\E[e_n^\pi]\to0.
\]
For the burn-in time $n_0$, which is finite almost surely, Cesaro averaging and Markov's inequality yield
\[
\frac1n\sum_{t=n_0+1}^n e_{t-1}^m\xrightarrow{p}0,
\qquad
\frac1n\sum_{t=n_0+1}^n e_{t-1}^\pi\xrightarrow{p}0.
\]

Define $\Omega_t=\E[G_t(\mu_0)G_t(\mu_0)^\top\mid\mathcal F_{t-1}].$
For $t>n_0$, the learned design uses $h_t(z)=\begin{pmatrix}1\\\widehat m_{t-1}(z)\end{pmatrix}$ and $\pi_t=p_{t-1}(Z_t)$.
Write $\bar m(z)=m(z)-\mu_0$.
By Lemma~\ref{lem:standing} \textup{(i)--(ii)}, $(Z_t,Y_t)$ is independent of $\mathcal F_{t-1}$ and $\E[R_t\mid\mathcal F_{t-1},Z_t,Y_t]=\pi_t.$
Conditional on $\mathcal F_{t-1}$, the fitted functions are fixed.
Using $R_t^2=R_t$ and iterated conditional expectation, we obtain
\[
\Omega_t=
\begin{pmatrix}
a_t & c_t^\top\\
c_t & M_t
\end{pmatrix},
\]
where
\[
a_t=\int\frac{\sigma^2(z)+\bar m(z)^2}{p_{t-1}(z)}\,\Pp_Z(dz),
\]
\[
c_t=\int\frac{1-p_{t-1}(z)}{p_{t-1}(z)}h_t(z)\bar m(z)\,\Pp_Z(dz),
\]
and
\[
M_t=\int\frac{1-p_{t-1}(z)}{p_{t-1}(z)}h_t(z)h_t(z)^\top\,\Pp_Z(dz).
\]
Let $a^*$, $c^*$, and $M^*$ denote the corresponding expressions with $p_{t-1}$ and $h_t$ replaced by $p^*$ and $h^*$.
Then
\[
\Omega^*=
\begin{pmatrix}
a^* & (c^*)^\top\\
c^* & M^*
\end{pmatrix}
\]
is the covariance matrix under the fixed oracle design.

For $p,q\in[\underline\pi,1]$,
\[
\left|\frac1p-\frac1q\right|
\le\underline\pi^{-2}|p-q|,
\qquad
\left|\frac{1-p}{p}-\frac{1-q}{q}\right|
\le\underline\pi^{-2}|p-q|.
\]
Moreover,
\[
\|h_t(z)\|\le\sqrt2,
\qquad
\|h^*(z)\|\le\sqrt2,
\qquad
\|h_t(z)-h^*(z)\|
=
|\widehat m_{t-1}(z)-m(z)|.
\]
The outer-product identity gives
\[
\|h_t(z)h_t(z)^\top-h^*(z)h^*(z)^\top\|
\le
2\sqrt2\,|\widehat m_{t-1}(z)-m(z)|.
\]
Since $|\bar m(z)|\le1$ and
\[
\sigma^2(z)+\bar m(z)^2=\E[(Y-\mu_0)^2\mid Z=z]\le1,
\]
these bounds and the Cauchy--Schwarz inequality imply
that, for a deterministic constant $C<\infty$,
\[
|a_t-a^*|\le C e_{t-1}^\pi, \qquad \|c_t-c^*\|+\|M_t-M^*\|\le C(e_{t-1}^\pi+e_{t-1}^m).
\]
Thus, after increasing $C$ if necessary,
\[
\|\Omega_t-\Omega^*\|
\le C(e_{t-1}^\pi+e_{t-1}^m),
\qquad t>n_0.
\]

During burn-in, $\pi_t=\rho$ and $h_t=(1,1/2)^\top$.
Hence $\|h_t\|\le\sqrt2$ throughout the procedure.
Choose a deterministic $C_G$ that bounds the estimating vectors at $\mu_0$ under both the learned and oracle designs.
Then $\|\Omega_t\|\le C_G^2$ and $\|\Omega^*\|\le C_G^2$, so the burn-in contribution satisfies
\[
\left\|\frac1n\sum_{t=1}^{n_0\wedge n}(\Omega_t-\Omega^*)\right\|\le\frac{2n_0C_G^2}{n}\xrightarrow{\mathrm{a.s.}}0.
\]
Consequently, for $n>n_0$,
\[
\begin{aligned}\left\|\frac1n\sum_{t=1}^n\Omega_t-\Omega^*\right\|
&\le\frac{2n_0C_G^2}{n}+\frac Cn\sum_{t=n_0+1}^n(e_{t-1}^\pi+e_{t-1}^m)\\
&\xrightarrow{p}0.
\end{aligned}
\]
Thus Condition~\ref{cond:cov} holds with $\Omega=\Omega^*$, which is positive definite by assumption.

Since $0\le m\le1$, the limiting auxiliary vector $h^*(Z)=(1,m(Z))^\top$ is bounded.
The lower-right block $\Omega_B^*=M^*$ is a principal submatrix of the positive definite matrix $\Omega^*$ and is therefore positive definite.
Together with $\underline\pi\le\pi^*\le1$, these facts verify the conditions of Theorem~\ref{thm:eff-decomp} for the fixed design $(\pi^*,h^*)$.
Taking $b=(-\mu_0,1)^\top$ gives $b^\top h^*(Z)=m(Z)-\mu_0.$
Hence the approximation term in Theorem~\ref{thm:eff-decomp} vanishes, and
\[
V_{\EL}=V_{\mathrm{eff}}(\pi^*)=\Var\{m(Z)\}+\E\left[\frac{\sigma^2(Z)}{\pi^*(Z)}\right]=V_{\mathrm{oracle}}.
\]
Moreover, $\|h_t\|\le\sqrt2$ throughout the procedure, so the boundedness requirement in Condition~\ref{cond:basic}\textup{(d)} holds.
Together with the covariance stabilization established above and the remaining basic conditions, Theorem~\ref{thm:adaptive-wilks} identifies $V_{\mathrm{oracle}}$ as the asymptotic variance of $\widehat\mu_n$.
Finally, recall that $e_n^\pi=\|\widehat\pi_n-\pi^*\|_2$.
The errors are uniformly bounded, so their convergence to zero in probability implies $\E[e_n^\pi]\to0$.

Set $q_t=\E[\pi_t\mid\mathcal F_{t-1}]$.
Since $n_0$ is a stopping time, $\{n_0<t\}\in\mathcal F_{t-1}$.
During burn-in, $\pi_t=\rho$.
After burn-in, $\pi_t=\widehat\pi_{t-1}(Z_t)$.
Thus, for $t\ge2$, independence of $Z_t$ from $\mathcal F_{t-1}$ gives
\[
q_t=\rho\,\mathbf{1}\{n_0\ge t\}+\mathbf{1}\{n_0<t\}\int \widehat\pi_{t-1}(z)\,\Pp_Z(dz).
\]
If necessary, fitted audit rules before the end of burn-in are defined arbitrarily as $\mathcal F_{t-1}$-measurable functions taking values in $[\underline\pi,1]$.
These values do not affect the procedure or the assumed convergence as $n\to\infty$, since $n_0<\infty$ almost surely.

Using $\int\pi^*(z)\,\Pp_Z(dz)=\rho$ and the Cauchy--Schwarz inequality, we obtain
\[
\begin{aligned}
|q_t-\rho|&=\mathbf{1}\{n_0<t\}\left|\int\{\widehat\pi_{t-1}(z)-\pi^*(z)\}\,\Pp_Z(dz)\right|\\&\le\mathbf{1}\{n_0<t\}\,e_{t-1}^\pi.
\end{aligned}
\]
Consequently,
\[
|\E[\pi_t]-\rho|\le \E|q_t-\rho|\le \E[e_{t-1}^\pi]\to0.
\]
Cesaro averaging therefore yields
\[
\frac1n\sum_{t=1}^n\E[\pi_t]\to\rho.
\]

Moreover, since $q_1,\rho\in[0,1]$,
\[
\E\left|\frac1n\sum_{t=1}^n q_t-\rho\right|\le\frac1n+\frac1n\sum_{t=2}^n\E[e_{t-1}^\pi]\to0.
\]
Finally, $\pi_t-q_t$ is a bounded martingale difference with respect to $(\mathcal F_t)_{t\ge0}$.
Orthogonality therefore gives
\[
\E\left[\left\{\frac1n\sum_{t=1}^n(\pi_t-q_t)\right\}^2\right]=\frac1{n^2}\sum_{t=1}^n\E[(\pi_t-q_t)^2]\le \frac1n\to0.
\]
Combining the last two displays proves
\[
\frac1n\sum_{t=1}^n\pi_t\xrightarrow{p}\rho.
\]
This proves the corollary.
\end{proof}

\subsection{Proof of Theorem~\ref{thm:anytime-validity}}
\label{app:subsection:A.8}
\label{app:anytime-validity}

\begin{proof}[Proof of Theorem~\ref{thm:anytime-validity}]
We work at the true mean $\mu_0$.
By Proposition~\ref{prop:estimating-equations}, $\E[G_t(\mu_0)\mid\mathcal F_{t-1}]=0_q.$
Since $\widehat a_{t-1}(\mu_0)$ is $\mathcal F_{t-1}$-measurable and bounded,
\[
\E[S_t(\mu_0)\mid\mathcal F_{t-1}]=\widehat a_{t-1}(\mu_0)^\top\E[G_t(\mu_0)\mid\mathcal F_{t-1}]=0.
\]
By Lemma~\ref{lem:standing}, using Condition~\ref{cond:basic}\textup{(d)} and $\mu_0\in\Theta$, we have $\|G_t(\mu_0)\|\le C_G$ almost surely.
Together with $\|\widehat a_{t-1}(\mu_0)\|\le C_a$, this gives
\[
|S_t(\mu_0)|\le C_aC_G=C_S\qquad\text{almost surely}.
\]

For each fixed $\beta\in I_\eta=[-\eta/C_S,\eta/C_S]$, define $M_n^\beta=\prod_{t=1}^n\{1+\beta S_t(\mu_0)\}$ and $M_0^\beta=1$.
Every factor satisfies
\[
1-\eta\le1+\beta S_t(\mu_0)\le1+\eta.
\]
Thus $M_n^\beta$ is nonnegative, adapted, and integrable, with $M_n^\beta\le(1+\eta)^n$.
Moreover,
\[
\begin{aligned}
\E[M_n^\beta\mid\mathcal F_{n-1}]
&=M_{n-1}^\beta\E[1+\beta S_n(\mu_0)\mid\mathcal F_{n-1}]\\
&=M_{n-1}^\beta.
\end{aligned}
\]
Hence $(M_n^\beta)_{n\ge0}$ is a nonnegative martingale with $\E[M_n^\beta]=1$ for every $n$.

The mixing distribution $\varpi$ is a prespecified probability measure supported on $I_\eta$.
For every fixed $n$, $M_n^\beta$ is a polynomial in $\beta$ with $\mathcal F_n$-measurable coefficients.
Therefore $(\beta,\omega)\mapsto M_n^\beta(\omega)$ is $\mathcal B(I_\eta)\otimes\mathcal F_n$-measurable.
The mixture $E_n(\mu_0)=\int_{I_\eta}M_n^\beta\,\varpi(d\beta)$ is consequently $\mathcal F_n$-measurable, and conditional Tonelli's theorem gives
\[
\begin{aligned}
\E[E_n(\mu_0)\mid\mathcal F_{n-1}]
&=
\int_{I_\eta}\E[M_n^\beta\mid\mathcal F_{n-1}]\,\varpi(d\beta)\\
&=\int_{I_\eta}M_{n-1}^\beta\,\varpi(d\beta)\\
&=E_{n-1}(\mu_0).
\end{aligned}
\]
Also, $E_0(\mu_0)=1$ and, by Tonelli's theorem,
\[
\E[E_n(\mu_0)]=\int_{I_\eta}\E[M_n^\beta]\,\varpi(d\beta)=1.
\]
Thus $(E_n(\mu_0))_{n\ge0}$ is a unit-mean nonnegative martingale.

Ville's inequality \citep{howard2021time} yields
\[
\Pp\left(\sup_{n\ge1}E_n(\mu_0)\ge\frac1\alpha\right)\le\alpha.
\]
Since $\mathcal C_n=\{\mu\in\Theta:E_n(\mu)<1/\alpha\}$ and $\mu_0\in\Theta$, we obtain
\[
\begin{aligned}
\Pp(\mu_0\in\mathcal C_n\text{ for all }n\ge1)
&=1-\Pp\left(\exists n\ge1:E_n(\mu_0)\ge\frac1\alpha\right)\\
&\ge1-\alpha.
\end{aligned}
\]

Finally, for every almost surely finite, positive integer-valued stopping time $N$, the inclusion
\[
\{\mu_0\in\mathcal C_n\text{ for all }n\ge1\}\cap\{N<\infty\}\subseteq\{\mu_0\in\mathcal C_N\}
\]
holds pointwise.
Consequently,
\[
\Pp(\mu_0\in\mathcal C_N)\ge1-\alpha.
\]
In fact, the same inclusion and coverage conclusion hold for any measurable, almost surely finite, positive integer-valued random time $N$, without requiring the stopping-time property.
\end{proof}

\subsection{Proof of Theorem~\ref{thm:vector-bridge}}
\label{app:subsection:A.9}
\label{app:vector-bridge}

\begin{lemma}
\label{lem:generic-laplace}
Let $d$ be fixed, let $\mathcal V\subset\R^d$ be a fixed compact convex set containing zero in its interior, and let $\ell_n$ be a random concave function that is three times continuously differentiable on an open neighborhood of $\mathcal V$, with maximizer $\widehat\vartheta_n$ over $\mathcal V$.
All statements below are on events of probability tending to one on which $\widehat\vartheta_n$ lies in the interior of $\mathcal V$.
Suppose that
\textup{(i)} $\widehat\vartheta_n=O_p(n^{-1/2})$;
\textup{(ii)} for a deterministic $c>0$, $-n^{-1}\nabla^2\ell_n(\vartheta)\succeq cI_d$ uniformly on $\mathcal V$ with probability tending to one;
\textup{(iii)} $H_n=-n^{-1}\nabla^2\ell_n(\widehat\vartheta_n)\xrightarrow{p}H$ for a positive definite $H$;
\textup{(iv)} the third derivatives of $\ell_n$ are bounded in norm by $Cn$ on $\mathcal V$ for a deterministic $C$; and
\textup{(v)} $\varphi$ is a probability density on $\mathcal V$ that is continuous and positive at zero.
Then
\[
\int_{\mathcal V}e^{\ell_n(\vartheta)}\varphi(\vartheta)\,d\vartheta
=e^{\ell_n(\widehat\vartheta_n)}\varphi(0)\Bigl(\frac{2\uppi}{n}\Bigr)^{d/2}\det(H)^{-1/2}\{1+o_p(1)\}.
\]
\end{lemma}

\begin{proof}
Let $r_n=(\log n)/\sqrt n$.
By \textup{(i)}, the ball of radius $r_n$ around $\widehat\vartheta_n$ lies in $\mathcal V$ with probability tending to one.
By \textup{(ii)} and $\nabla\ell_n(\widehat\vartheta_n)=0$, $\ell_n(\vartheta)\le\ell_n(\widehat\vartheta_n)-cn\|\vartheta-\widehat\vartheta_n\|^2/2$ on $\mathcal V$.
Since $\varphi$ integrates to one, the integral outside this ball is at most $e^{\ell_n(\widehat\vartheta_n)}e^{-c(\log n)^2/2}=o_p(e^{\ell_n(\widehat\vartheta_n)}n^{-d/2})$.
Inside the ball, write $\vartheta=\widehat\vartheta_n+x/\sqrt n$.
Taylor's theorem and \textup{(iv)} give $\ell_n(\widehat\vartheta_n+x/\sqrt n)=\ell_n(\widehat\vartheta_n)-x^\top H_nx/2+O\{(\log n)^3/\sqrt n\}$ uniformly for $\|x\|\le\log n$.
By \textup{(v)} and \textup{(i)}, $\sup_{\|x\|\le\log n}|\varphi(\widehat\vartheta_n+x/\sqrt n)-\varphi(0)|\xrightarrow{p}0$.
The bound in \textup{(ii)} provides the Gaussian envelope $e^{-c\|x\|^2/2}$, so dominated convergence and \textup{(iii)} give
\[
\int_{\|x\|\le\log n}e^{-x^\top H_nx/2}\,dx\xrightarrow{p}(2\uppi)^{d/2}\det(H)^{-1/2}.
\]
Combining the two regions and multiplying by the Jacobian $n^{-d/2}$ proves the claim.
\end{proof}

\begin{proof}[Proof of Theorem~\ref{thm:vector-bridge}]
For fixed $\lambda\in\Lambda_\eta$, write $M_n(\lambda)=\prod_{t\le n}\{1+\lambda^\top G_t(\mu_0)\}$.
Every factor lies in $[1-\eta,1+\eta]$, and Proposition~\ref{prop:estimating-equations} gives $\E[1+\lambda^\top G_t(\mu_0)\mid\mathcal F_{t-1}]=1$.
Hence $M_n(\lambda)$ is a nonnegative martingale with unit mean.
Since $M_n(\lambda)$ is a polynomial in $\lambda$ with $\mathcal F_n$-measurable coefficients, the conditional Tonelli argument in the proof of Theorem~\ref{thm:anytime-validity} shows that $E_n^{\mathrm{vec}}(\mu_0)$ is a unit-mean nonnegative martingale, and Ville's inequality gives the stated coverage.

For the predictable representation, fix $\mu$ and write $M_{t}(\lambda,\mu)=\prod_{s\le t}\{1+\lambda^\top G_s(\mu)\}$.
Then
\[
E_t^{\mathrm{vec}}(\mu)
=\int_{\Lambda_\eta}M_{t-1}(\lambda,\mu)\{1+\lambda^\top G_t(\mu)\}\,\nu(d\lambda)
=E_{t-1}^{\mathrm{vec}}(\mu)\bigl\{1+\bar\lambda_{t-1}(\mu)^\top G_t(\mu)\bigr\}.
\]
Iterating from $E_0^{\mathrm{vec}}=1$ gives the product form.
The multiplier $\bar\lambda_{t-1}(\mu)$ is a $\mathcal F_{t-1}$-measurable average of points in the convex set $\Lambda_\eta$, so it lies in $\Lambda_\eta$.

Now assume Condition~\ref{cond:cov} and apply Lemma~\ref{lem:generic-laplace} with $d=q$, $\mathcal V=\Lambda_\eta$, and $\ell_n(\lambda)=\sum_{t\le n}\log\{1+\lambda^\top G_t(\mu_0)\}$.
The proof of Lemma~\ref{lem:local-el} at $u=0$ shows that the unconstrained empirical-likelihood dual solution $\lambda_n(0)$ satisfies $\|\lambda_n(0)\|=O_p(n^{-1/2})$.
It therefore lies in the interior of $\Lambda_\eta$ with probability tending to one, where it coincides with the maximizer $\widehat\lambda_n$ of $\ell_n$ over $\Lambda_\eta$.
This verifies \textup{(i)} and gives $T_n^{\mathrm{vec}}(\mu_0)=2\ell_n(\widehat\lambda_n)=-2\log\mathcal R_n(\mu_0)$ with probability tending to one.
On $\Lambda_\eta$,
\[
-\frac1n\nabla^2\ell_n(\lambda)=\frac1n\sum_{t=1}^n\frac{G_t(\mu_0)G_t(\mu_0)^\top}{\{1+\lambda^\top G_t(\mu_0)\}^2}\succeq\frac{\widehat\Sigma_n(\mu_0)}{(1+\eta)^2},
\]
and $\widehat\Sigma_n(\mu_0)\xrightarrow{p}\Omega$ by Lemma~\ref{lem:appendix-martingale-limits}, which verifies \textup{(ii)}.
Since $\max_t|\widehat\lambda_n^\top G_t(\mu_0)|\le C_G\|\widehat\lambda_n\|=o_p(1)$, the same lemma gives $H_n\xrightarrow{p}\Omega$, which verifies \textup{(iii)}.
The third derivatives are bounded in norm by $2nC_G^3/(1-\eta)^3$, which verifies \textup{(iv)}.
Lemma~\ref{lem:generic-laplace} therefore yields
\[
E_n^{\mathrm{vec}}(\mu_0)=\mathcal R_n(\mu_0)^{-1}\varphi_\nu(0)\Bigl(\frac{2\uppi}{n}\Bigr)^{q/2}\det(\Omega)^{-1/2}\{1+o_p(1)\}.
\]
Taking twice the logarithm proves (22).
\end{proof}

\subsection{Proof of Theorem~\ref{thm:profile-bridge}}
\label{app:subsection:A.10}
\label{app:profile-bridge}

Throughout this subsection, assume the conditions of
Theorem~4.6.
Retain the notation $\mathbb G_n=\frac1{\sqrt n}\sum_{t=1}^nG_t(\mu_0),$ $\widehat\Sigma_n(\mu)=\frac1n\sum_{t=1}^nG_t(\mu)G_t(\mu)^\top,$ and $D=-e_1,$ and define $\mathcal I=D^\top\Omega^{-1}D$ and $a^*=\Omega^{-1}D/\sqrt{\mathcal I}$.
Then $(a^*)^\top\Omega a^*=1$.
All directions are computed separately for each candidate mean, as in (11)--(12).
Throughout this subsection, $\widehat a_{t-1}$ denotes $\widehat a_{t-1}(\mu_0)$, which is $\mathcal F_{t-1}$-measurable, and $S_t(\mu_0)=\widehat a_{t-1}(\mu_0)^\top G_t(\mu_0)$.

\begin{lemma}
\label{lem:predictable-direction}
Under the predictable construction
(11)--(12),
with $\varepsilon_t>0$, $\varepsilon_t\downarrow0$,
and $C_a>\|a^*\|$, we have
\[
\widehat a_t\xrightarrow{p}a^*,
\qquad
\frac1n\sum_{t=1}^n
\|\widehat a_{t-1}-a^*\|^2
\xrightarrow{p}0.
\]
Moreover,
\[
\frac1{\sqrt n}\sum_{t=1}^nS_t(\mu_0)
=(a^*)^\top\mathbb G_n+o_p(1),
\qquad
\frac1n\sum_{t=1}^nS_t(\mu_0)^2\xrightarrow{p}1.
\]
\end{lemma}

\begin{proof}[Proof of Lemma~\ref{lem:predictable-direction}]
For every $\mu\in\Theta$, $G_s(\mu)=G_s(\mu_0)-(\mu-\mu_0)(R_s/\pi_s)e_1$, $\|G_s(\mu_0)\|\le C_G$ and $R_s/\pi_s\le\underline\pi^{-1}$ give
\[
\|\widehat\Omega_t(\mu)-\widehat\Omega_t(\mu_0)\|\le\frac{2C_G|\mu-\mu_0|}{\underline\pi}+\frac{(\mu-\mu_0)^2}{\underline\pi^2}.
\]
Since $\widehat\Omega_t(\mu_0)=\varepsilon_tI_q+\widehat\Sigma_t(\mu_0)$ with $\varepsilon_t\downarrow0$, Lemma~\ref{lem:appendix-martingale-limits} gives $\widehat\Omega_t(\mu_0)\xrightarrow{p}\Omega$.
Since $\Omega$ is positive definite, continuity of matrix inversion gives $\widetilde a_t(\mu_0)\xrightarrow{p} a^*$.
Because $C_a>\|a^*\|$, truncation is inactive with probability tending to one, hence $\widehat a_t\xrightarrow{p} a^*$.

Write $\Delta_{t-1}=\widehat a_{t-1}-a^*$.
The differences are uniformly bounded, so their convergence to zero in probability implies $\E\|\Delta_t\|^2\to0$.
Consequently,
\[
\E\left[\frac1n\sum_{t=1}^n\|\Delta_{t-1}\|^2\right]\to0.
\]
Markov's inequality proves the stated Cesaro convergence.

By Proposition~\ref{prop:estimating-equations} and predictability of $\Delta_{t-1}$, $\Delta_{t-1}^\top G_t(\mu_0)$ is a martingale difference.
Orthogonality of martingale differences and $\|G_t(\mu_0)\|\le C_G$ imply
\[
\begin{aligned}
\E\left[\left\{\frac1{\sqrt n}\sum_{t=1}^n\Delta_{t-1}^\top G_t(\mu_0)\right\}^2\right]
&=\frac1n\sum_{t=1}^n\E[(\Delta_{t-1}^\top G_t(\mu_0))^2]\\
&\le\frac{C_G^2}{n}\sum_{t=1}^n\E\|\Delta_{t-1}\|^2\to0.
\end{aligned}
\]
Therefore,
\[
\frac1{\sqrt n}\sum_{t=1}^nS_t(\mu_0)=(a^*)^\top\mathbb G_n+o_p(1).
\]

For the second moments,
\[
\begin{aligned}
\left|S_t(\mu_0)^2-\{(a^*)^\top G_t(\mu_0)\}^2\right|\le C_G^2(C_a+\|a^*\|)\|\Delta_{t-1}\|.
\end{aligned}
\]
The Cauchy--Schwarz inequality and Cesaro convergence give
\[
\frac1n\sum_{t=1}^nS_t(\mu_0)^2=(a^*)^\top\widehat\Sigma_n(\mu_0)a^*+o_p(1)\xrightarrow{p}(a^*)^\top\Omega a^*=1.
\]
\end{proof}

\begin{proof}[Proof of Theorem~\ref{thm:profile-bridge}]
For brevity, write $S_t=S_t(\mu_0)$, $\overline S_n=\frac1n\sum_{t=1}^nS_t$ and $ v_n=\frac1n\sum_{t=1}^nS_t^2$.
Lemma~\ref{lem:predictable-direction} gives
\[
\sqrt n\,\overline S_n=(a^*)^\top\mathbb G_n+o_p(1)=O_p(1),\qquad v_n\xrightarrow{p}1.
\]
Also, $|S_t|\le C_S$ almost surely.

With probability tending to one, the sample contains both positive and negative scores.
Indeed, if all scores were nonnegative, then $S_t^2\le C_SS_t$ for every $t$, so $v_n\le C_S|\overline S_n|$.
The same bound holds if all scores are nonpositive.
Since $v_n\xrightarrow{p}1$ and $\overline S_n=O_p(n^{-1/2})$, the probability that all scores have the same weak sign tends to zero.
Thus zero lies in the interior of the convex hull of $\{S_t\}_{t\le n}$ with probability tending to one.
On this event, the scalar empirical-likelihood constraint is strictly feasible, and Lemma~\ref{lem:scalar-dual} guarantees a unique interior dual solution $\widehat\beta_n$, satisfying
\[
0=\frac1n\sum_{t=1}^n
\frac{S_t}{1+\widehat\beta_nS_t}.
\]
If $\widehat\beta_n\ne0$, write $\widehat\beta_n=\kappa s$, where $\kappa=|\widehat\beta_n|$ and $s\in\{-1,1\}$.
The dual equation implies
\[
s\overline S_n=\frac{\kappa}{n}\sum_{t=1}^n\frac{S_t^2}{1+\kappa sS_t}\ge\frac{\kappa v_n}{1+\kappa C_S}.
\]
Hence
\[
\kappa\{v_n-C_S|\overline S_n|\}\le|\overline S_n|.
\]
The same resulting bound holds when $\widehat\beta_n=0$.
Since $v_n\xrightarrow{p}1$ and $\overline S_n=O_p(n^{-1/2})$, we conclude that $\widehat\beta_n=O_p(n^{-1/2})$.

Moreover,
\[
\max_{t\le n}|\widehat\beta_nS_t|
\le C_S|\widehat\beta_n|
=O_p(n^{-1/2}).
\]
Thus, with probability tending to one,
\[
\min_{t\le n}(1+\widehat\beta_nS_t)\ge\frac12.
\]
Expanding the dual equation gives
\[
0=\overline S_n-v_n\widehat\beta_n+\mathfrak r_n,
\qquad
\mathfrak r_n=
\frac1n\sum_{t=1}^n
\frac{\widehat\beta_n^2S_t^3}
{1+\widehat\beta_nS_t}.
\]
The denominator bound implies
\[
|\mathfrak r_n|
\le2C_S^3\widehat\beta_n^2
=O_p(n^{-1}),
\]
and therefore
\[
\widehat\beta_n
=
\frac{\overline S_n}{v_n}+O_p(n^{-1}).
\]

Taylor expansion of the dual objective yields
\[
\begin{aligned}
-2\log\widetilde{\mathcal R}_n(\mu_0)
&=
2\sum_{t=1}^n\log(1+\widehat\beta_nS_t)\\
&=
2n\widehat\beta_n\overline S_n
-n\widehat\beta_n^2v_n+o_p(1)\\
&=
\frac{n\overline S_n^2}{v_n}
-nv_n\left(
\widehat\beta_n-\frac{\overline S_n}{v_n}
\right)^2+o_p(1)\\
&=
\frac{n\overline S_n^2}{v_n}+o_p(1)\\
&=
\{(a^*)^\top\mathbb G_n\}^2+o_p(1).
\end{aligned}
\]
Here the summed cubic remainder is
$O_p(n|\widehat\beta_n|^3)=o_p(1)$, and the
completed-square correction is $O_p(n^{-1})$.

The proof of Theorem~\ref{thm:adaptive-wilks} gives
\[
\begin{aligned}
T_{n,\mathrm{prof}}(\mu_0)
&=\frac{(D^\top\Omega^{-1}\mathbb G_n)^2}{D^\top\Omega^{-1}D}+o_p(1)\\
&=\{(a^*)^\top\mathbb G_n\}^2+o_p(1).
\end{aligned}
\]
Combining these expansions proves (23).

Suppose additionally that $\varpi$ has a density $\varphi_\varpi$ on $I_\eta$ that is continuous and positive at zero.
Define $\ell_n(\beta)=\sum_{t=1}^n\log(1+\beta S_t)$ for $\beta\in I_\eta$, which is well defined because $1+\beta S_t\ge1-\eta>0$ for all such $\beta$.
Since $\widehat\beta_n=O_p(n^{-1/2})$ and zero is an interior point of $I_\eta$, the scalar dual solution lies in the interior of $I_\eta$ with probability tending to one.
On this event it maximizes $\ell_n$ over $I_\eta$, and $\ell_n(\widehat\beta_n)=-\log\widetilde{\mathcal R}_n(\mu_0)$.

For every $\beta\in I_\eta$,
\[
-\frac1n\ell_n''(\beta)=\frac1n\sum_{t=1}^n\frac{S_t^2}{(1+\beta S_t)^2}\ge\frac{v_n}{(1+\eta)^2}.
\]
Since $v_n\xrightarrow{p}1$, there exists a deterministic $c>0$ such that, with probability tending to one, $-\frac1n\ell_n''(\beta)\ge c$ for every $\beta\in I_\eta$.
Also,
\[
\sup_{\beta\in I_\eta}|\ell_n'''(\beta)|\le\frac{2nC_S^3}{(1-\eta)^3}.
\]
The curvature at the maximizer satisfies
\[
H_n:=-\frac1n\ell_n''(\widehat\beta_n)=\frac1n\sum_{t=1}^n\frac{S_t^2}{(1+\widehat\beta_nS_t)^2}\xrightarrow{p}1,
\]
because $\max_{t\le n}|\widehat\beta_nS_t|=o_p(1)$ and $v_n\xrightarrow{p}1$.

Set $r_n=(\log n)/\sqrt n$.
Uniform strong concavity and $\ell_n'(\widehat\beta_n)=0$ imply
\[
\ell_n(\beta)\le\ell_n(\widehat\beta_n)-\frac{cn}{2}(\beta-\widehat\beta_n)^2.
\]
Since $\varphi_\varpi$ integrates to one, the contribution outside $|\beta-\widehat\beta_n|\le r_n$ is at most
\[
\exp\{\ell_n(\widehat\beta_n)\}\exp\{-c(\log n)^2/2\},
\]
which is $o_p(\exp\{\ell_n(\widehat\beta_n)\}n^{-1/2})$.

Within this neighborhood, put $\beta=\widehat\beta_n+x/\sqrt n$.
Taylor's theorem and the third-derivative bound give
\[
\ell_n\left(\widehat\beta_n+\frac{x}{\sqrt n}\right)=\ell_n(\widehat\beta_n)-\frac12H_nx^2+o_p(1)
\]
uniformly for $|x|\le\log n$, since the remainder is $O((\log n)^3/\sqrt n)=o(1)$.
Continuity of $\varphi_\varpi$ at zero and $\widehat\beta_n=O_p(n^{-1/2})$ imply
\[
\sup_{|x|\le\log n}\left|\varphi_\varpi\left(\widehat\beta_n+\frac{x}{\sqrt n}\right)-\varphi_\varpi(0)\right|\xrightarrow{p}0.
\]
The integration neighborhood lies inside $I_\eta$ with probability tending to one.

Changing variables and using the uniform expansions, together with $\varphi_\varpi(0)>0$, yields
\[
\begin{aligned}
E_n(\mu_0)
&=\int_{I_\eta}e^{\ell_n(\beta)}\varphi_\varpi(\beta)\,d\beta\\
&=\frac{e^{\ell_n(\widehat\beta_n)}}{\sqrt n}\varphi_\varpi(0)\int_{|x|\le\log n}e^{-H_nx^2/2}\,dx\{1+o_p(1)\}\\
&=e^{\ell_n(\widehat\beta_n)}\varphi_\varpi(0)\left(\frac{2\uppi}{nH_n}\right)^{1/2}\{1+o_p(1)\}.
\end{aligned}
\]
The last equality follows because $H_n\xrightarrow{p}1$, so the omitted Gaussian tails are negligible.
Since $H_n^{-1/2}\xrightarrow{p}1$, we obtain
\[
\begin{aligned}
E_n(\mu_0)
&=e^{\ell_n(\widehat\beta_n)}\varphi_\varpi(0)\left(\frac{2\uppi}{n}\right)^{1/2}\{1+o_p(1)\}\\
&=\widetilde{\mathcal R}_n(\mu_0)^{-1}\varphi_\varpi(0)\left(\frac{2\uppi}{n}\right)^{1/2}\{1+o_p(1)\}.
\end{aligned}
\]
Taking twice the logarithm gives
\[
2\log E_n(\mu_0)=-2\log\widetilde{\mathcal R}_n(\mu_0)-\log n+2\log\varphi_\varpi(0)+\log(2\uppi)+o_p(1).
\]
Combining this with (23) establishes
\[
2\log E_n(\mu_0)+\log n-C_{\mathrm{prof}}=T_{n,\mathrm{prof}}(\mu_0)+o_p(1),\qquad C_{\mathrm{prof}}=2\log\varphi_\varpi(0)+\log(2\uppi),
\]
which proves (24).
\end{proof}

\subsection{Proof of Corollary~\ref{cor:alternatives}}
\label{app:subsection:A.11}
\label{app:alternatives}

Throughout this subsection, assume the conditions of Theorem~\ref{thm:profile-bridge}, including the assumption that $\varpi$ has a density $\varphi_\varpi$ continuous and positive at zero.
The data-generating distribution is fixed, with mean $\mu_0$.
Retain the notation $\mathbb G_n=\frac1{\sqrt n}\sum_{t=1}^nG_t(\mu_0),$ $\mathcal I=D^\top\Omega^{-1}D=V_{\EL}^{-1},$ $a^*=\frac{\Omega^{-1}D}{\sqrt{\mathcal I}},$ and $D=-e_1.$
Then $(a^*)^\top\Omega a^*=1,$ $(a^*)^\top D=\sqrt{\mathcal I},$ and $a_1^*=-\sqrt{\mathcal I}.$

\begin{lemma}
\label{lem:alternative-scores}
Let $(\mu_n)\subseteq\Theta$ be a deterministic sequence
with $\Delta_n=\mu_n-\mu_0\to0$.
Then
\[
\frac1n\sum_{t=1}^n
\E\left[
\|\widehat a_{t-1}(\mu_n)-a^*\|^2
\right]\to0.
\]
Moreover,
\[
\frac1{\sqrt n}\sum_{t=1}^nS_t(\mu_n)
=
(a^*)^\top\mathbb G_n
+\sqrt{n\mathcal I}\,\Delta_n
+o_p(1+\sqrt n|\Delta_n|)
\]
and
\[
\frac1n\sum_{t=1}^nS_t(\mu_n)^2\xrightarrow{p}1.
\]
\end{lemma}

\begin{proof}[Proof of Lemma~\ref{lem:alternative-scores}]
In this proof, write $\widehat a_{t-1}=\widehat a_{t-1}(\mu_n)$.
Since $\mu_n$ is deterministic, $\widehat a_{t-1}$ is $\mathcal F_{t-1}$-measurable.

The bound in the proof of Lemma~\ref{lem:predictable-direction} gives
\[
\sup_{t\ge1}\|\widehat\Omega_t(\mu_n)-\widehat\Omega_t(\mu_0)\|\le
d_n:=\frac{2C_G|\Delta_n|}{\underline\pi}+\frac{\Delta_n^2}{\underline\pi^2}\to0.
\]

For a positive definite matrix $M$, define
\[
v(M)=\frac{M^{-1}D}{\{D^\top M^{-1}D\}^{1/2}},
\quad
F(M)=v(M)\min\{1,C_a/\|v(M)\|\}.
\]
The map $F$ is continuous at $\Omega$.
Since $C_a>\|a^*\|$, we have $F(\Omega)=a^*$, and $\widehat a_t(\mu)=F(\widehat\Omega_t(\mu))$ for $t\ge1$.

Fix $\epsilon>0$.
By continuity, there exists $\delta>0$ such that, for every positive definite $M$,
\[
\|M-\Omega\|<\delta\quad\Longrightarrow\quad\|F(M)-a^*\|<\epsilon.
\]
For all sufficiently large $n$, $d_n<\delta/2$.
Consequently, for every $t\ge1$,
\[
\Pp\bigl(\|\widehat a_t(\mu_n)-a^*\|>\epsilon\bigr)\le\Pp\bigl(\|\widehat\Omega_t(\mu_0)-\Omega\|\ge\delta/2\bigr).
\]
Because $\widehat\Omega_t(\mu_0)\xrightarrow{p}\Omega$ as $t\to\infty$, it follows that
\[
\lim_{T\to\infty}\limsup_{n\to\infty}\sup_{t\ge T}\Pp\bigl(\|\widehat a_t(\mu_n)-a^*\|>\epsilon\bigr)=0.
\]

Set $B=C_a+\|a^*\|$.
The deterministic bound
$\|\widehat a_t(\mu_n)-a^*\|\le B$ gives
\[
\E\|\widehat a_t(\mu_n)-a^*\|^2\le\epsilon^2+B^2\Pp\bigl(\|\widehat a_t(\mu_n)-a^*\|>\epsilon\bigr).
\]
Thus, for every fixed integer $T\ge1$ and $n>T$,
\[
\begin{aligned}
\frac1n\sum_{t=1}^n\E\|\widehat a_{t-1}(\mu_n)-a^*\|^2
&\le
\frac{TB^2}{n}+\epsilon^2+
B^2\sup_{s\ge T}
\Pp\bigl(
\|\widehat a_s(\mu_n)-a^*\|>\epsilon
\bigr).
\end{aligned}
\]
Taking $\limsup_{n\to\infty}$, then letting $T\to\infty$ and finally $\epsilon\downarrow0$, proves the first claim.

Since $G_t(\mu_n)=G_t(\mu_0)-\Delta_n(R_t/\pi_t)e_1$,
\[
\frac1{\sqrt n}\sum_{t=1}^nS_t(\mu_n)
=\frac1{\sqrt n}\sum_{t=1}^n\widehat a_{t-1}^\top G_t(\mu_0)
-\sqrt n\,\Delta_n\frac1n\sum_{t=1}^n\frac{R_t}{\pi_t}\widehat a_{t-1,1}.
\]
Because $\widehat a_{t-1}-a^*$ is predictable and satisfies the first claim, the martingale argument in the proof of Lemma~\ref{lem:predictable-direction} gives that the first term equals $(a^*)^\top\mathbb G_n+o_p(1)$.

For the second term, write
\[
\frac1n\sum_{t=1}^n\frac{R_t}{\pi_t}\widehat a_{t-1,1}
=\frac1n\sum_{t=1}^n\widehat a_{t-1,1}
+\frac1n\sum_{t=1}^n\left(\frac{R_t}{\pi_t}-1\right)\widehat a_{t-1,1}.
\]
The first average converges in probability to $a_1^*$ by the direction convergence and Cauchy--Schwarz.
The second average is $O_p(n^{-1/2})$, since its summands are uniformly bounded martingale differences.
Using $a_1^*=-\sqrt{\mathcal I}$ proves the asserted score-sum expansion.

Finally,
\[
S_t(\mu_n)-(a^*)^\top G_t(\mu_0)
=(\widehat a_{t-1}-a^*)^\top G_t(\mu_0)-\Delta_n\frac{R_t}{\pi_t}\widehat a_{t-1,1},
\]
so that
\[
|S_t(\mu_n)-(a^*)^\top G_t(\mu_0)|
\le C_G\|\widehat a_{t-1}-a^*\|+\frac{C_a|\Delta_n|}{\underline\pi}.
\]
Both scores are uniformly bounded.
The direction convergence, Markov's inequality, and Cauchy--Schwarz consequently give
\[
\frac1n\sum_{t=1}^nS_t(\mu_n)^2
=(a^*)^\top\widehat\Sigma_n(\mu_0)a^*+o_p(1)
\xrightarrow{p}1.
\]
\end{proof}

\begin{proof}[Proof of Corollary~\ref{cor:alternatives}]
Write $\Delta_n=\mu_n-\mu_0,$ $S_{t,n}=S_t(\mu_n),$ $\overline S_n=\frac1n\sum_{t=1}^nS_{t,n},$ and $v_n=\frac1n\sum_{t=1}^nS_{t,n}^2.$
By Lemma~\ref{lem:alternative-scores},
\[
\sqrt n\,\overline S_n=(a^*)^\top\mathbb G_n+\sqrt{n\mathcal I}\,\Delta_n+o_p(1+\sqrt n|\Delta_n|),
\qquad
v_n\xrightarrow{p}1.
\]
Also, $|S_{t,n}|\le C_S$.

Since $\Delta_n\to0$, the score expansion implies $\overline S_n=o_p(1)$.
If all scores had the same weak sign, then $v_n\le C_S|\overline S_n|$.
Thus the sample contains both positive and negative scores with probability tending to one.
Let $\widehat\beta_n$ be the unique interior scalar dual solution:
\[
0=\frac1n\sum_{t=1}^n\frac{S_{t,n}}{1+\widehat\beta_nS_{t,n}}.
\]
The scalar dual bound established in the proof of Theorem~\ref{thm:profile-bridge} gives
\[
|\widehat\beta_n|\{v_n-C_S|\overline S_n|\}\le|\overline S_n|.
\]
Hence
\[
\widehat\beta_n=O_p(|\overline S_n|)=o_p(1).
\]
All denominators are therefore at least $1/2$ with probability tending to one.
Expanding the dual equation yields
\[
0=\overline S_n-v_n\widehat\beta_n+O_p(\widehat\beta_n^2),
\qquad\widehat\beta_n=\frac{\overline S_n}{v_n}+O_p(\overline S_n^2).
\]

Define $\ell_n(\beta)=\sum_{t=1}^n\log(1+\beta S_{t,n}).$
Taylor expansion and completion of the square give
\[
\begin{aligned}
2\ell_n(\widehat\beta_n)
&=2n\widehat\beta_n\overline S_n-nv_n\widehat\beta_n^2+O_p(n|\widehat\beta_n|^3)\\
&=\frac{n\overline S_n^2}{v_n}-nv_n\left(\widehat\beta_n-\frac{\overline S_n}{v_n}\right)^2+O_p(n|\widehat\beta_n|^3).
\end{aligned}
\]

Since $\widehat\beta_n=o_p(1)$, it lies in the interior of $I_\eta$ with probability tending to one.
For every $\beta\in I_\eta$,
\[
-\frac1n\ell_n''(\beta)\ge\frac{v_n}{(1+\eta)^2},
\qquad|\ell_n'''(\beta)|\le\frac{2nC_S^3}{(1-\eta)^3}.
\]
Moreover,
\[
H_n:=-\frac1n\ell_n''(\widehat\beta_n)=\frac1n\sum_{t=1}^n\frac{S_{t,n}^2}{(1+\widehat\beta_nS_{t,n})^2}\xrightarrow{p}1.
\]

The Laplace argument in the proof of Theorem~\ref{thm:profile-bridge} applies around this
moving maximizer.
Indeed, its neighborhood of radius $(\log n)/\sqrt n$ lies inside $I_\eta$ with probability
tending to one.
The uniform curvature bound controls the tails, and the third-derivative bound gives a local remainder $O((\log n)^3/\sqrt n)=o(1)$.
Since $\widehat\beta_n=o_p(1)$, continuity of $\varphi_\varpi$ at zero gives uniform replacement of the local density by $\varphi_\varpi(0)$.
Consequently,
\[
E_n(\mu_n)=e^{\ell_n(\widehat\beta_n)}\varphi_\varpi(0)\left(\frac{2\uppi}{nH_n}\right)^{1/2}\{1+o_p(1)\}.
\]
Using $H_n\xrightarrow{p}1$ yields
\[
2\log E_n(\mu_n)+\log n-C_{\mathrm{prof}}=2\ell_n(\widehat\beta_n)+o_p(1),
\]
where $C_{\mathrm{prof}}=2\log\varphi_\varpi(0)+\log(2\uppi).$

Suppose $\Delta_n=u/\sqrt n$ for fixed $u\in\R$.
Lemma~\ref{lem:alternative-scores} gives
\[
\sqrt n\,\overline S_n=(a^*)^\top\mathbb G_n+u\sqrt{\mathcal I}+o_p(1).
\]
By Lemma~\ref{lem:appendix-martingale-limits} and $(a^*)^\top\Omega a^*=1$,
\[
\sqrt n\,\overline S_n\xrightarrow{d} N(u\sqrt{\mathcal I},1).
\]
Here $\overline S_n=O_p(n^{-1/2})$, so the scalar dual expansion implies
\[
2\ell_n(\widehat\beta_n)=\frac{n\overline S_n^2}{v_n}+o_p(1)\xrightarrow{d}\chi_1^2(u^2\mathcal I).
\]
Combining this limit with the mixture approximation and $\mathcal I=V_{\EL}^{-1}$ proves
\[
2\log E_n(\mu_n)+\log n-C_{\mathrm{prof}}
\xrightarrow{d}\chi_1^2(u^2/V_{\EL}).
\]

Suppose $\Delta_n\to0$ and $n\Delta_n^2/\log n\to\infty$.
Then $\sqrt n|\Delta_n|\to\infty$, and Lemma~\ref{lem:alternative-scores} implies
\[
\frac{\overline S_n}{\Delta_n}\xrightarrow{p}\sqrt{\mathcal I}.
\]
In particular, $\overline S_n=O_p(|\Delta_n|)$ and $\widehat\beta_n=O_p(|\Delta_n|)$.
The scalar dual expansion gives
\[
2\ell_n(\widehat\beta_n)=\frac{n\overline S_n^2}{v_n}+O_p(n|\Delta_n|^3+n\Delta_n^4).
\]
Since $\Delta_n\to0$ and $v_n\xrightarrow{p}1$,
\[
2\ell_n(\widehat\beta_n)=n\mathcal I\Delta_n^2\{1+o_p(1)\}.
\]
Thus the mixture approximation yields
\[
2\log E_n(\mu_n)=n\mathcal I\Delta_n^2\{1+o_p(1)\}-\log n+C_{\mathrm{prof}}+o_p(1).
\]
Finally, $\log n=o(n\Delta_n^2)$ and $\mathcal I=V_{\EL}^{-1}$ imply
\[
2\log E_n(\mu_n)=\frac{n\Delta_n^2}{V_{\EL}}\{1+o_p(1)\},
\]
which proves the second assertion.
\end{proof}

\subsection{Proof of Proposition~\ref{prop:evidence-bound}}
\label{app:subsection:A.12}
\label{app:evidence-bound}

\begin{proof}
Write $V=V_{\mathrm{oracle}}$ and define $s_Z(z)=\frac{m(z)-\mu_0}{V}$ and $s_c(y,z)=\frac{y-m(z)}{\pi^*(z)V}$.
These scores satisfy $\E_0[s_Z]=0$, $\E_0[s_c\mid Z]=0$ and $|s_Z+s_c|\le\frac{2}{\underline\pi V}.$
Thus $dP_\delta/dP_0=1+\delta(s_Z+s_c)$ is a probability density in $[1/2,3/2]$ for $|\delta|\le\underline\pi V/4$.
Moreover,
\[
\E_0[Y(s_Z+s_c)]=\frac{\Var_0\{m(Z)\}+\E_0[\sigma^2(Z)/\pi^*(Z)]}{V}=1,
\]
so $\mu(P_\delta)=\mu_0+\delta$.

Let $P_\delta^{(n)}$ denote the law of the observed history through time $n$, including the initial information.
Using the same decision rules and the same independent initial-information law under both distributions gives
\[
\frac{dP_\delta^{(n)}}{dP_0^{(n)}}=\prod_{t=1}^n(1+\delta q_t),
\qquad q_t=s_Z(Z_t)+R_ts_c(Y_t,Z_t).
\]
Indeed, integrating out an unaudited label leaves the factor $1+\delta s_Z(Z_t)$.
Under $P_0$,
\[
\E_0[q_t\mid\mathcal F_{t-1}]=0,
\]
and
\[
I_t:=\E_0[q_t^2\mid\mathcal F_{t-1}]=\E_0[s_Z(Z)^2]+\E_0\left[\frac{\pi_t\sigma^2(Z_t)}{\pi^*(Z_t)^2V^2}\,\middle|\,\mathcal F_{t-1}\right].
\]

The clipped oracle rule satisfies, for $p\in[\underline\pi,1]$,
\[
\{p-\pi^*(z)\}\frac{\sigma^2(z)}{\pi^*(z)^2}\le(\tau^*)^2\{p-\pi^*(z)\}.
\]
Equality holds where $\underline\pi<\pi^*(z)<1$; where $\pi^*(z)=1$, $p-\pi^*(z)\le0$ and $\sigma(z)\ge\tau^*$; where $\pi^*(z)=\underline\pi$, $p-\pi^*(z)\ge0$ and $\sigma(z)\le\tau^*\underline\pi$.
We also use $\E_0[\pi^*(Z_t)\mid\mathcal F_{t-1}]=\rho$ and $\E_0[s_Z^2]+\E_0[\sigma^2/(\pi^*V^2)]=1/V$.
Taking conditional expectations with $p=\pi_t$ yields
\[
I_t\le\frac1V+\frac{(\tau^*)^2}{V^2}\{\E_0[\pi_t\mid\mathcal F_{t-1}]-\rho\}.
\]
Since $|q_t|\le2/(\underline\pi V)$, we have $|\delta q_t|\le1/2$ for $|\delta|\le\underline\pi V/4$, so the inequality
\[
-\log(1+x)\le-x+\frac{x^2}{2}+|x|^3,\qquad |x|\le1/2,
\]
and the chain rule for relative entropy give
\begin{equation}
\begin{aligned}
\operatorname{KL}
(P_0^{(n)}\Vert P_\delta^{(n)})\le{}\frac{\delta^2}{2}\left(1+\frac{4|\delta|}{\underline\pi V}\right)\times\left[\frac nV+\frac{(\tau^*)^2}{V^2}\sum_{t=1}^n\{\E_0[\pi_t]-\rho\}\right].
\end{aligned}
\label{eq:supp-evidence-kl}
\end{equation}
Consequently, under the long-run expected budget,
\begin{equation}
\operatorname{KL}(P_0^{(n)}\Vert P_{\delta_n}^{(n)})\le\frac{n\delta_n^2}{2V}\{1+o(1)\}.
\label{eq:supp-evidence-kl-asym}
\end{equation}

For part~(i), put
$E=\mathcal E_n(\mu_0+\delta)$ and
$\Lambda=dP_\delta^{(n)}/dP_0^{(n)}$.
For fixed $n$ and sufficiently small $\delta$,
$\Lambda$ is bounded above and away from zero.
Thus $\E_0[E\Lambda]\le1$ implies
$\E_0[(\log E)^+]<\infty$.
If $\E_0[\log E]=-\infty$, the desired inequality is
immediate.
Otherwise, Jensen's inequality gives
\[
\begin{aligned}
\E_0[\log E]
&=
\E_0[\log(E\Lambda)]+\E_0[-\log\Lambda]\\
&\le
\log\E_{P_\delta}[E]
+\operatorname{KL}
(P_0^{(n)}\Vert P_\delta^{(n)})\\
&\le
\operatorname{KL}
(P_0^{(n)}\Vert P_\delta^{(n)}).
\end{aligned}
\]
Part~(i) follows from \eqref{eq:supp-evidence-kl-asym}.

For part~(ii), fix $\epsilon>0$ and let $c_n=\frac{(1+\epsilon)n\delta_n^2}{2V}$ and $A_n=\{\log\mathcal E_n(\mu_0+\delta_n)\ge c_n\}$.
Validity under $P_{\delta_n}$ and Markov's inequality imply $P_{\delta_n}(A_n)\le e^{-c_n}.$
For Bernoulli relative entropy,
\[
\operatorname{kl}(p,q)
\ge p\log(1/q)-\log2,
\]
with the usual extended-value conventions at the boundaries.
The data-processing inequality therefore yields
\[
\begin{aligned}
P_0(A_n)c_n-\log2
&\le
\operatorname{kl}
\{P_0(A_n),P_{\delta_n}(A_n)\}\\
&\le
\operatorname{KL}
(P_0^{(n)}\Vert P_{\delta_n}^{(n)})\\
&\le
\frac{n\delta_n^2}{2V}\{1+o(1)\}.
\end{aligned}
\]
Dividing by $c_n$ and using $n\delta_n^2\to\infty$
proves part~(ii).
\end{proof}

\subsection{Scalar empirical-likelihood dual}
\label{app:subsection:A.13}
\label{app:scalar-dual}

\begin{lemma}
\label{lem:scalar-dual}
Fix $\mu\in\Theta$ and $n\ge1$, and let $\mathcal K_n(\mu)=\{\beta\in\R:1+\beta S_t(\mu)>0\text{ for all }t\le n\}$.
If $0$ lies in the interior of the convex hull of $\{S_t(\mu)\}_{t\le n}$, then
$\sup_{\beta\in\mathcal K_n(\mu)}\sum_{t=1}^n\log\{1+\beta S_t(\mu)\}$
is attained at a unique $\widehat\beta_n(\mu)$ and equals $-\log\widetilde{\mathcal R}_n(\mu)$.
Moreover, $I_\eta\subseteq\mathcal K_n(\mu)$, and if $\widehat\beta_n(\mu)\in I_\eta$, then
$\sup_{\beta\in I_\eta}\sum_{t=1}^n\log\{1+\beta S_t(\mu)\}=-\log\widetilde{\mathcal R}_n(\mu)$.
\end{lemma}

\begin{proof}
Write $S_t=S_t(\mu)$ and $g(\beta)=\sum_{t=1}^n\log(1+\beta S_t)$ for $\beta\in\mathcal K_n(\mu)$.
Since $0$ lies in the interior of the convex hull of $\{S_t\}_{t\le n}$, some $S_t$ are positive and some are negative.
Hence $\mathcal K_n(\mu)$ is a bounded open interval containing zero, and $g(\beta)\to-\infty$ at both endpoints.
Since $g''(\beta)=-\sum_{t}S_t^2/(1+\beta S_t)^2<0$, $g$ has a unique maximizer $\widehat\beta_n$, characterized by $\sum_{t}S_t/(1+\widehat\beta_nS_t)=0$.

For feasible weights $(p_t)$ and any $\beta\in\mathcal K_n(\mu)$, the constraints give $\sum_tp_t(1+\beta S_t)=1$, and the arithmetic--geometric mean inequality yields $\prod_t\{np_t(1+\beta S_t)\}\le1$.
Hence $\log\prod_tnp_t\le-g(\beta)$, so $-\log\widetilde{\mathcal R}_n(\mu)\ge g(\widehat\beta_n)$.
Conversely, $\widehat p_t=\{n(1+\widehat\beta_nS_t)\}^{-1}>0$ satisfies $\sum_t\widehat p_tS_t=0$ by the first-order condition and
$\sum_t\widehat p_t=n^{-1}\sum_t\{1-\widehat\beta_nS_t/(1+\widehat\beta_nS_t)\}=1$,
with $\log\prod_tn\widehat p_t=-g(\widehat\beta_n)$.
Therefore $-\log\widetilde{\mathcal R}_n(\mu)=g(\widehat\beta_n)$.

Finally, $|\beta S_t|\le\eta<1$ for $\beta\in I_\eta$, so $I_\eta\subseteq\mathcal K_n(\mu)$, and the last claim follows.
\end{proof}

\section{Additional Experimental Details}
\label{app:section:B}
\subsection{Implementation Details}
\label{app:subsection:B.1}
We set $\underline\pi=0.05$ ($\underline\pi=0.025$ for $\rho=0.05$, so that $\underline\pi<\rho$), $C_a=5$, $\eta=0.9$, $\varepsilon_t=0.1/\sqrt{t}$ and $n_{\mathrm{up}}=50$, take $\varpi$ uniform over $1001$ equally spaced betting parameters in $[-1,1]$, and evaluate confidence sequences on a grid of $401$ equally spaced candidate means in $[0,1]$.
Each betting fraction is truncated to $|\beta|\le\eta/\widehat C_t(\mu)$ with
$\widehat C_t(\mu)=\max_{w\in\{0,1,1/\underline\pi\},\,y\in\{0,1\},\,f\in\{0,1\}}|\widehat a_{t-1}(\mu)^\top(w(y-\mu),\,w-1,\,(w-1)f)^\top|$, with the same $\underline\pi$ for all audit rules.
For each fixed betting-grid point $\beta_j$, let $\beta_{j,t}(\mu)$ denote its predictable truncation.
We accumulate wealth as $\prod_{s\le t}\{1+\beta_{j,s}(\mu)S_s(\mu)\}$.
Each component, and hence their finite average, is a unit-mean nonnegative martingale at $\mu_0$.
Exact anytime validity therefore holds before numerical inversion.
The reported intervals are numerical approximations obtained by evaluating candidate means on a finite grid.

For the real-data applications, the pre-audit information $Z_t$ consists of, for each of the three low-cost judges, the verbalized probabilities of all classes, renormalized to sum to one, together with their entropy.
No additional text features are used.

\subsection{Baselines}
\label{app:subsection:B.2}
All learning-based baselines use the same IPW-weighted random forest as AAELI, refitted every $n_{\mathrm{up}}=50$ items, and the same audit uniforms.
\begin{itemize}
\item \textbf{PPI} \citep{waudby2024time,zrnic2024active}: uniform auditing with $\pi_t=\rho$.
The confidence sequence applies the hedged-capital betting construction \citep{waudbysmith2024betting} to the AIPW increments $\Delta_t=f_t+R_t(Y_t-f_t)/\pi_t$, with known bounds $\Delta_t\in[1-1/\rho,\,1/\rho]$.
\item \textbf{ASI} \citep{zrnic2024active}: uncertainty $u_t=2\min\{f_t,1-f_t\}$ and $\pi_t=(1-\theta)\,\clip_{[0,1]}(u_t/\tau_u)+\theta\rho$ with mixing weight $\theta=0.5$, where the threshold $\tau_u$ is calibrated on unlabeled data so that $\E[\pi_t]=\rho$.
The same betting construction is applied to $\Delta_t$ with bounds determined by $\pi_t\ge\theta\rho$.
\item \textbf{RASI} \citep{li2026robust}: audit probabilities on the geometric path between the ASI rule and uniform auditing, with the path parameter chosen by the $\ell_2$-robust criterion and the radius by $5$-fold cross-validation, after a burn-in of $100$ fully labeled items.
We extend this audit design to a confidence sequence by applying the hedged-capital betting construction \citep{waudbysmith2024betting} to the AIPW increments $\Delta_t$.
Let $\underline\pi_{\mathrm R}>0$ denote the deterministic audit-probability floor used by this baseline.
We use the uniform bounds $\Delta_t\in[1-1/\underline\pi_{\mathrm R},\,
1/\underline\pi_{\mathrm R}]$.
These bounds hold uniformly over the current pre-audit information, so the final betting fractions depend only on past observations and the candidate mean.
\item \textbf{FTRL} \citep{sfyraki2026revisiting}: $\pi_t=\clip_{[\underline\pi_{\mathrm F},\rho]} (\gamma_{\mathrm F}\sum_{s<t}\Phi_s/\pi_s^2)$ with lower bound $\underline\pi_{\mathrm F}=\rho/8$ and step size $\gamma_{\mathrm F}=n^{-1/2}$, where $\Phi_s$ is a linear prediction of the squared residual fitted on half of the labels.
We extend this audit design to a confidence sequence by applying the same hedged-capital betting construction to $\Delta_t$.
In this implementation, $\pi_t$ is selected using only past observations, before the current pre-audit information $Z_t$ is observed.
We therefore use the history-measurable bounds $\Delta_t\in[1-1/\pi_t,\,1/\pi_t]$, which hold uniformly over the current predictor value.
\item \textbf{ABPPI++} \citep{kilian2026anytime,angelopoulos2026ppi++}: uniform auditing, the PPI++ estimator with the average judge score as predictor, and a Laplace prior with scale $(\rho n)^{-1/2}$; error level $\alpha$ is split as $3\alpha/4$ for the rectifier and $\alpha/4$ for the measure of fit.
\item \textbf{BAI} \citep{chen2026balanced}: a predictor trained on $100$ fully labeled pilot items (counted in the budget), audit probabilities mixing uncertainty-proportional and uniform sampling with weight $0.5$, cube sampling balanced on the uncertainty, and a terminal Wald interval whose variance adds the population term $\Var(Y)/N$ because the target is $\E[Y]$.
\end{itemize}

\section{Additional Experimental Results}
\label{app:section:C}

\begin{figure}[H]
\centering
\includegraphics[width=0.7\textwidth]{Figure/Legend.pdf}
\subfloat[Confidence Sequences]{%
    \includegraphics[width=0.33\textwidth]{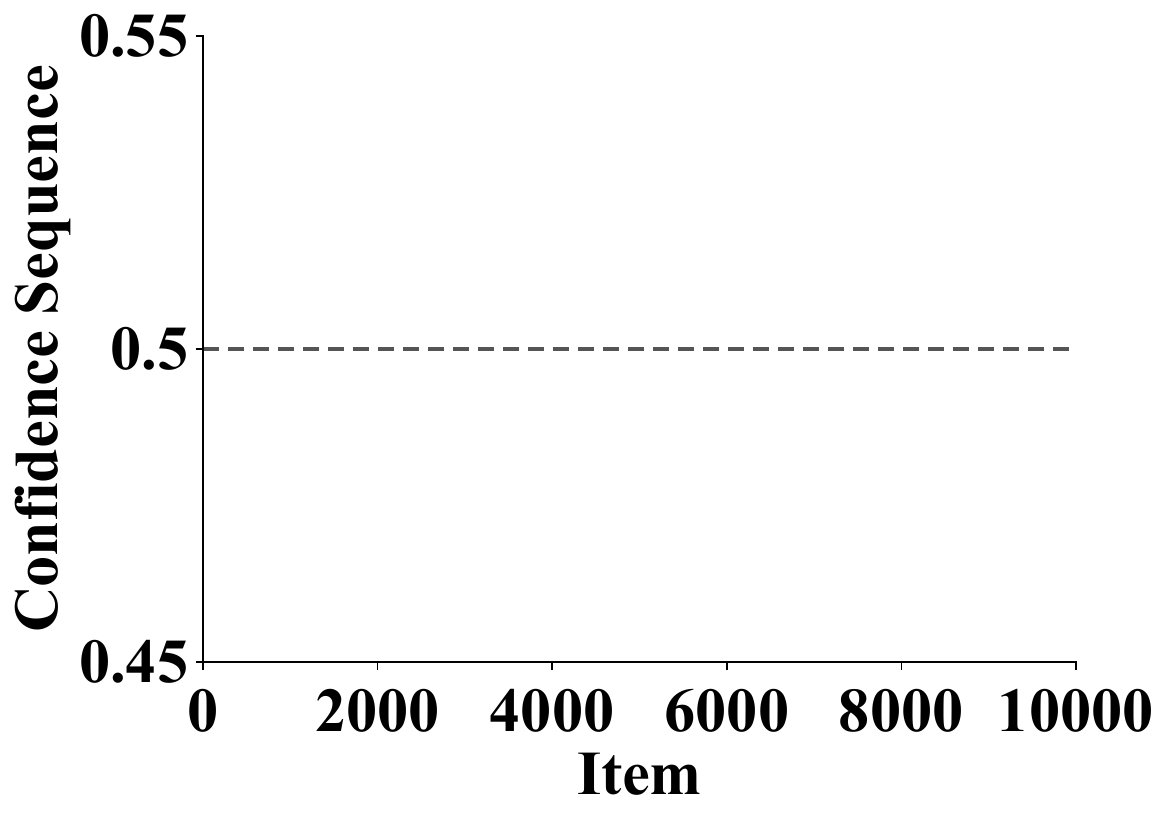}
}
\subfloat[Anytime Coverage]{%
    \includegraphics[width=0.33\textwidth]{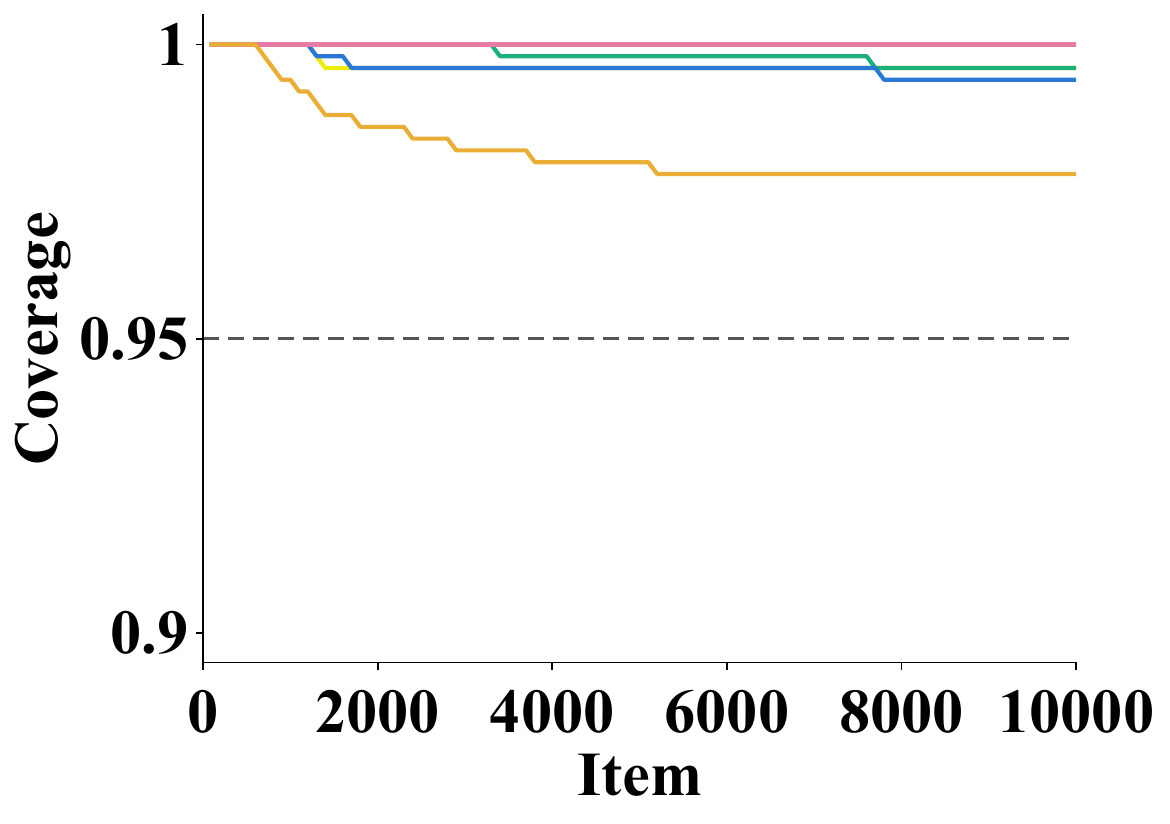}
}
\subfloat[Width]{%
    \includegraphics[width=0.33\textwidth]{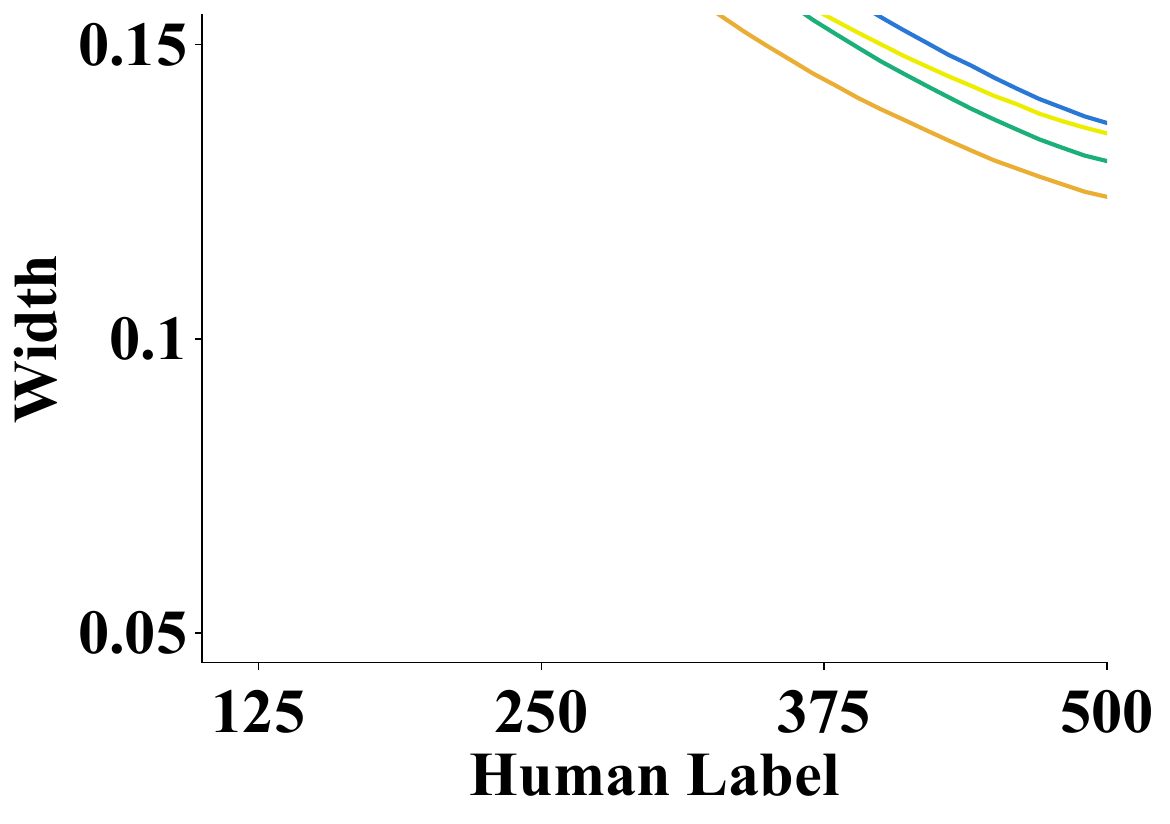}
}
\caption{Ablation on the human audit budget: $\rho=0.05$, corresponding to 500
expected human labels.}
\label{figure:a6}
\end{figure}

\begin{figure}[H]
\centering
\includegraphics[width=0.7\textwidth]{Figure/Legend.pdf}
\subfloat[Confidence Sequences]{%
    \includegraphics[width=0.33\textwidth]{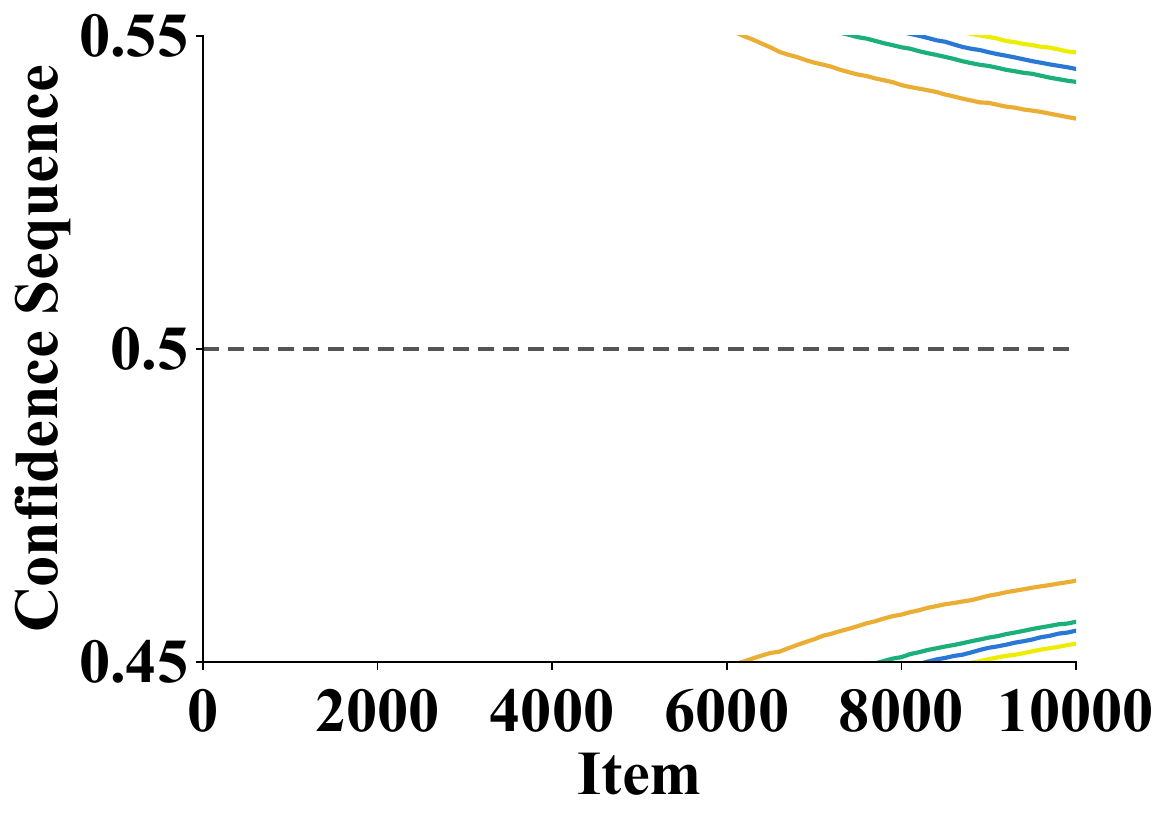}
}
\subfloat[Anytime Coverage]{%
    \includegraphics[width=0.33\textwidth]{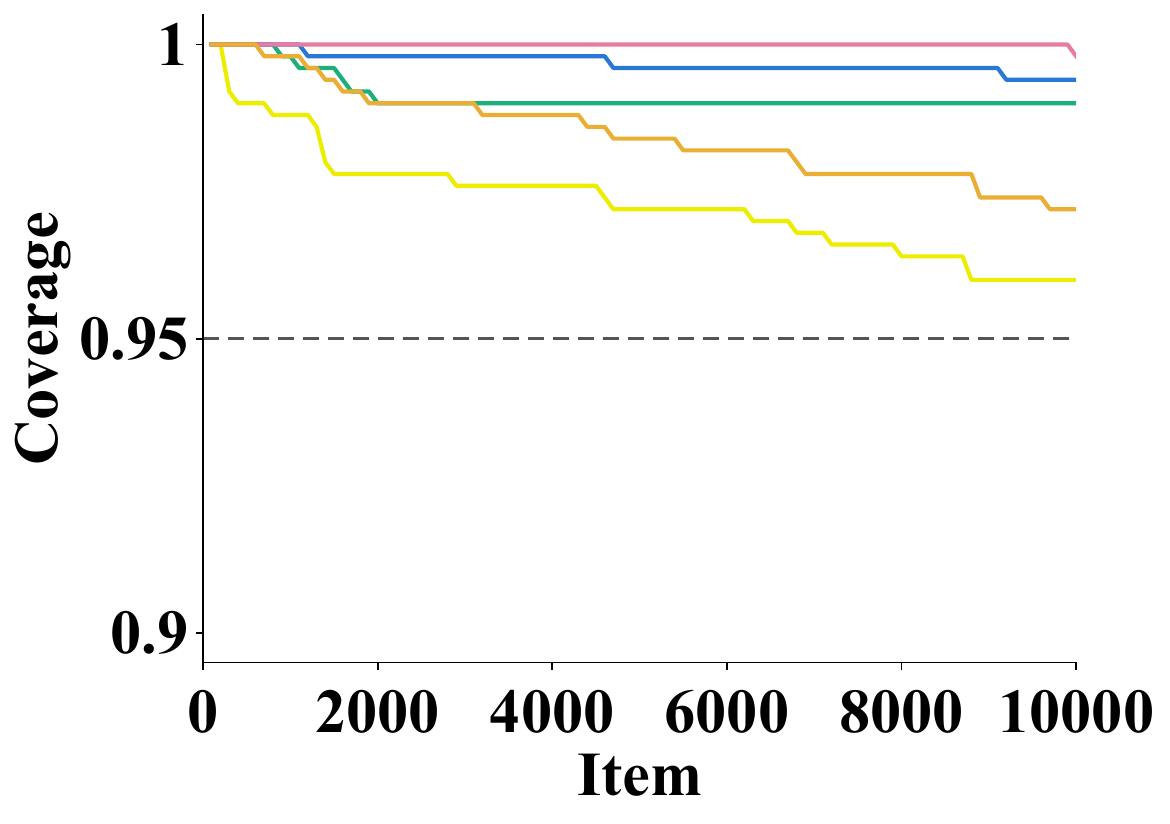}
}
\subfloat[Width]{%
    \includegraphics[width=0.33\textwidth]{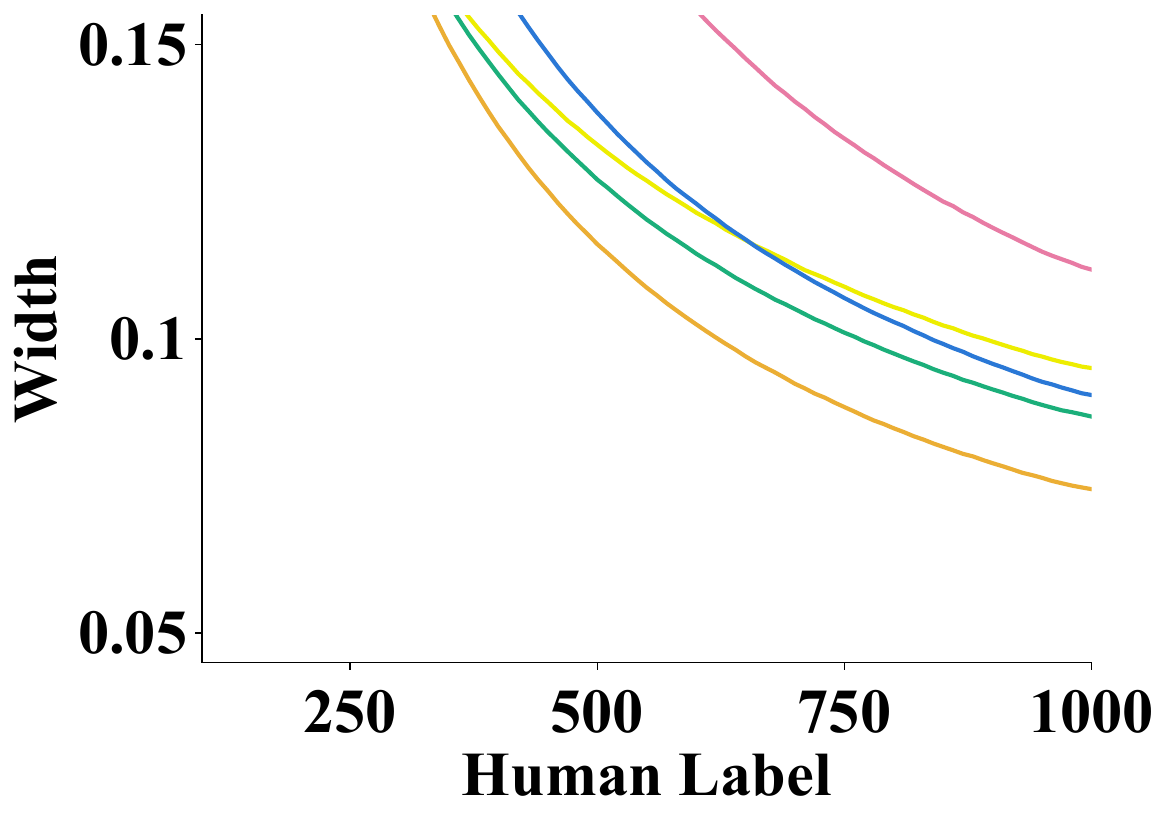}
}
\caption{Ablation on the human audit budget: $\rho=0.1$, corresponding to 1,000
expected human labels.}
\label{figure:a7}
\end{figure}

\begin{figure}[H]
\centering
\includegraphics[width=0.7\textwidth]{Figure/Legend.pdf}
\subfloat[Confidence Sequences]{%
    \includegraphics[width=0.33\textwidth]{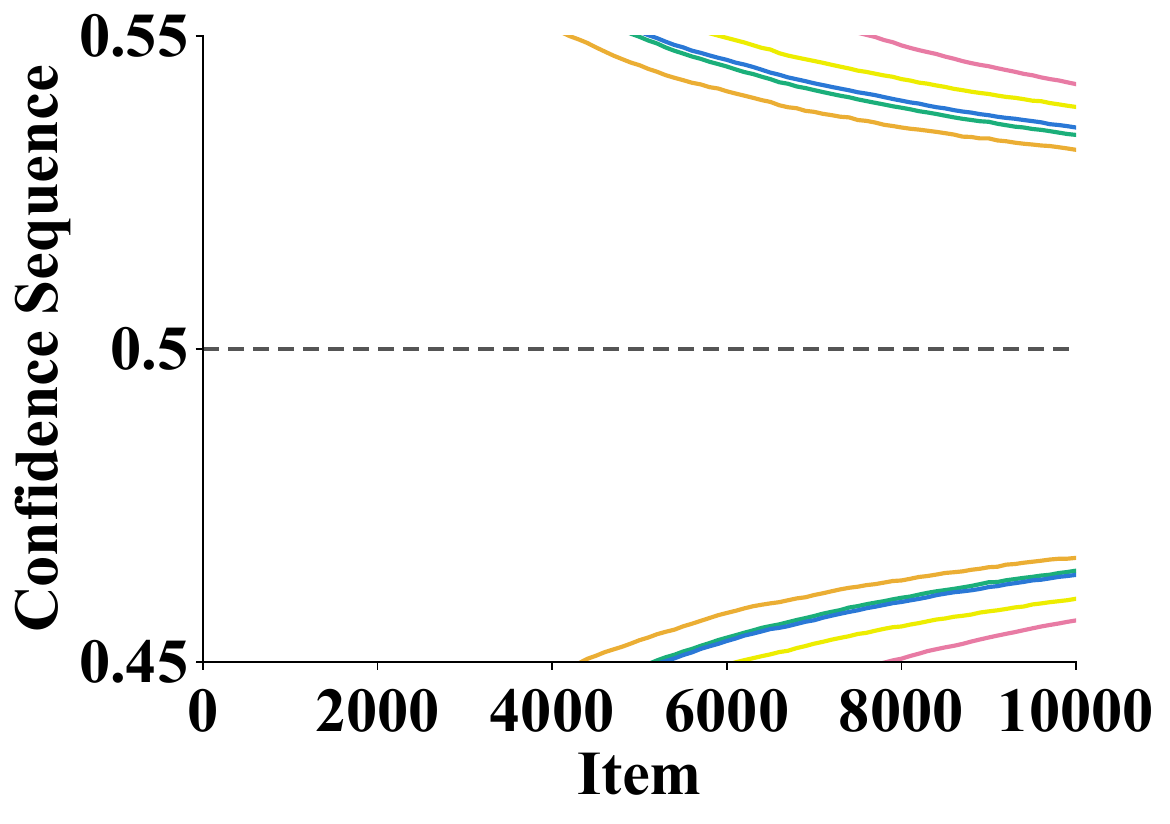}
}
\subfloat[Anytime Coverage]{%
    \includegraphics[width=0.33\textwidth]{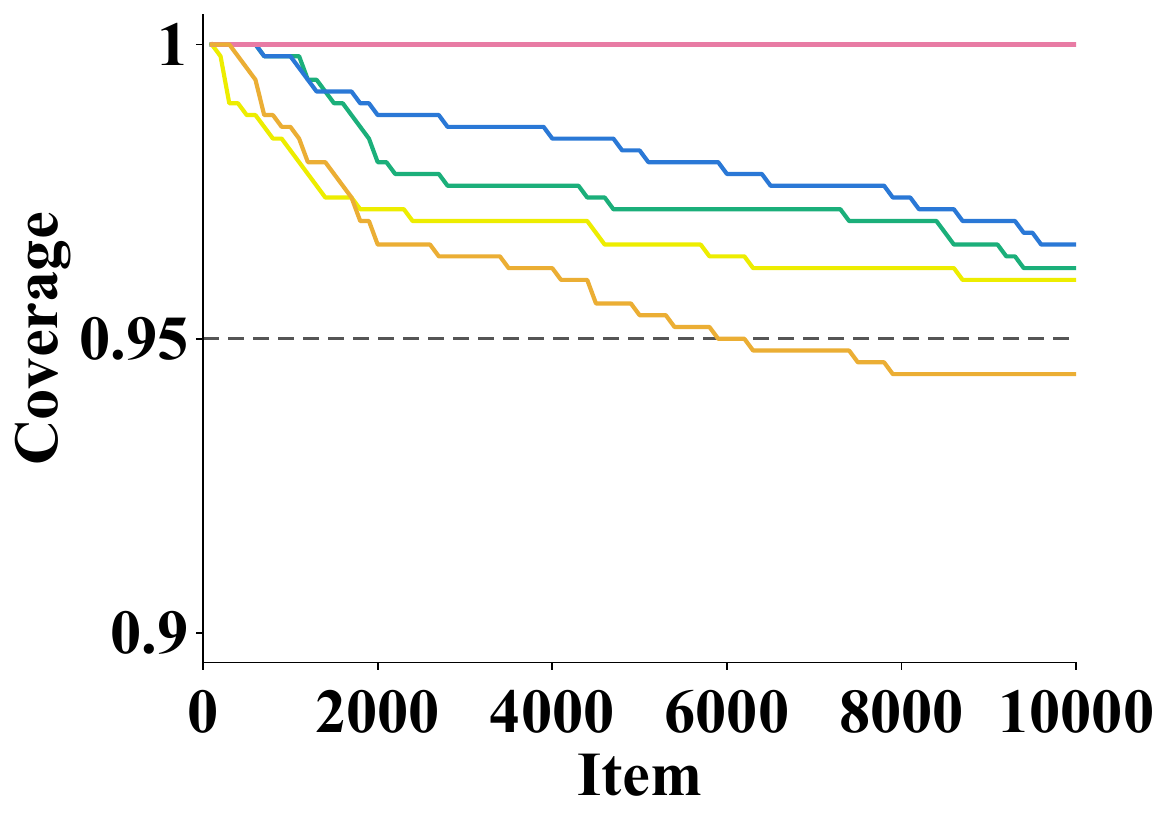}
}
\subfloat[Width]{%
    \includegraphics[width=0.33\textwidth]{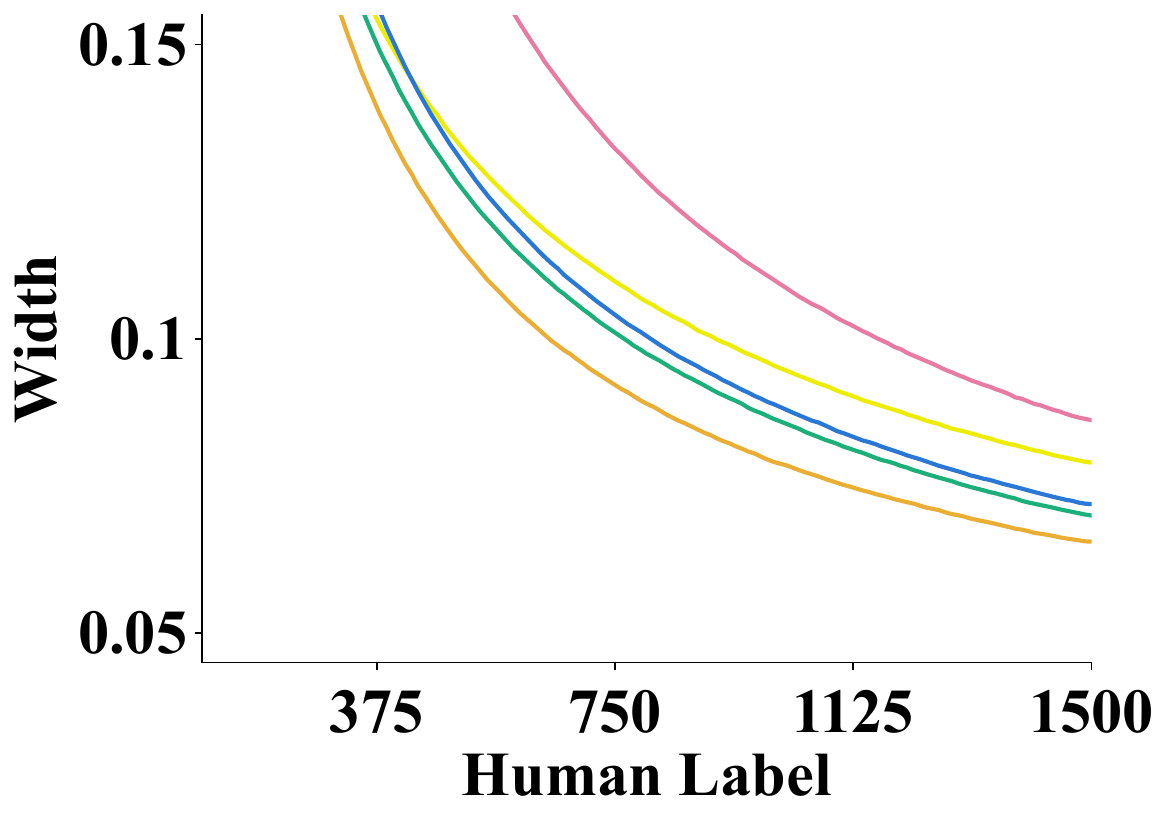}
}
\caption{Ablation on the human audit budget: $\rho=0.15$, corresponding to 1,500
expected human labels.}
\label{figure:a8}
\end{figure}

\begin{figure}[H]
\centering
\includegraphics[width=0.7\textwidth]{Figure/Legend.pdf}
\subfloat[Confidence Sequences]{%
    \includegraphics[width=0.33\textwidth]{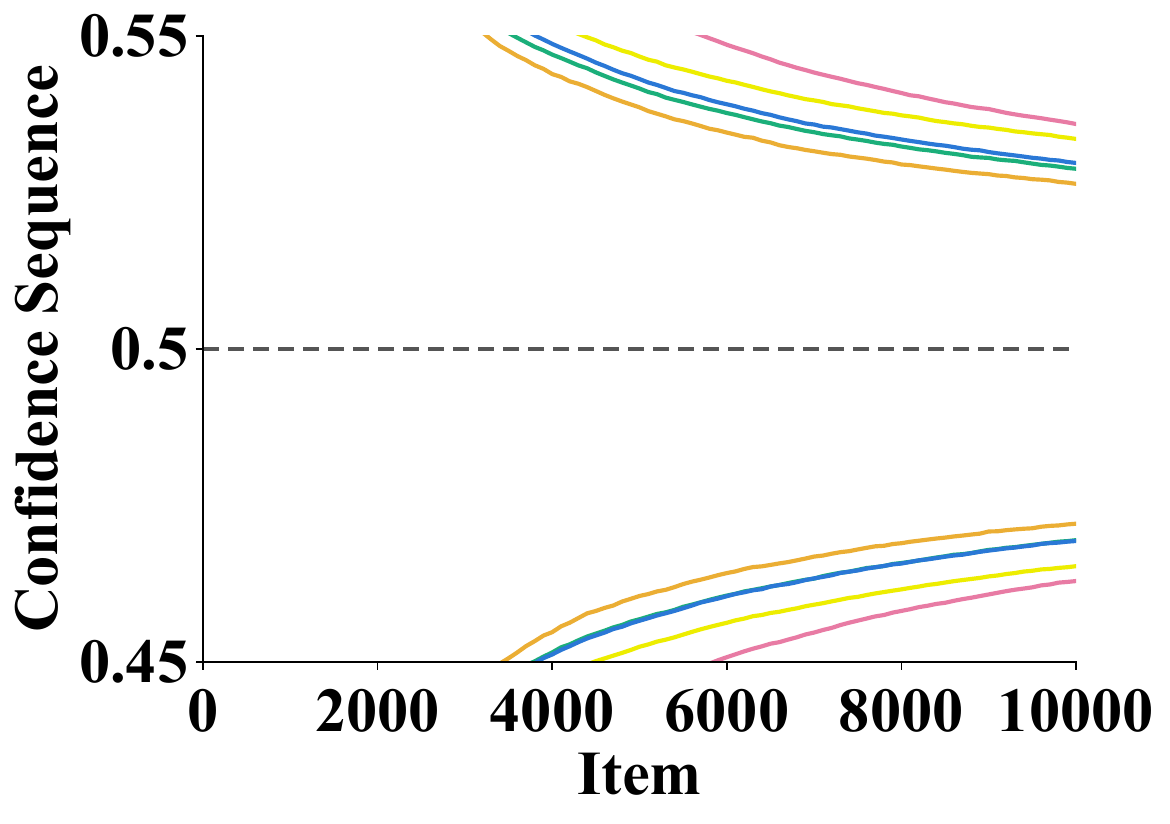}
}
\subfloat[Anytime Coverage]{%
    \includegraphics[width=0.33\textwidth]{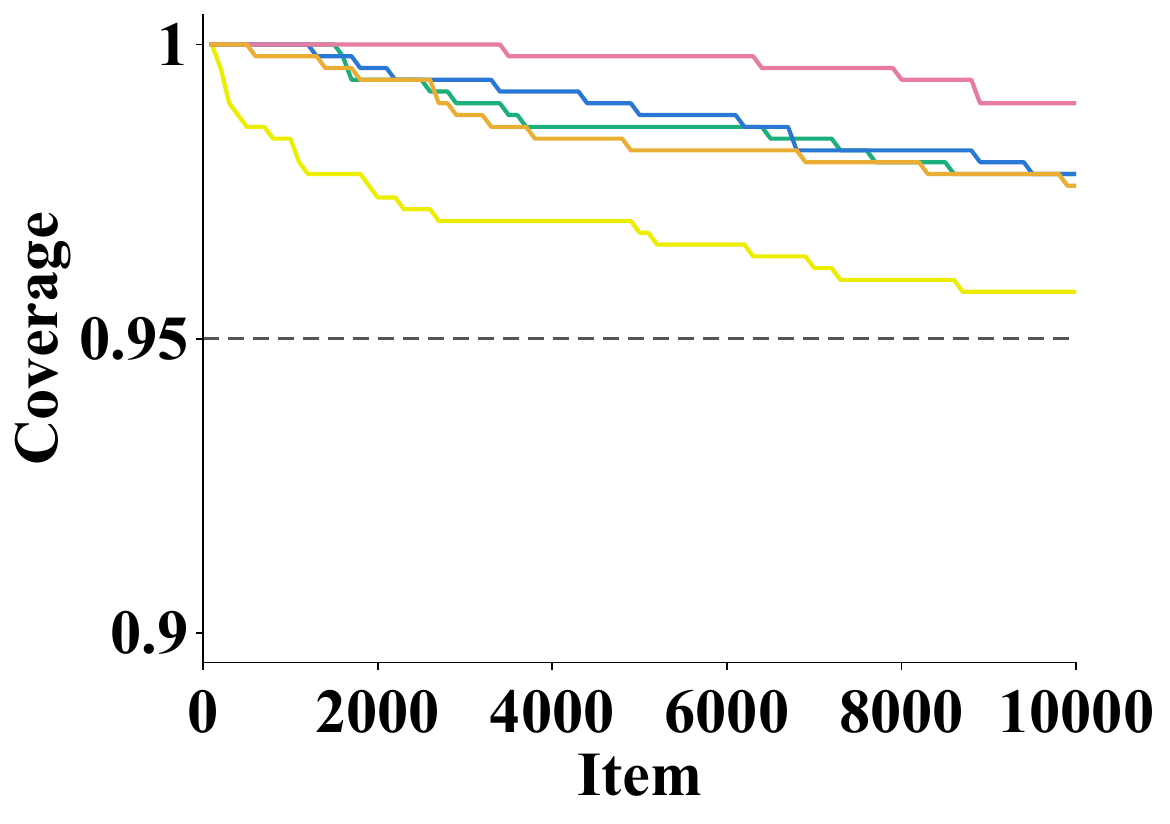}
}
\subfloat[Width]{%
    \includegraphics[width=0.33\textwidth]{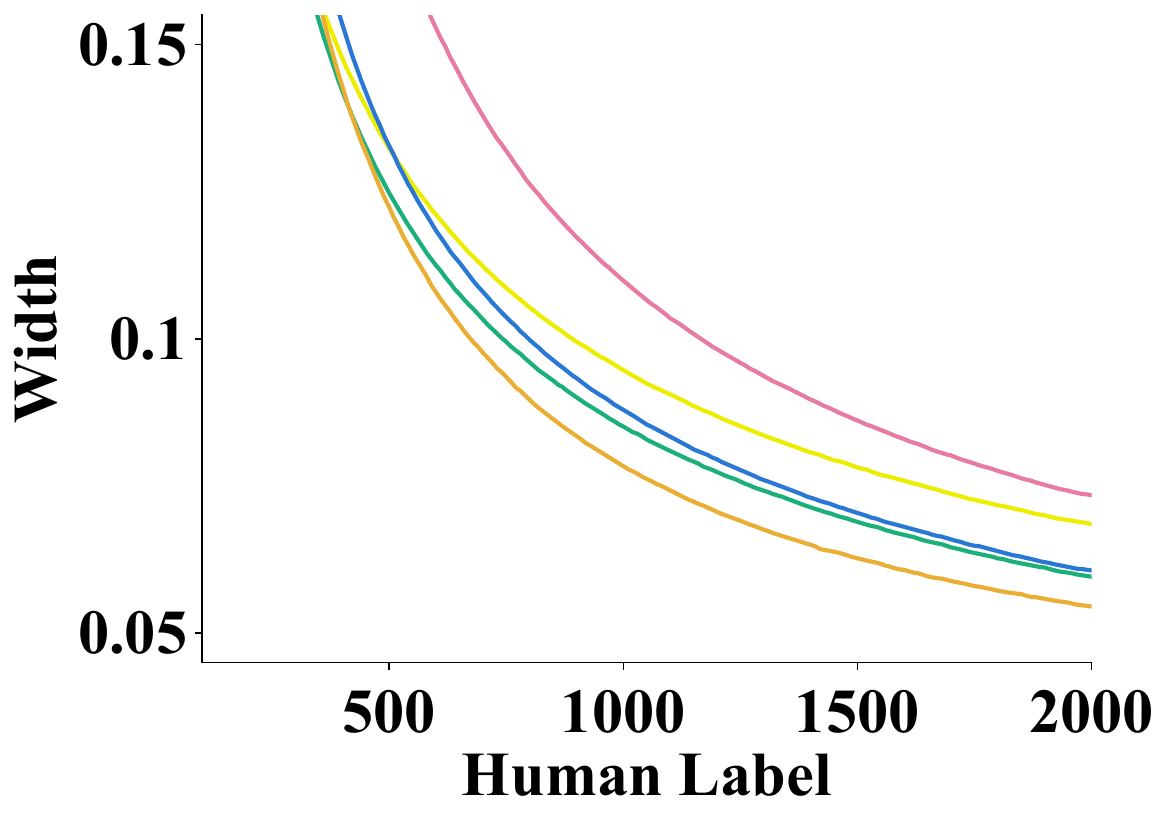}
}
\caption{Robustness to the nuisance learner: $\widehat m$ and $\widehat\sigma$ are fitted by a multilayer perceptron (MLP) with two hidden layers.}
\label{figure:a9}
\end{figure}

\begin{figure}[H]
\centering
\includegraphics[width=0.7\textwidth]{Figure/Legend.pdf}
\subfloat[Confidence Sequences]{%
    \includegraphics[width=0.33\textwidth]{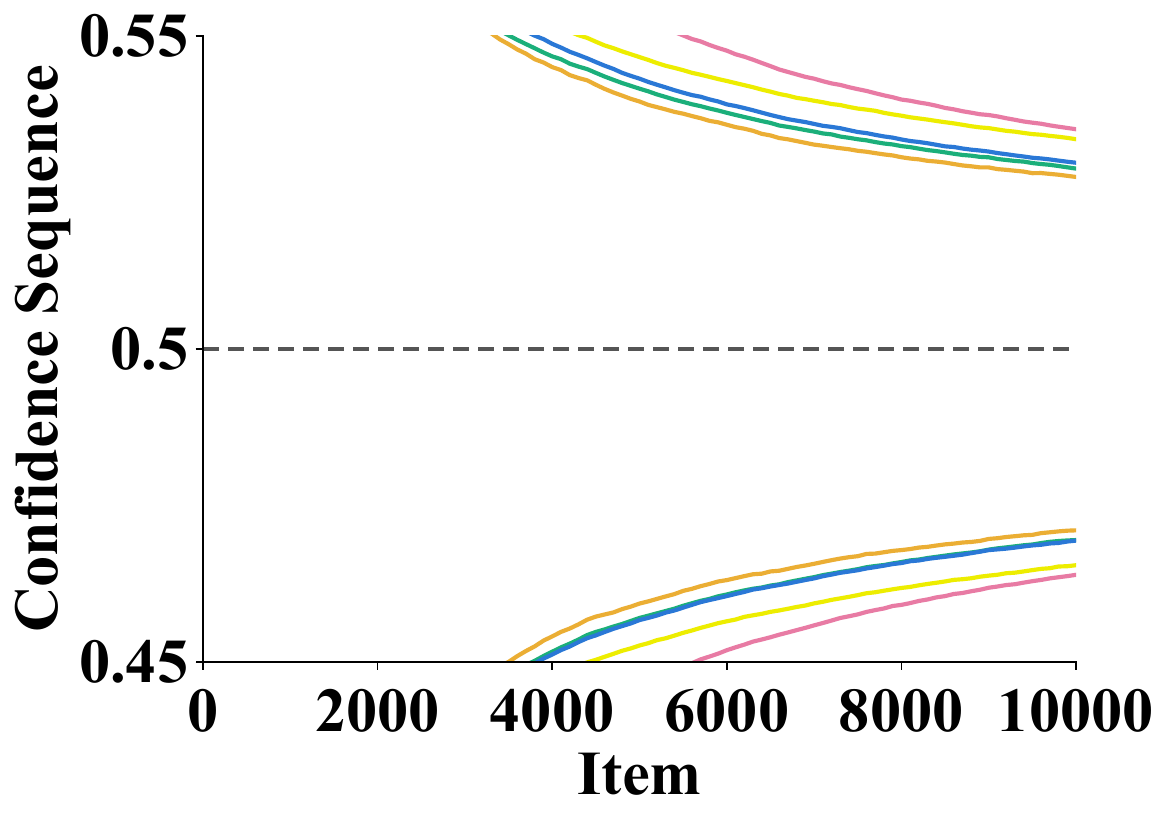}
}
\subfloat[Anytime Coverage]{%
    \includegraphics[width=0.33\textwidth]{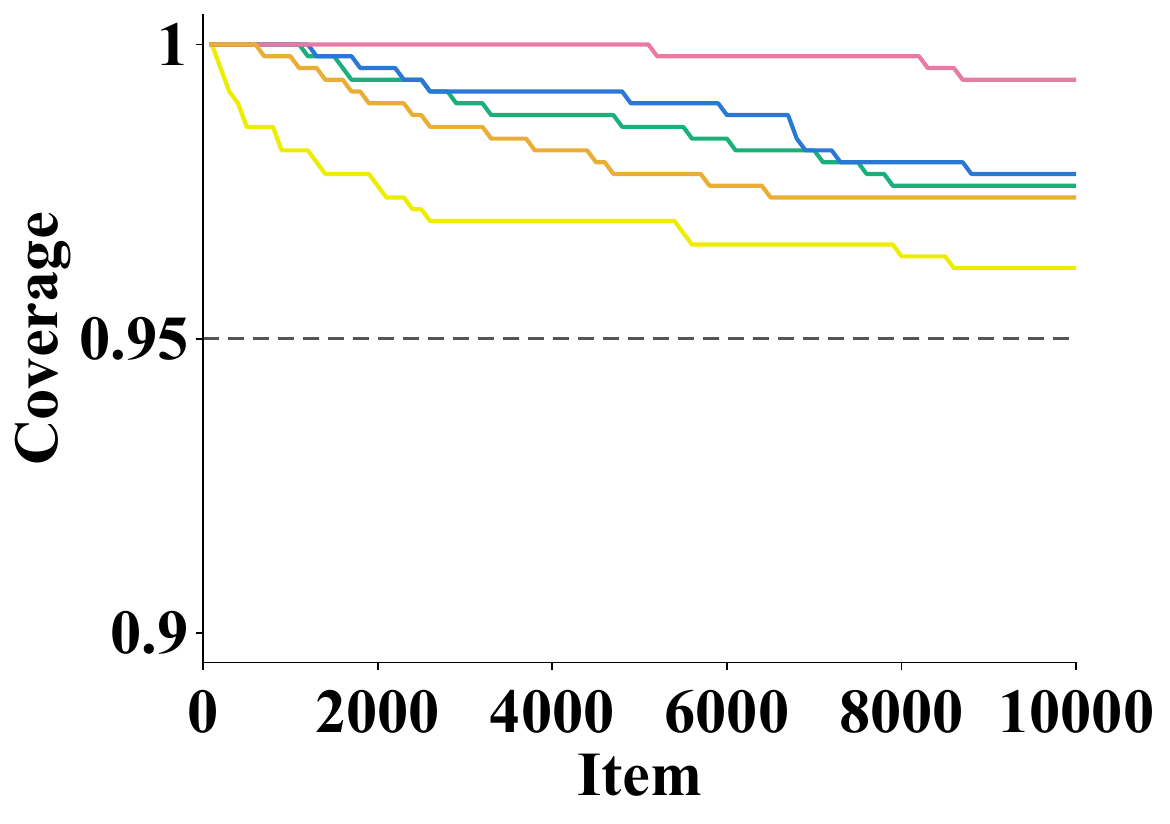}
}
\subfloat[Width]{%
    \includegraphics[width=0.33\textwidth]{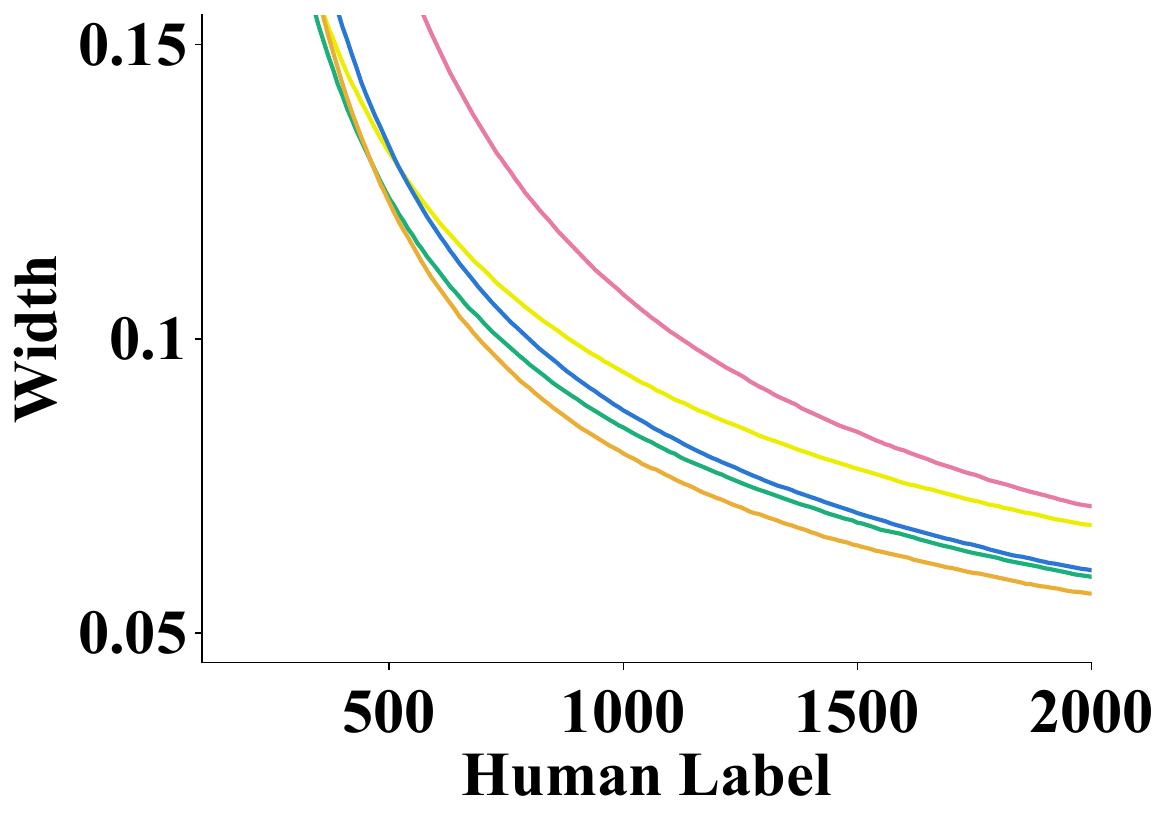}
}
\caption{Robustness to the nuisance learner: $\widehat m$ and $\widehat\sigma$ are fitted by support vector regression (SVR) with an RBF kernel.}
\label{figure:a10}
\end{figure}

\begin{figure}[H]
\centering
\includegraphics[width=0.7\textwidth]{Figure/Legend.pdf}
\subfloat[Confidence Sequences]{%
    \includegraphics[width=0.34\textwidth]{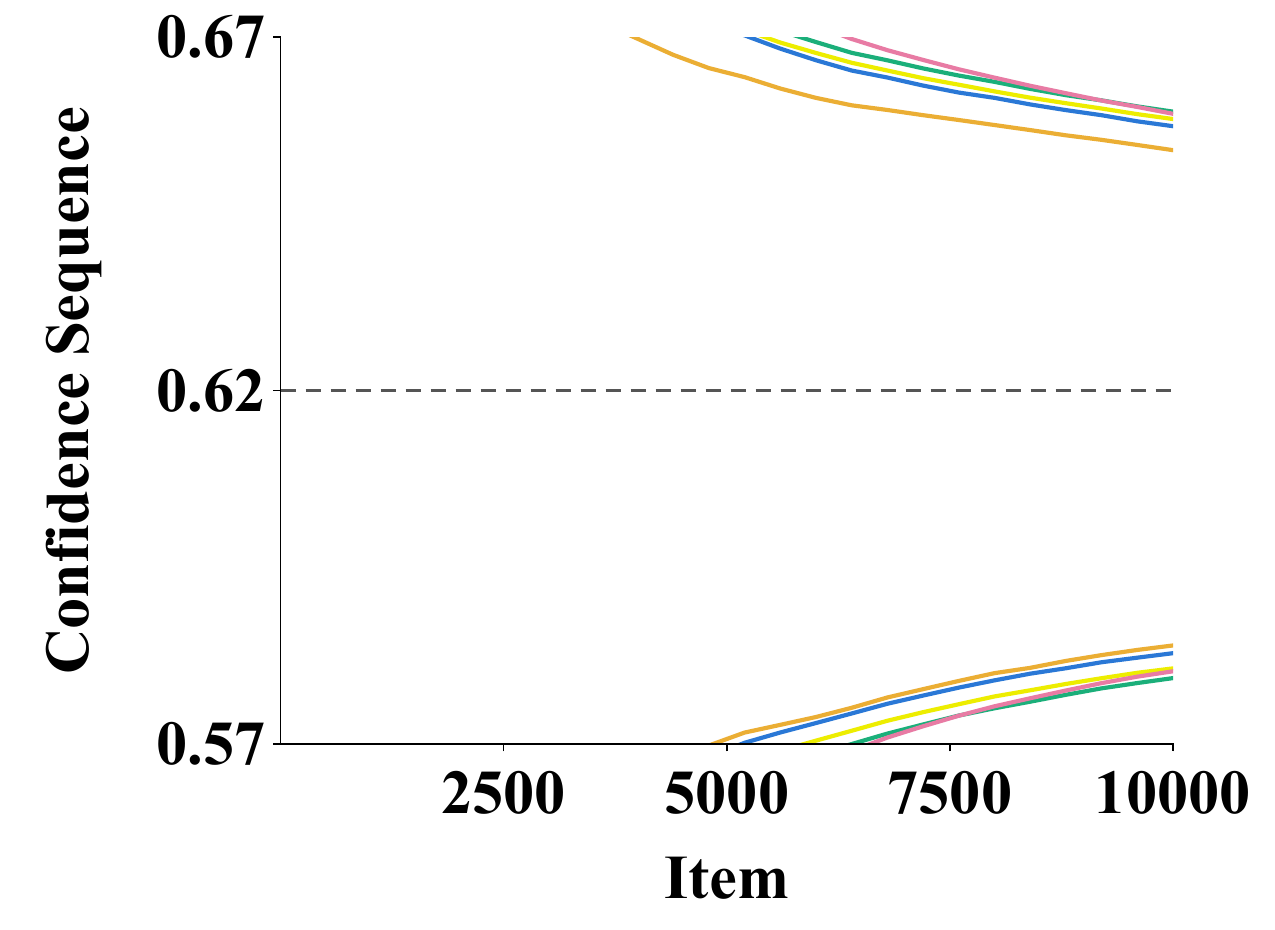}
}
\subfloat[Anytime Coverage]{%
    \includegraphics[width=0.34\textwidth]{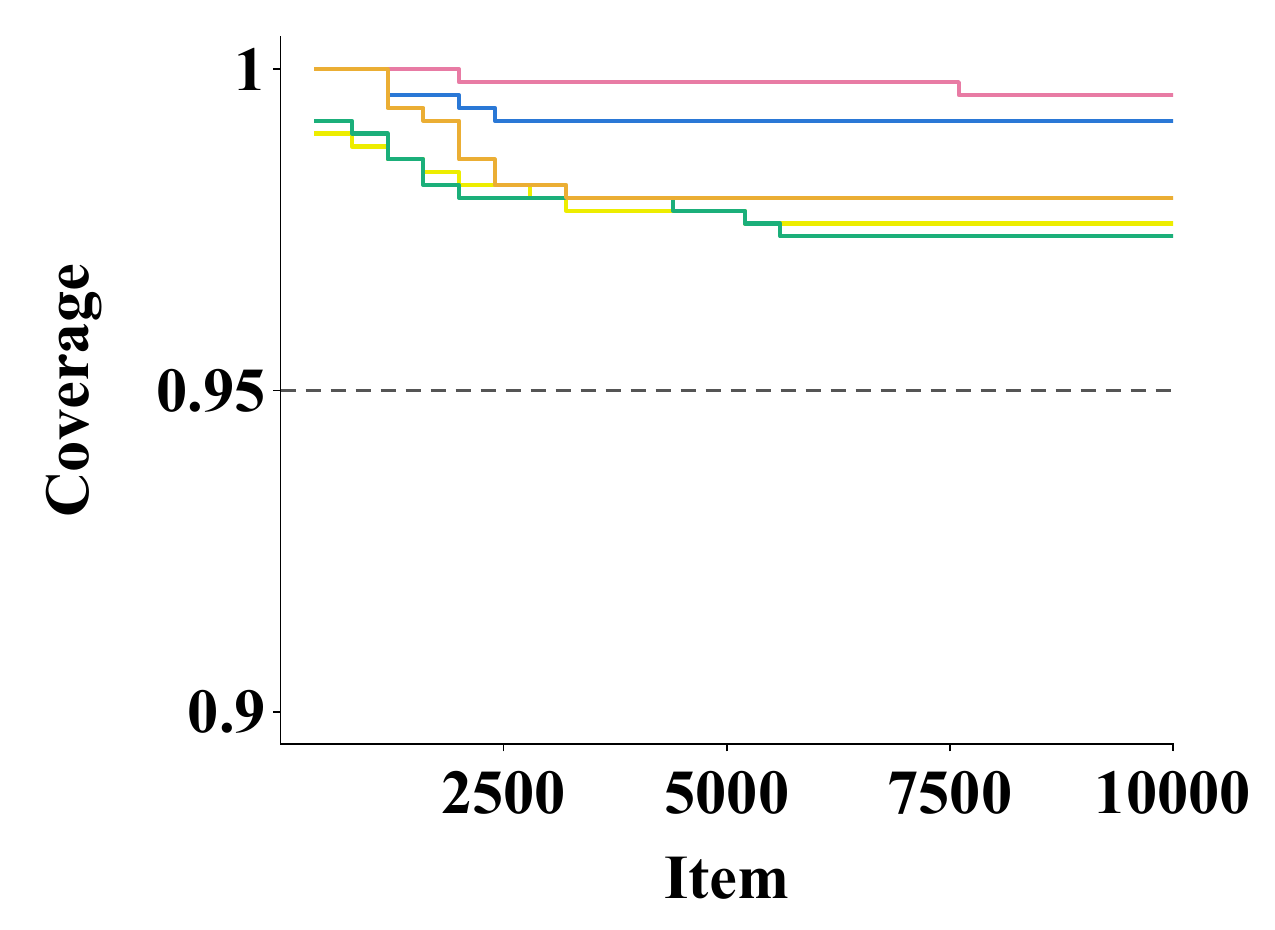}
}
\subfloat[Width]{%
    \includegraphics[width=0.34\textwidth]{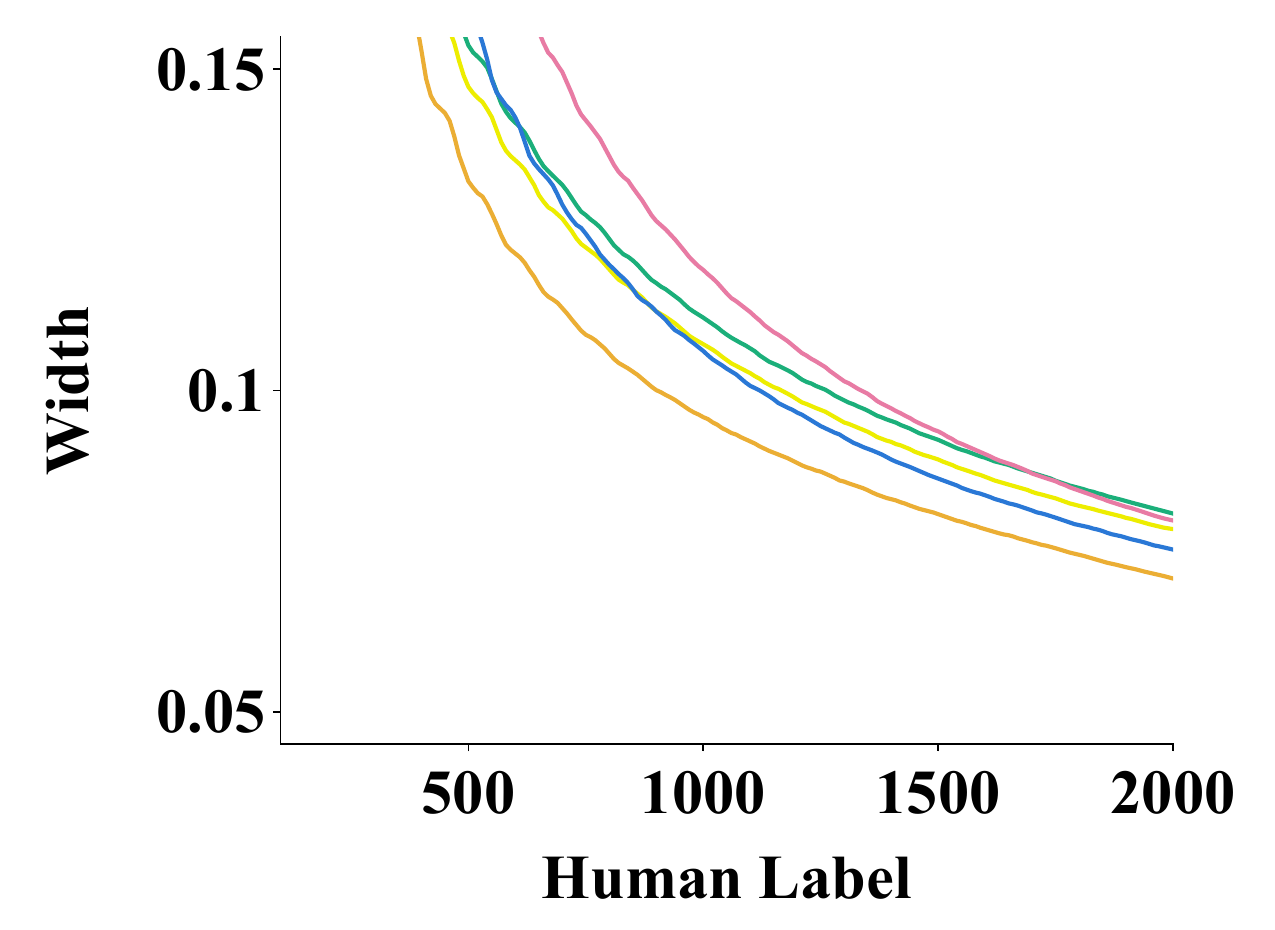}
}
\caption{Anytime-valid inference on Chatbot Arena with a single LLM judge (GPT-6-Luna) and human budget $\rho=0.2$.}
\label{figure:chatbox1}
\end{figure}

\begin{figure}[H]
\centering
\includegraphics[width=0.7\textwidth]{Figure/Legend.pdf}
\subfloat[Confidence Sequences]{%
    \includegraphics[width=0.34\textwidth]{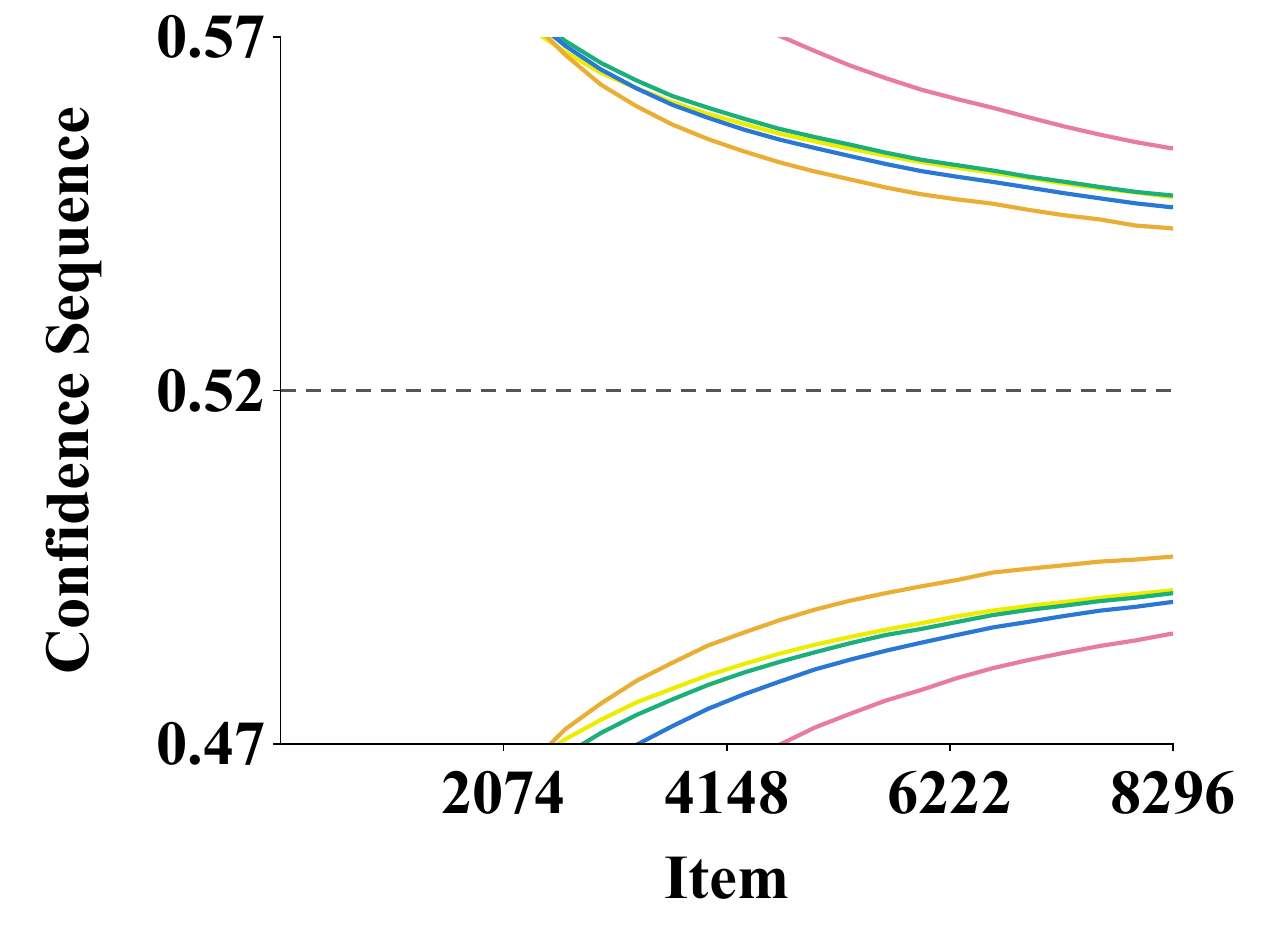}
}
\subfloat[Anytime Coverage]{%
    \includegraphics[width=0.34\textwidth]{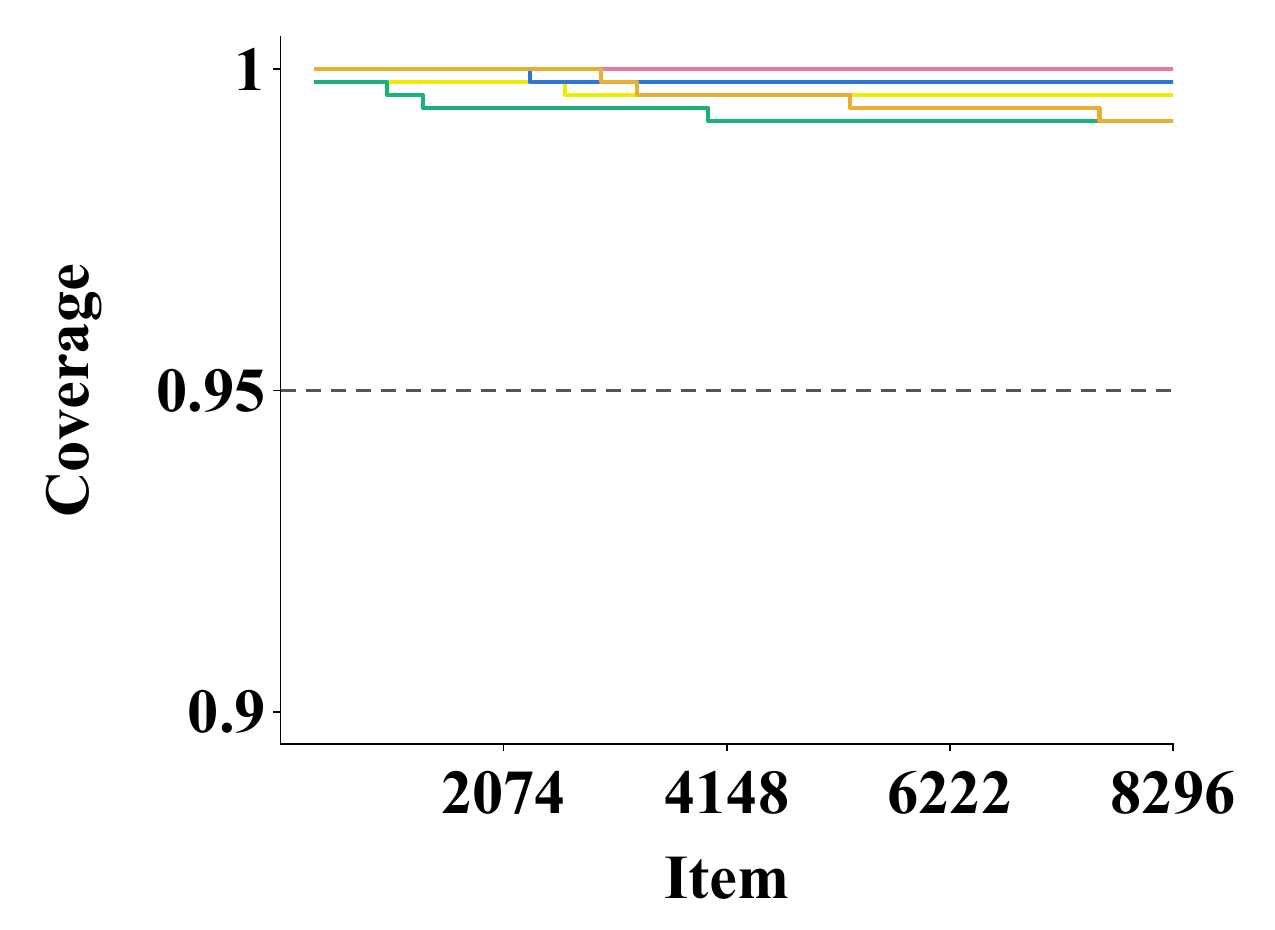}
}
\subfloat[Width]{%
    \includegraphics[width=0.34\textwidth]{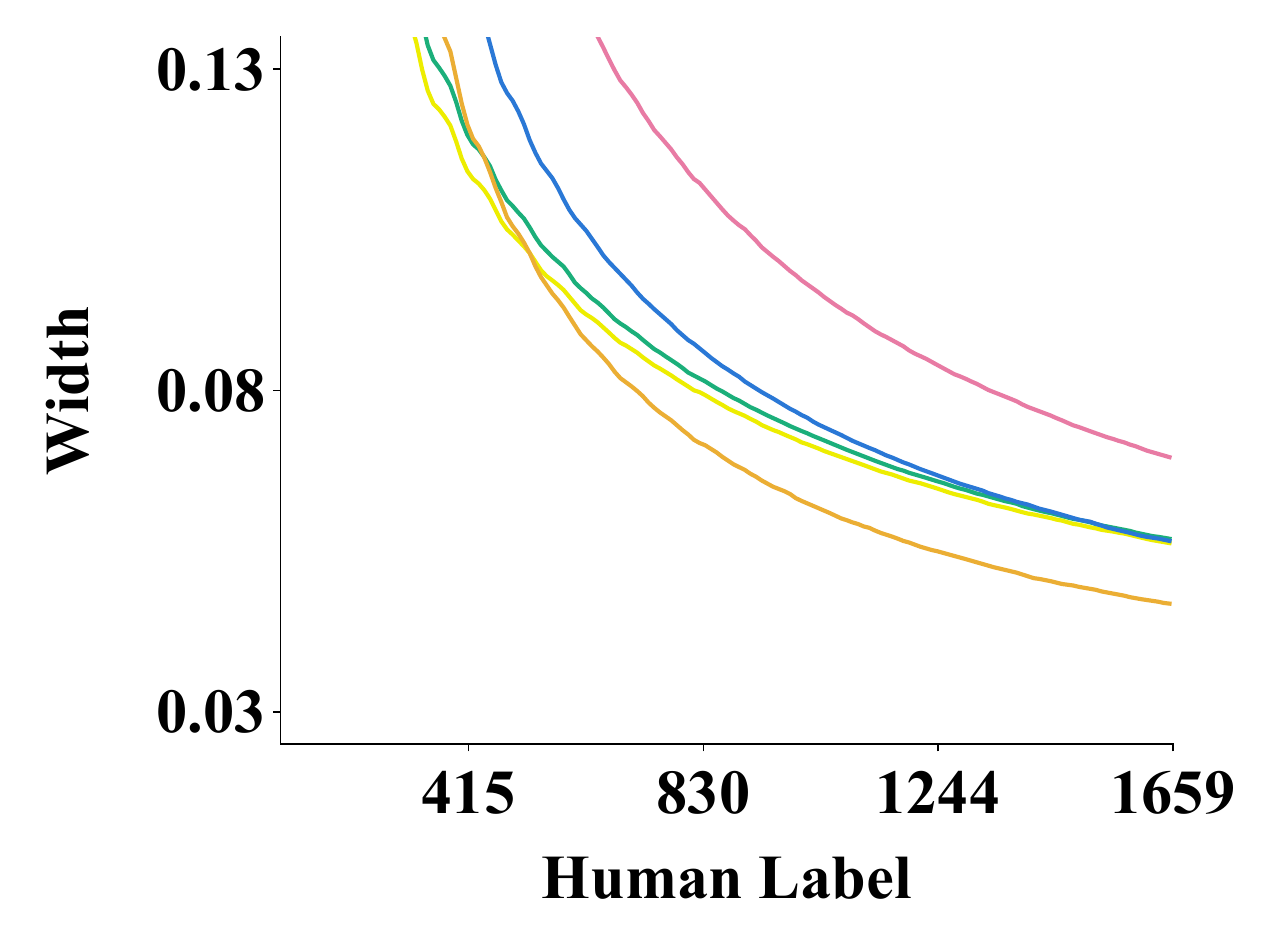}
}
\caption{Anytime-valid inference on AllSides with a single LLM judge (GPT-6-Luna) and human budget $\rho=0.2$.}
\label{figure:allsides1}
\end{figure}

\begin{figure}[H]
\centering
\includegraphics[width=0.7\textwidth]{Figure/Legend.pdf}
\subfloat[Confidence Sequences]{\includegraphics[width=0.34\textwidth]{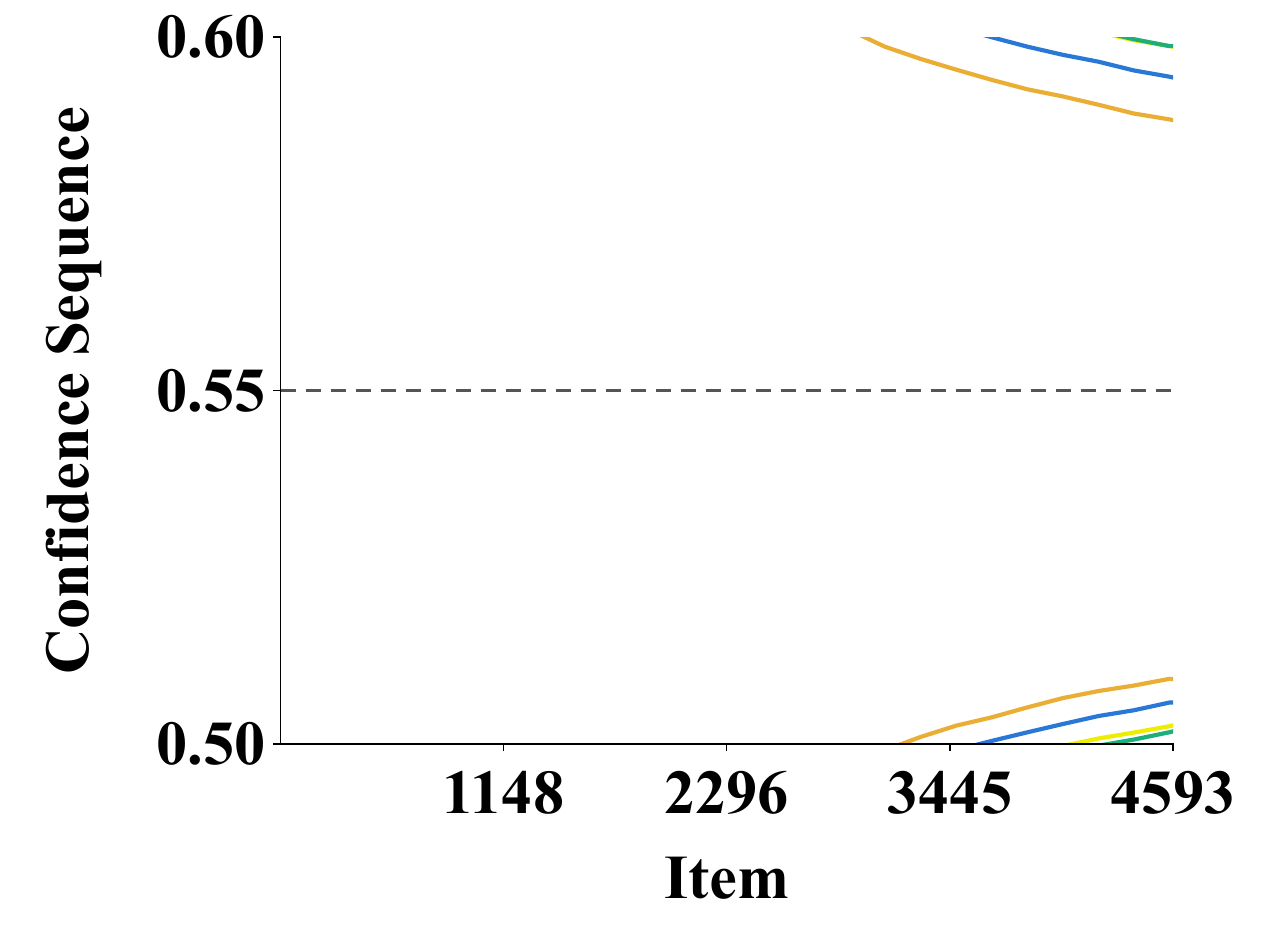}}
\subfloat[Anytime Coverage]{\includegraphics[width=0.34\textwidth]{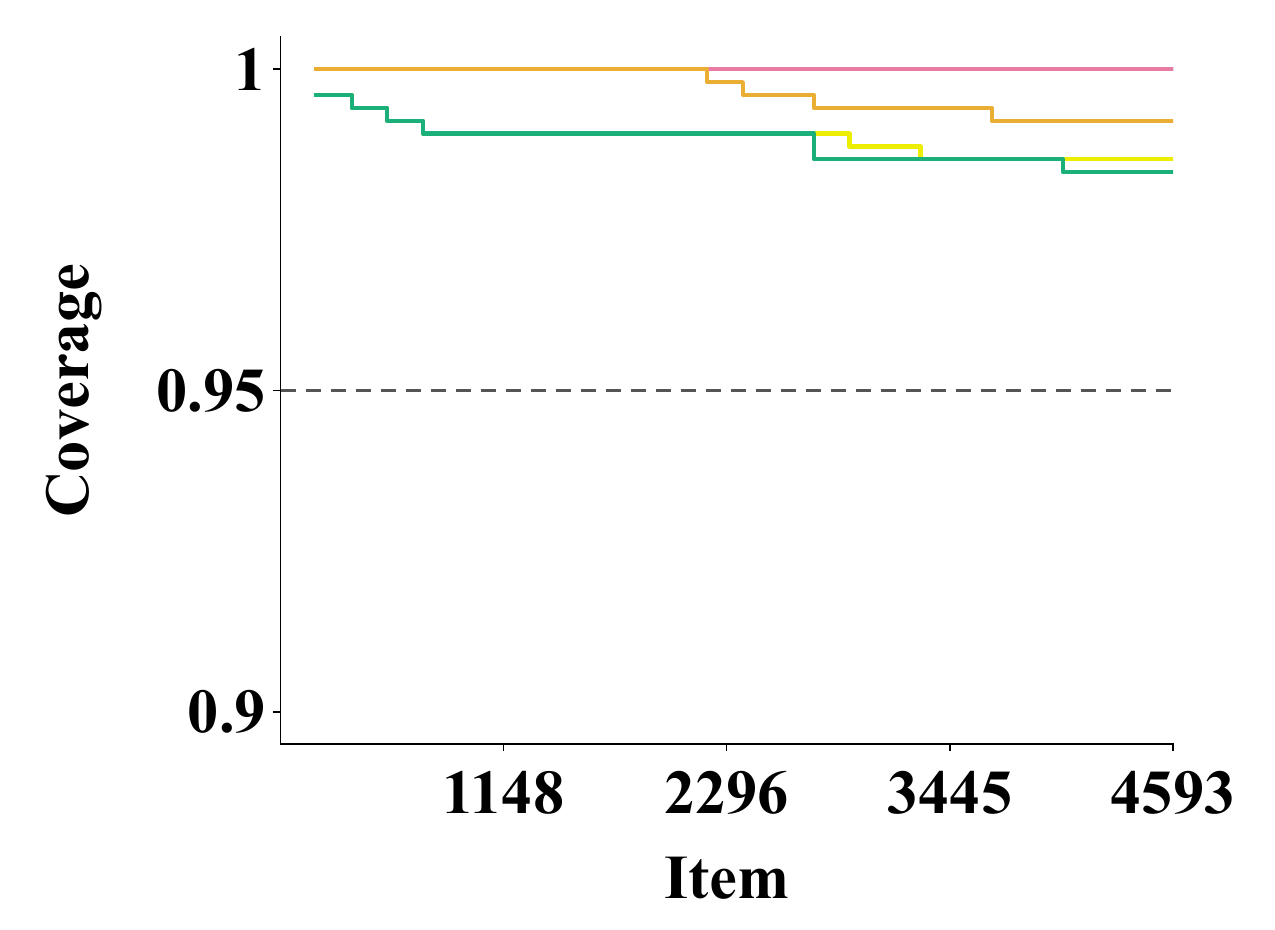}}
\subfloat[Width]{\includegraphics[width=0.34\textwidth]{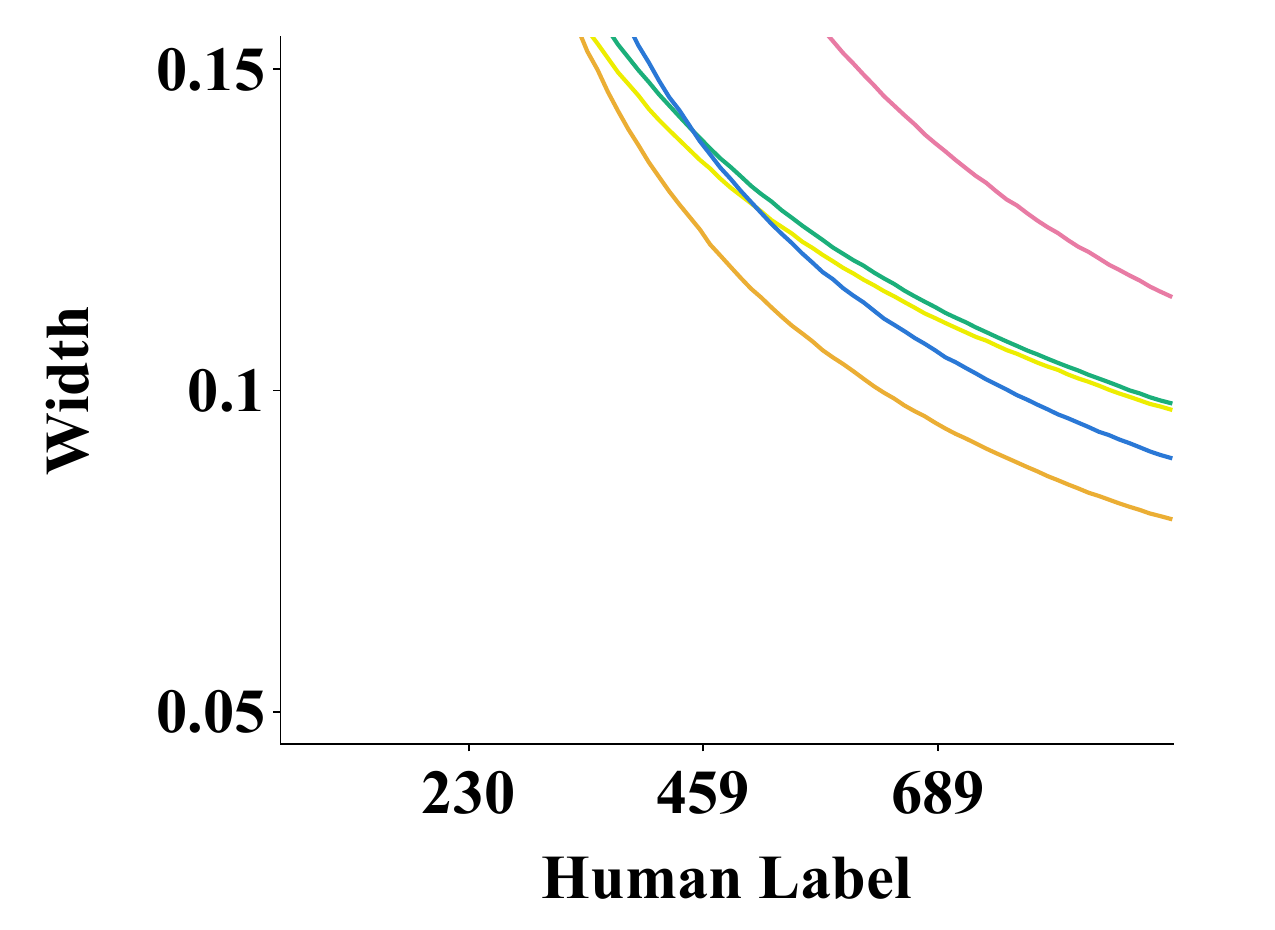}}
\caption{Anytime-valid inference on Misinfo with a single LLM judge (GPT-6-Luna) and human budget $\rho=0.2$.}
\label{figure:misinfo1}
\end{figure}

\end{document}